\documentclass[11pt,reqno]{article}

\usepackage{newtxtext,newtxmath}
\usepackage{parskip}
\usepackage{microtype}
\usepackage[labelformat=simple]{subcaption}

\usepackage{booktabs}
\usepackage{multirow}
\usepackage{natbib}
\usepackage{authblk}
\usepackage{wrapfig}
\usepackage{float}
\usepackage{graphicx}
\usepackage{titletoc}

\usepackage{url}
\usepackage{enumitem}
\usepackage{amsfonts}
\usepackage{bm}

\usepackage{amsthm}

\usepackage[dvipsnames,table]{xcolor}
\usepackage{colortbl}
\definecolor{Amaranth}{rgb}{0.9, 0.17, 0.31}
\usepackage[colorlinks=true]{hyperref}%
\usepackage[capitalize, nameinlink, noabbrev]{cleveref}
\hypersetup{
	hidelinks,
    colorlinks=true,
    linkcolor=Blue,
    filecolor=Blue,
    citecolor=Blue,
    urlcolor=Blue,
    pdftitle={Sequential Local Operator Alignment for Training-Free Model Merging},
}

\usepackage[margin=1in]{geometry}
\usepackage{mathtools}
\usepackage{amsmath,amsfonts,bm}

\newcommand{\vgamma}{\bm{\gamma}}
\newcommand{\vbeta}{\bm{\beta}}

\def\eqref#1{equation~\ref{#1}}

\def\1{\bm{1}}

\def\vb{{\bm{b}}}

\def\vu{{\bm{u}}}

\def\vx{{\bm{x}}}
\def\vy{{\bm{y}}}

\def\mA{{\bm{A}}}
\def\mB{{\bm{B}}}
\def\mC{{\bm{C}}}

\def\mG{{\bm{G}}}
\def\mH{{\bm{H}}}
\def\mI{{\bm{I}}}

\def\mM{{\bm{M}}}

\def\mR{{\bm{R}}}
\def\mS{{\bm{S}}}

\def\mU{{\bm{U}}}
\def\mV{{\bm{V}}}
\def\mW{{\bm{W}}}
\def\mX{{\bm{X}}}
\def\mY{{\bm{Y}}}
\def\mZ{{\bm{Z}}}

\def\mLambda{{\bm{\Lambda}}}
\def\mSigma{{\bm{\Sigma}}}

\DeclareMathAlphabet{\mathsfit}{\encodingdefault}{\sfdefault}{m}{sl}
\SetMathAlphabet{\mathsfit}{bold}{\encodingdefault}{\sfdefault}{bx}{n}

\definecolor{alizarin}{RGB}{227,38,54}
\definecolor{ultramarine}{RGB}{24,13,191}
\definecolor{lightblue}{RGB}{219, 234, 254}
\definecolor{lightolive}{RGB}{220, 230, 180}
\definecolor{lightyellow}{RGB}{255, 249, 196}
\definecolor{lightpurple}{RGB}{238,226,241}
\definecolor{lightred}{RGB}{251,220,221}

\definecolor{avggray}{RGB}{245,245,245}
\definecolor{avgpurple}{RGB}{220,205,245}
\definecolor{avgblue}{RGB}{205,225,250}

\newcommand{\avgcell}[1]{\cellcolor{avggray}#1}

\newif\ifrevisions
\revisionsfalse

\ifrevisions
    
\else
    
\fi

\usepackage{times}

\title{Training-Free Transformer Merging via Sequential \\ Local Operator Alignment}
\author[1]{Akansh Maurya}
\author[2, 3]{Ya-Wei Eileen Lin}
\author[2,3, 4]{Stefanie Jegelka}
\author[1]{\\Sebastian U.~Stich}
\author[1]{Rotem Mulayoff}

\affil[1]{CISPA Helmholtz Center for Information Security}
\affil[2]{Munich Center for Machine Learning}
\affil[3]{Technical University of Munich, School of Computation, Information and Technology}
\affil[4]{Massachusetts Institute of Technology, Department of EECS, CSAIL}

\date{ }
\begin{document}

\setlength{\abovedisplayskip}{3pt}
\setlength{\belowdisplayskip}{3pt}
\frenchspacing
\raggedbottom

\maketitle
\begin{abstract}
    Training-free model merging aims to combine multiple fine-tuned models into a single model without further optimization on labeled data. Yet, in transformers,  independently merging individual layers can affect a shared attention computation because the query–key and value–output operators depend on composed matrices,  overlooking the \emph{functional} structure.
    Moreover, when merging earlier components, downstream components receive different activations than they do in the original model, thus, the merged and original execution paths no longer match.
    In this paper, we introduce \emph{Sequential Local Operator Alignment}, a training-free method that merges transformers along the execution path of the partially merged model.
    Our method uses calibration data to estimate the local behavior of each functional component, aligns operators sequentially under the intermediate activation of the partially merged model, and subsequently factorizes the merged operators back into valid transformer parameters. We empirically show that this sequential step reduces error accumulation across layers. Furthermore,  the proposed operator factorization step enables rank expansion, providing a principled mechanism for increasing multi-task capacity. We demonstrate that our approach generalizes across modalities, model scales, and varying numbers of tasks, from CLIP and RoBERTa to billion-parameter LLMs, and further extends naturally to the merging of LoRA-fine-tuned models.
    The results indicate improvements over strong merging baselines without requiring rank expansion, while optional expansion provides a further accuracy–inference-cost trade-off. Project link: \url{https://akansh12.github.io/SLOA-Merge/}
\end{abstract}

\section{Introduction}
\label{sec:introduction}
Modern neural networks are often adapted into many task-specific variants from the same pretrained initialization~\citep{bommasani2021opportunities, devlin2019bert, raffel2020exploring, radford2021learning, ding2023parameter}.
However, maintaining a separate model for every task is expensive \citep{guo2021parameter, wortsman2022model, yadav2024matters}, while retraining a single multi-task model can require substantial data, compute, and engineering effort \citep{ruder2017overview, crawshaw2020multi, yang2024adamerging}.
\textit{Training-free model merging }\citep{xu2024training, fu2025training, matena2022merging, ilharco2023editing, ortizjimenez2023task, jin2023dataless}
provides a practical alternative: given several independently fine-tuned models, the goal is to combine them into one model without further training.
An effective merged model should preserve the
task-specific capabilities of the original models while offering the deployment efficiency of a single network \citep{yang2026model}.\looseness=-1

Many existing training-free merging methods operate directly on \emph{individual} parameter tensors or \emph{individual} activation spaces.
For example, they average corresponding weight matrices \citep{wortsman2022model}, interpolate task vectors \citep{ilharco2023editing}, or align intermediate activations layer by layer \citep{ainsworth2023git, singh2020model}.
While simple and often effective, this layer-wise perspective is parameterization-dependent: a model's function is not uniquely tied to the particular tensors used to represent it \citep{zhao2025symmetry,burgin2026beyond,  bona2023parameter, elhage2021mathematical, dar2023analyzing}.
For instance, in a transformer attention head \citep{vaswani2023attentionneed}, the matrices $\mathbf{W}_Q$, $\mathbf{W}_K$, $\mathbf{W}_V$, and $\mathbf{W}_O$ are not used as independent functional units.
The query and key determine the pre-softmax attention scores through the combined bilinear operator $\mW_Q\mW_K^\top$, while the value and output determine the post-attention contribution through the composed operator $\mW_V\mW_O$.
Thus, the functionally meaningful objects are the composed operators $\mW_Q\mW_K^\top$ and $\mW_V\mW_O$.
A merging procedure that independently combines $\mathbf{W}_Q$ and $\mathbf{W}_K$, or $\mathbf{W}_V$ and $\mathbf{W}_O$, may therefore fail to preserve a shared attention computation even when that computation is functionally equivalent.
\begin{figure*}[t]
    \centering
     \includegraphics[width=1\textwidth]{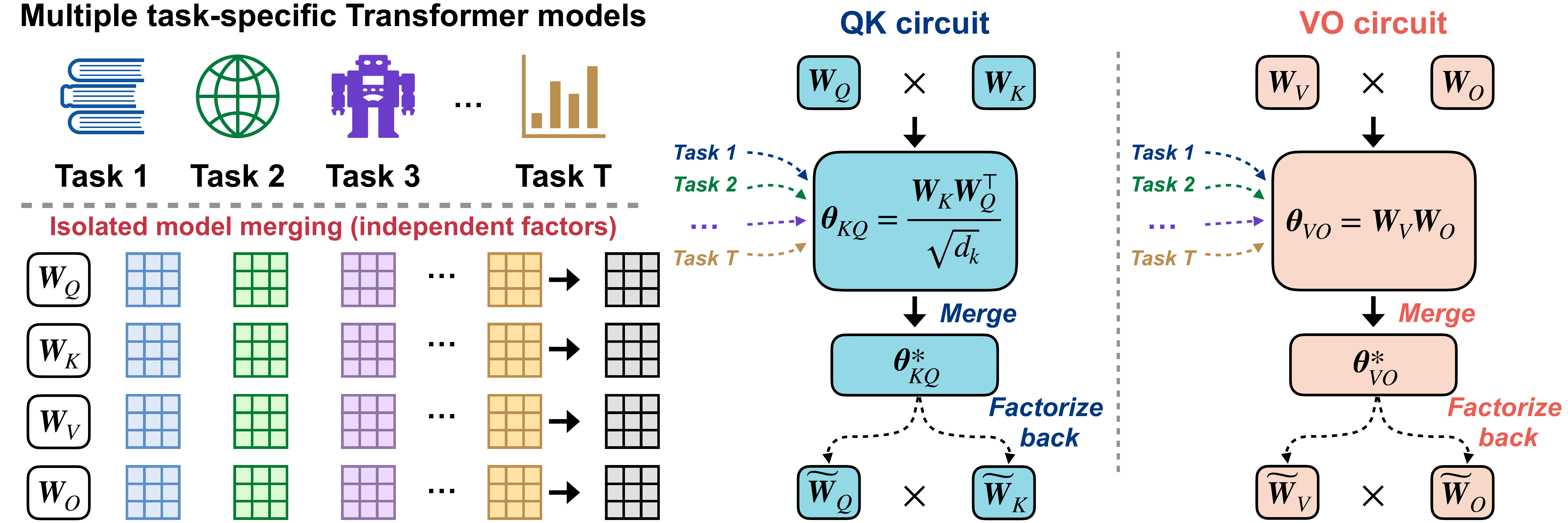}
    \caption{
    Illustration of \emph{Sequential Local Operator Alignment}.
Our method aligns the effective local functional operators, merges them
sequentially, and then factorizes the aligned operators.
This sequential operator-level view respects coupled attention computation, reduces error accumulation across layers, and enables rank expansion to have additional multi-task capacity.
    }
\label{fig:illstration}
    \vspace{-1 em}
\end{figure*}

A natural way is to merge objects at the level at which they enter the model's computation, rather than at the level of an arbitrary parameter decomposition.
To this end, we introduce \emph{Sequential Local Operator Alignment} (SLOA), a training-free framework for merging transformer models by aligning \emph{local functional operators} (see Fig.~\ref{fig:illstration} for the illustration).
SLOA associates each model component with the local operator.
For each component, SLOA constructs a merged operator from the task models.
SLOA considers two alignment criteria: \textit{(i)} \emph{behavior matching},  on calibration inputs to the component, the merged operator should produce outputs close to those of the task-specific operators, and \textit{(ii)} \emph{data-aware operator regularization}, the merged operator is kept close to the task-specific operators under a metric induced by the calibration activations.
This metric emphasizes directions that are active on the data.
Importantly, our contribution is to instantiate them as a unified local-operator alignment framework for training-free transformer merging.  The same objective applies across LayerNorm modules, linear maps, QK attention operators, and VO attention operators by changing only the behavior map and the induced metric.\looseness=-1

A key feature of our framework is that alignment is performed \emph{sequentially}.
When earlier operators have already been merged, the  hidden activations  entering later components are no longer identical to the activations produced by any individual task model.
If later components are merged only under the original task-model activations, the local objective assumes that all previous merges were perfect. Our approach instead evaluates the activation-matching term on the activations produced by the partially merged model. This makes each local merge corrective, aligning downstream operators along the actual execution path of the merged network and reducing error accumulation across layers.\looseness=-1

Our operator view separates the merged operator from its eventual factorization.
For QK and VO components, each individual task head has rank at most the head dimension, but the merged multi-task operator may require a higher-rank representation, as different tasks can use different subspaces. Therefore, we treat rank as a \emph{capacity-control mechanism}. After merging an operator, we can factorize it at the original rank to preserve the architecture, or \emph{expand} the rank to retain additional singular directions that capture task-specific behavior. Empirically, our method achieves state-of-the-art training-free merging performance across vision benchmarks, RoBERTa models, and billion-parameter LLMs. Moreover, SLOA naturally extends to models fine-tuned with LoRA~\citep{hu2022lora}, further broadening its applicability to parameter-efficient adaptation settings.
It consistently outperforms baselines that merge individual weights or rely on isolated activation alignment, supporting the central claim that model merging should be performed at local functional operators.

\section{Motivation: limitations of isolated attention-weight merging}\label{sec:motivation}
In a transformer attention head, the query, key, value, and output layers $\mW_Q$, $\mW_K$, $\mW_V$, and $\mW_O$ are often merged independently, or their corresponding intermediate activations are aligned separately. Given token activations $\mX$, the query and key weights determine the pre-softmax attention scores only through the composed QK operator $\mM_{\mathrm{QK}}=\mW_Q \mW_K^\top$ and $\mathrm{scores}(\mX)=\mX \mM_{\mathrm{QK}} \mX^\top$.
Similarly, for a fixed attention pattern $\mA$, the value and output weights determine the head contribution only through the composed VO operator $
    \mM_{\mathrm{VO}} = \mW_V \mW_O$ and $\mathrm{head}(\mX) = \mA \mX \mM_{\mathrm{VO}}$. Thus, the functional objects are the composed operators $\mM_{\mathrm{QK}}$ and $\mM_{\mathrm{VO}}$, whereas the individual matrices $\mW_Q,\mW_K,\mW_V,\mW_O$ are only one possible factorization of these operators.\looseness=-1

\begin{wrapfigure}[14]{r}{0.44\textwidth}
    \centering
     \includegraphics[width=0.43\textwidth]{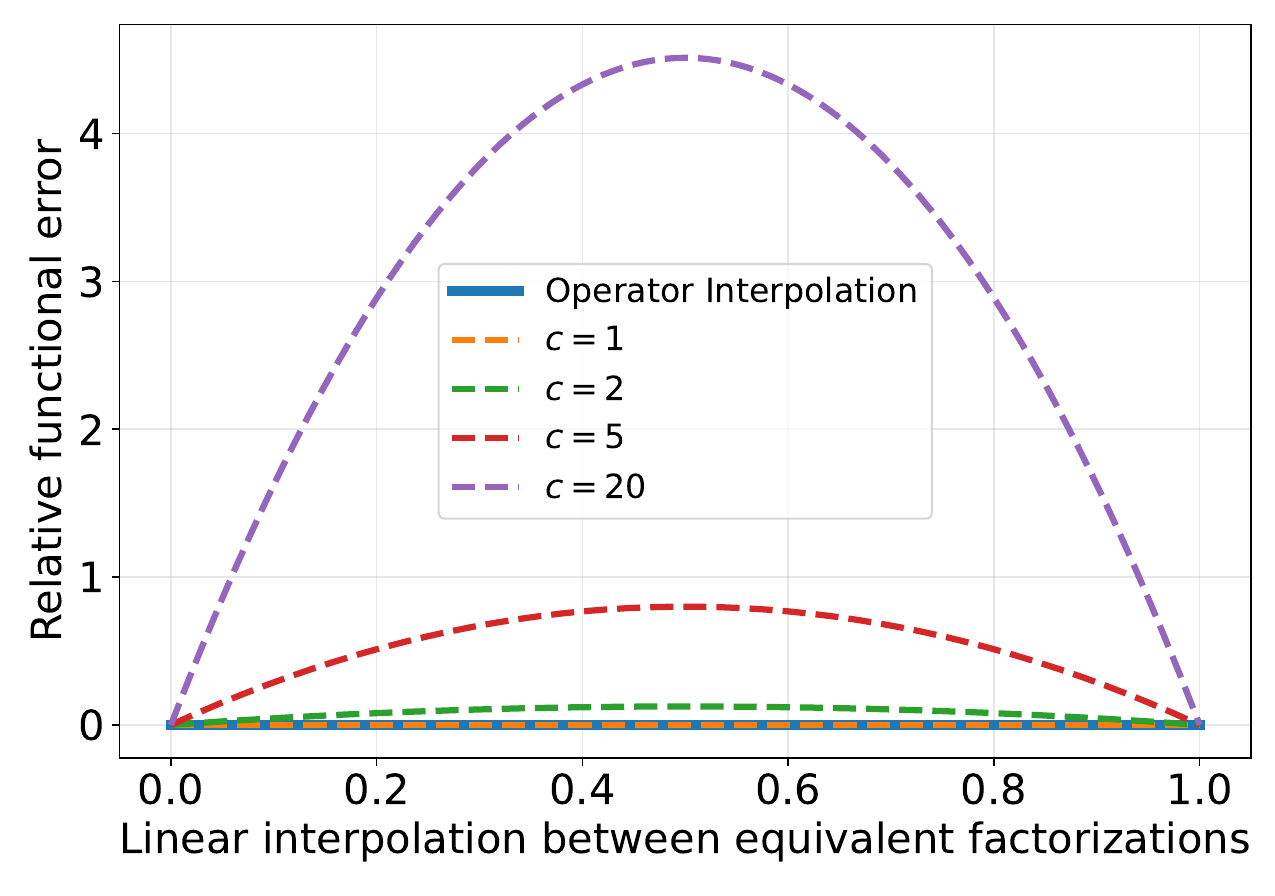}
    \caption{Illustration of the non-invariance of independent merging in attn. operator.}
    \label{fig:scalar_exp}
\end{wrapfigure}
\noindent\textbf{Example~1.}
Consider two QK factorizations that represent the same attention-score operator~$\mM_{\mathrm{QK}}^{(1)} = \mW_Q \mW_K^\top$, and $\mM_{\mathrm{QK}}^{(2)} = (\mW_Q \mR)(\mW_K \mR^{-\top})^\top =\mW_Q \mW_K^\top$. Although the two models are functionally identical at the QK level, independently averaging the query and key weights gives $ \bar{\mW}_Q = \frac{1}{2}\mW_Q(\mI+\mR)$, and $\bar{\mW}_K = \frac{1}{2}\mW_K(\mI+\mR^{-\top})$.
The resulting merged QK operator is $\bar{\mM}_{\mathrm{QK}} = \bar{\mW}_Q \bar{\mW}_K^\top = \frac{1}{4}\mW_Q (\mI+\mR)(\mI+\mR^{-1}) \mW_K^\top$, which is not equal to $\mW_Q\mW_K^\top$.
As a special case, let $\mR = c\mI$ with $c>0$.  Then the merged QK operator becomes $\bar{\mM}_{\mathrm{QK}} = \frac{(1+c)(1+c^{-1})}{4} \mM_{\mathrm{QK}}$. The multiplicative factor equals $1$ only when $c=1$, and it grows without bound as $c\rightarrow\infty$. Thus, for any input $\mX$ s.t. $\mX\mM_{\mathrm{QK}}\mX^\top\neq 0$, the merged attention scores can differ arbitrarily from the shared scores of the two original models
\begin{equation}
    \left\|
    \mX\bar{\mM}_{\mathrm{QK}}\mX^\top
    -
    \mX\mM_{\mathrm{QK}}\mX^\top
    \right\|_F
    =
    \left|
    \frac{(1+c)(1+c^{-1})}{4}
    -
    1
    \right|
    \left\|
    \mX\mM_{\mathrm{QK}}\mX^\top
    \right\|_F.
\end{equation}
The same argument applies to VO under $(\mW_V\mR,\ \mR^{-1}\mW_O)$. Our solution is to treat $\mW_Q\mW_K^\top$ (and $\mW_V\mW_O$) as a single unit for merging (Fig.~\ref{fig:scalar_exp}) and re-factorize after merging. This has two advantages: \textit{(1)} we forgo the parameterization ambiguity and any search for a good aligning transformation; \textit{(2)} the rank of the merged operator can be chosen freely at factorization time, allowing us to adjust capacity.

\section{Methodology: Sequential local operator alignment (SLOA)}
\label{sec:method}
The above example shows that independent merging can produce arbitrary functional errors when computation is performed inside a transformer. To address this, we introduce \emph{Sequential Local Operator Alignment}, a training-free method that merges models by aligning the local functional operators.

\subsection{Local functional operators}
\label{sec:local_operators_transformer}
We define local functional operators as the parameterized object that determines a component's local computation once its input activation and any non-learned part are fixed.
For example, the local operator of a linear layer is its weight matrix, while the local operator of an attention head is more naturally expressed through the composed QK and VO operators (see Fig.~\ref{fig:illstration}).

We write $\bm{\theta}$ for a generic local operator and $\mathcal{A}(\mZ;\bm{\theta})$ for its behavior map, where $\mZ$ denotes the local input needed to evaluate the operator.
For most operators, $\mZ$ is the activation $\mX$, and for VO, it includes both the attention matrix and the residual state. Table~\ref{tab:ops_metrics} summarizes the local functional operators used in our transformer merging procedure.
The key design criterion is that, for fixed local input $\mZ$, the behavior map $\mathcal{A}(\mZ;\bm{\theta})$ is linear in the local operator $\bm{\theta}$.
Specifically, LayerNorm is linear in $(\vgamma,\vbeta)$ after the normalization statistics are fixed; a linear layer is linear in $\mW$; QK is linear in the composed score operator $\mW_Q\mW_K^\top/\sqrt{d_k}.$; and VO is linear in the composed operator $\mW_V\mW_O$. \looseness=-1

The same template can be applied beyond the operators in Table~\ref{tab:ops_metrics} whenever a module admits a local representation that is linear in the chosen operator.
For example, in a gated MLP, the down-projection can be treated as a local linear operator after the gated hidden activation has been formed.
Operators that remain nonlinear in their parameters after conditioning on the local input would require either a smaller linear suboperator or an additional approximation, which are not the focus of this work.
\begin{table}[t]
\caption{
Local functional operators and their data-dependent metrics.
Here, $\nu(\mX)$ denotes the normalized input before the learned LayerNorm's scale and shift, and $\mA$ denotes the attention matrix
}
\begin{equation*}
\begin{array}{c|c|c|c}
\text{Component}
&
\text{Local operator $\bm{\theta}$}
&
\text{Behavior map $\mathcal{A}(\mZ;\bm{\theta})$}
&
\text{Metric}
\\
\hline
\text{LayerNorm}
&
(\vgamma,\vbeta)
&
\mathcal{A}(\mX;\vgamma,\vbeta)
=
\nu(\mX)\operatorname{diag}(\vgamma)
+
\mathbf{1}\vbeta^\top
& \operatorname{Cov}(\nu(\chi))
\\[1.5mm]
\text{Linear map}
&
\mW
&
\mathcal{A}(\mX;\mW)
=
\mX\mW
&
\mLambda
\\[1.5mm]
\text{QK component}
&
\mW_{Q}\mW_{K}^\top/\sqrt{d_k}
&
\mathcal{A}(\mX;\bm{\theta})
=
\mX\bm{\theta}\mX^\top
&
\mLambda\otimes\mLambda
\\[1.5mm]
\text{VO component}
&
\mW_{V}
\mW_{O}
&
\mathcal{A}(\mA\mX;\bm{\theta})
=
\mA\mX\bm{\theta}
&
\mLambda
\end{array}
\label{eq:local_operator_summary}
\end{equation*}
\label{tab:ops_metrics}
\end{table}

\subsection{Activation-based local behavior matching}
\label{sec:activation_matching}

Suppose we merge $T$ task-specific models.
For a fixed local operator, let $\bm{\theta}_t$ denote the corresponding operator in model $t$, and let $n_t$ be the number of calibration samples for that model.
For calibration example $i$, let $\mZ_{t,i}$ be the local input received by the operator when running the original task model $t$, and let $\widehat{\mZ}_{t,i}$ be the local input received by the same operator when running the partially merged model.
We also define the corresponding local state collections $\chi_t=\{\mZ_{t,i}\}_{i=1}^{n_t}$ and $\widehat{\chi}_t=\{\widehat{\mZ}_{t,i}\}_{i=1}^{n_t}$.

The task-model output for this local operator is $\mathcal{A}(\mZ_{t,i};\bm{\theta}_t)$, while the merged operator, evaluated on the state it will actually receive during the merged model's forward pass, produces $\mathcal{A}(\widehat{\mZ}_{t,i};\bm{\theta})$. Thus, the activation-based local alignment loss is then given by
\begin{equation}
\mathcal{L}_{\mathrm{act}}(\bm{\theta})
=
\frac{1}{T}
\sum_{t=1}^{T}
\frac{1}{n_t}
\sum_{i=1}^{n_t}
\left\|
\mathcal{A}(\widehat{\mZ}_{t,i};\bm{\theta})
-
\mathcal{A}(\mZ_{t,i};\bm{\theta}_t)
\right\|_F^2 .
\label{eq:local_activation_matching}
\end{equation}
Equivalently, Eq.~(\ref{eq:local_activation_matching}) first samples a model uniformly from $\{1,\ldots,T\}$ and then samples a calibration example uniformly from that model.
This balanced empirical average prevents models with more calibration examples from dominating the local objective.
Note that Eq.~(\ref{eq:local_activation_matching}) is functional: the merged operator is asked to reproduce the local outputs of the task-specific operators.
However, with a small calibration set, activation matching alone can be underdetermined.
Many operators may agree on the observed local activations while behaving differently on unobserved activations.
We therefore add a data-dependent weight-space geometry term introduced in the next subsection.

\subsection{Data-dependent weight geometry}
\label{sec:data_dependent_geometry}

Pure weight-space merging is stable and sample-efficient, but it treats all parameter directions as equally important.
Pure activation matching focuses on functionality, but with only a small calibration set, it can be
underdetermined and sensitive to the observed samples.
SLOA considers these two views jointly: the activation term matches local behavior, while the geometry term keeps the merged operator close to the task-specific operators in directions that are important under the calibrations.
For a standard one-sided linear operator, such as $\mX\mW$, we define
\begin{equation}
\label{eq:one_sided_metric}
    \left\|
\bm{\theta}
-
\bm{\theta}_t
\right\|_{\widehat{\chi}_t}^{2}
=
\operatorname{Tr}
\left[
(\bm{\theta}-\bm{\theta}_t)^\top
\mLambda_t
(\bm{\theta}-\bm{\theta}_t)
\right],
\end{equation} where $\mLambda_t
=
\alpha
\operatorname{diag}
\left(
\operatorname{Cov}_{\vu\in \phi(\widehat{\chi}_t)}(\vu)
\right)
+
\epsilon \mI$.
Here, $\phi(\widehat{\chi}_t)$ denotes the collection of token-level vectors that multiply the operator in the local behavior map.
For a linear map, $\phi(\mX)=\mX$; for the LayerNorm scale parameter, $\phi(\mX)=\nu(\mX)$; and for VO, $\phi((\mA,\mX))=\mA\mX$.
The coefficient $\alpha$ controls the strength of the data-dependent geometry, and $\epsilon \mI$ provides a ridge term for numerical stability.
Note that Eq.~(\ref{eq:one_sided_metric}) is a Mahalanobis distance in operator space: the merged operator is kept close to $\bm\theta_t$ along directions that are active on the partially merged states, and is regularized less along weakly observed ones.
For QK operators, the behavior is bilinear, and thus, we use
\begin{equation}
\left\|
\bm{\theta}
-
\bm{\theta}_t
\right\|_{\widehat{\chi}_t}^{2}
=
\left\|
\mLambda_t^{1/2}
(\bm{\theta}-\bm{\theta}_t)
\mLambda_t^{1/2}
\right\|_F^2 .
\label{eq:qk_two_sided_metric}
\end{equation}
After vectorization, this corresponds to a Mahalanobis metric with covariance $\mLambda_t\otimes\mLambda_t$.
Considering the activation-matching loss and the data-dependent geometry gives the objective
\begin{equation}
\bm{\theta}_{\mathrm{merge}}
=
\underset{\bm{\theta}\in\Theta_{\mathrm{amb}}}{\arg\min}
\;
\mathcal{L}_{\mathrm{act}}(\bm{\theta})
+
\frac{1}{T}
\sum_{t=1}^{T}
\left\|
\bm{\theta}
-
\bm{\theta}_t
\right\|_{\widehat{\chi}_t}^{2}.
\label{eq:unified_operator_alignment}
\end{equation}
Here, $\Theta_{\mathrm{amb}}$ is the ambient operator space.
For linear maps and LayerNorm affine parameters, this is the usual parameter space.
For QK and VO operators, $\Theta_{\mathrm{amb}}=\mathbb{R}^{d\times d}$ is the residual-space operator space; the rank constraint induced by attention-head width is imposed later during factorization.
Because $\mathcal{A}(\widehat{\mZ};\bm{\theta})$ is linear in $\bm{\theta}$ for fixed calibration states, Eq.~(\ref{eq:unified_operator_alignment}) is a quadratic objective.
Thus, each local operator can be merged by solving a closed-form regularized least-squares problem.
Further derivations of the local operator alignment optimal solutions are in \S\ref{app: local operator  alignment}.
\subsection{Sequential merging}
\label{sec:sequential}
The objective in Eq.~(\ref{eq:unified_operator_alignment}) is applied sequentially.
Let $\mathcal{O}$ be the set of local operators and write $o_a\prec o_b$ if the output or activation produced by operator $o_a$ can affect the input to operator $o_b$. We choose a topological order $o_1,o_2,\ldots,o_M$ that is in line with this dependency relation. At step $m$, the operators $o_1,\ldots,o_{m-1}$ have been merged. For each model $t$ and calibration $i$, we have two local inputs for $o_m$. First, $\mZ_{t,i}^{(m)}$ is the local input produced by the original task model $t$.
Second, $\widehat{\mZ}_{t,i}^{(m)}$ is the local input produced by the partially merged model, whose predecessors $o_1,\ldots,o_{m-1}$ use the merged operators.
The current operator is then merged by solving Eq.~(\ref{eq:unified_operator_alignment}) with $
    \mZ_{t,i}=\mZ_{t,i}^{(m)}$,
    $\widehat{\mZ}_{t,i}=\widehat{\mZ}_{t,i}^{(m)}$, and $
    \bm{\theta}_t=\bm{\theta}_t^{(m)}$.
This sequential construction is important. If the current operator were aligned only on $\mZ_{t,i}^{(m)}$, the local objective would implicitly assume that all previous operators had been merged perfectly. In practice, earlier merging steps may shift the activations seen by later operators.
By evaluating $\widehat{\mZ}_{t,i}^{(m)}$, SLOA aligns each operator on the activation distribution that it will actually receive inside the partially merged model. The target remains $\mathcal{A}(\mZ_{t,i}^{(m)};\bm{\theta}_t^{(m)})$, therefore,  the current operator corrects for accumulated upstream error while preserving the task-specific local behavior.

\subsection{Rank as capacity control}
\label{sec:rank_increase}
Our method allows the merged model to have a larger capacity than the task-specific models. For example, let $\mW_Q,\mW_K,\mW_V\in\mathbb{R}^{d\times d_k}$ and $\mW_O\in\mathbb{R}^{d_k\times d}$ be the parameters for an attention head from the task-specific models, then the local operators for this head are given by
\begin{equation}
    \bm{\theta}_{QK}=\mW_Q\mW_K^\top/\sqrt{d_k}
    \quad \text{and} \quad
    \bm{\theta}_{VO}=\mW_V\mW_O,
    \quad \text{s.t.} \quad
    \bm{\theta}_{QK},\bm{\theta}_{VO}\in\mathbb{R}^{d\times d}.
\end{equation}
Since these operators are realized through a width-$d_k$ bottleneck, they have rank at most $d_k$. The merged operator, however, needs to combine $T$ such operators that may occupy different subspaces.  Consequently, it is likely that using the same model width for the merged model, i.e., a rank of at most $d_k$, would result in underfitting. Therefore, it is important to let the merged model be more expressive, which in this case means a rank exceeding $d_k$. To this end, note that SLOA solves for $\bm{\theta}_{\mathrm{merge}} \in\mathbb{R}^{d\times d} $ without any constraints, while the factorization rank $r$ for the merged parameters is chosen only afterward. Let $\bm{\theta}_{\mathrm{merge}}^r = \mU_r\mSigma_r\mV_r^\top$ be the best rank-$r$ approximation of $ \bm{\theta}_{\mathrm{merge}} $, and
\begin{equation}\label{eq:rank_factorization}
    \begin{aligned}
        \text{QK:}&\quad \bar{\mW}_Q=\mU_r(\tau\mSigma_r)^{1/2}, && \ \bar{\mW}_K=\mV_r(\tau\mSigma_r)^{1/2},\\
        \text{VO:}&\quad \bar{\mW}_V=\mU_r\mSigma_r^{1/2}, && \ \bar{\mW}_O=\mSigma_r^{1/2}\mV_r^\top,
    \end{aligned}
    \qquad \text{s.t.} \qquad
    \begin{aligned}
        &\bar{\mW}_Q, \bar{\mW}_K \in \mathbb{R}^{d\times r},\\
        &\bar{\mW}_V \in \mathbb{R}^{d \times r}, \ \bar{\mW}_O \in \mathbb{R}^{r \times d},
    \end{aligned}
\end{equation}
where $\tau = \sqrt{r}$ is the attention scaling. With this choice of parameters for the merged model, we have $\bar{\mW}_Q \bar{\mW}_K^\top/\tau=\bm{\theta}_{\mathrm{merge},QK}^r$ and $\bar{\mW}_V \bar{\mW}_O = \bm{\theta}_{\mathrm{merge},VO}^r$, where the rank of the merged head operators is $r$. Setting $r=d_k$ preserves the architecture, whereas $r>d_k$ increases the rank of each head. Naturally, adding more parameters to the model increases the computational cost at inference. Empirically, we find that moderately increasing the rank improves performance while incurring only a small computational overhead. Further discussion about rank expansion in \S\ref{app:rank_details}.

\noindent\textbf{Additional remarks.} We conclude this section with two remarks:
\textit{(i)}
RegMean~\citep{jin2023dataless} regularizes each linear-layer weight through Gram-matrix shrinkage. However, in attention, similar projection matrices do not necessarily imply similar computations. For example, changes to the query and key matrices can cancel each other and leave the QK operator unchanged, even though a penalty on the two matrices would still be nonzero.
We build $\mLambda_t$ from calibration states of the partially merged model. The resulting metric, $\mLambda_t\otimes\mLambda_t$, weights differences in the QK operator along both inputs to the attention-score map. Unlike separate penalties on $\mW_Q$ and $\mW_K$, it constrains the computation they define together while keeping the objective quadratic in the operator variable.
Thus, SLOA keeps the local operators close to their task-specific counterparts without forcing their internal attention factors to use the same coordinates.
\textit{(ii)}
ProDistill~\citep{xu2025scalable} progressively matches block outputs
on paired partially merged and task-model activations by optimizing
element-wise task-vector coefficients.
SLOA shares this progressive matching principle, but changes both
the merging variables and the level at which behavior is matched.
Specifically, it aligns composed QK and VO operators through
pre-softmax score matching and attention-weighted output matching,
respectively.
For fixed local inputs, these behavior maps are linear, yielding quadratic alignment objectives that can be solved with structured linear algebra.
Merely replacing projection factors with composed operators in a
block-level distillation objective would not yield this property.

Importantly, the main novelty of SLOA over existing methods is that SLOA treats merging as a sequence of operator-level regression problems. At each step, it matches local behavior, regularizes the operators using the data, and reconstructs them at a chosen rank. As it works directly with the composed QK and VO operators, it does not depend on how these operators are factorized. It also lets us choose the rank used by the deployed model, giving the merged model greater capacity.

\section{Experiments}\label{sec:exp}
\begin{figure}[t]
\centering
\includegraphics[width=\linewidth, trim=13cm 0 13cm 15cm, clip]{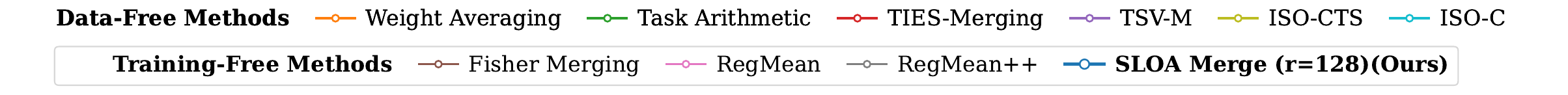}
\begin{subfigure}[t]{0.25\textwidth}
    \includegraphics[width=\linewidth, trim=0 2.4cm 50cm 0, clip]{figures/radar_all_four.pdf}
    \caption{ViT-B/32 vs.\ data-free}
    \label{subfig:fig1}
\end{subfigure}%
\begin{subfigure}[t]{0.25\textwidth}
    \includegraphics[width=\linewidth, trim=17cm 2.4cm 33cm 0, clip]{figures/radar_all_four.pdf}
    \caption{ViT-B/32 vs.\ train-free}
    \label{subfig:fig2}
\end{subfigure}%
\begin{subfigure}[t]{0.25\textwidth}
    \includegraphics[width=\linewidth, trim=34cm 2.4cm 16cm 0, clip]{figures/radar_all_four.pdf}%
    \caption{RoBERTa vs.\ data-free}
    \label{subfig:fig3}
\end{subfigure}%
\begin{subfigure}[t]{0.25\textwidth}
    \includegraphics[width=\linewidth, trim=50cm 2.4cm 0 0, clip]{figures/radar_all_four.pdf}
    \caption{RoBERTa vs. train-free}
    \label{subfig:fig4}
\end{subfigure}%
\caption{Per-task comparison of SLOA ($r{=}128$) with data-free and training-free baselines on ViT-B/32 with 20 vision tasks, Panels~\protect\subref{subfig:fig1},\,\protect\subref{subfig:fig2} and RoBERTa with 7 GLUE tasks, Panels~\protect\subref{subfig:fig3},\,\protect\subref{subfig:fig4}. SLOA consistently outperforms both data-free and training-free baselines.}
\label{fig:four_subfigures}
    \vspace{-1 em}
\end{figure}
In this section, we demonstrate SLOA for model merging across vision, natural language understanding, and large language model settings, using both fully fine-tuned and LoRA-adapted experts.
In \S\ref{sec:Merging ViT models}, we evaluate Vision Transformers~\citep{dosovitskiy2020image} on image classification benchmarks comprising 8, 14, and 20 tasks, focusing on the ViT-B/32 architecture.
In \S\ref{sec:Merging RoBERTa}, we consider merging RoBERTa models~\citep{liu2019robertarobustlyoptimizedbert} for natural language understanding (NLU).
Finally, in \S\ref{sec:Merging LLM LoRA}, we evaluate SLOA at a larger model scale by merging LoRA-adapted LLaMA-2-7B experts~\citep{touvron2023llama2openfoundation} trained for math, code, and instruction-following tasks. Fig.~\ref{fig:four_subfigures} previews the per-task gains of SLOA over all baselines.
Unless otherwise stated, all training-free methods that require calibration, including SLOA, use $N{=}256$ samples per task drawn exclusively from the corresponding training split; no test data is used for merging. All results are averaged over three random seeds.
Additional experiments in~\S\ref{sec:ablation}, and full experimental details are provided in~\S\ref{app:Experimental details}.

\subsection{Merging Vision Transformers (ViT-B)}
\label{sec:Merging ViT models}
We first evaluate SLOA on multi-task merging of ViT-B/32 experts using the task-specific checkpoints provided by~\citet{tang2024fusionbench}. All experts are fine-tuned from a shared CLIP initialization~\citep{radford2021learningtransferablevisualmodels}, each on a distinct image classification task. We consider three increasingly challenging settings involving 8, 14, and 20 experts.
Our data-free baselines are weight averaging~\citep{wortsman2022model}, Task Arithmetic~\citep{ilharco2023editing}, TIES-Merging~\citep{yadav2023ties}, TSV-M~\citep{gargiulo2025task}, and Iso-C and Iso-CTS~\citep{marczak2025no}. We also compare against training-free methods that use calibration data: Fisher Merging~\citep{matena2022merging}, RegMean~\citep{jin2023dataless}, and RegMean++~\citep{nguyen2026regmean}. \looseness=-1

Table~\ref{tab:vision_main_results_b32} summarizes the results. At the original rank, SLOA outperforms all baselines across the three settings. Its advantage over the strongest baseline, RegMean++, increases with the number of merged tasks, from $0.6$ percentage points for 8 tasks to $1.9$ for 14 tasks and $5.5$ for 20 tasks. These results demonstrate that SLOA handles an increasing number of tasks more effectively than the competing methods on these benchmarks.

\begin{wraptable}[18]{r}{85mm}%
\vspace{-\baselineskip}%
\caption{Performance comparison of model merging methods on ViT-B/32 across 8, 14, and 20 vision classification tasks. For SLOA, $r$ denotes the retained factorization rank used for both QK and VO local operators. The setting $r=64$ matches the original ViT-B/32 attention-head dimension.}%
\resizebox{0.52\textwidth}{!}{
 \begin{tabular}{clcccc}
\toprule
\multirow{2}{*}{}
& \multirow{2}{*}{Method}
& \multirow{2}{*}{\# Samples}
& \multicolumn{3}{c}{Avg. Accuracy $\uparrow$}  \\
\cmidrule(lr){4-6}
& & & 8 Tasks & 14 Tasks & 20 Tasks \\
\midrule
& Fine-tuned            & --  & 90.39 & 89.26 & 89.78 \\
\midrule
\multirow{6}{*}{\rotatebox[origin=c]{90}{Data-Free}}
& Weight Averaging       & --  & 66.29 & 65.37 & 61.08 \\
& Task Arithmetic        & --  & 69.49 & 66.47 & 60.60 \\
& TIES-Merging           & --  & 71.70 & 67.58 & 62.76 \\
& TSV-M                  & --  & 83.04 & 79.11 & 73.66 \\
& Iso-C                  & --  & 82.64 & 77.94 & 71.16 \\
& Iso-CTS                & --  & 82.87 & 79.24 & 73.41 \\
\midrule
\multirow{5}{*}{\rotatebox[origin=c]{90}{Training-Free}}
& Fisher Merging         & 256 & 69.12 & 66.26 & 61.74 \\
& RegMean                & 256 & 82.63 & 77.93 & 72.20 \\
& RegMean++              & 256 & 84.37 & 80.06 & 74.51 \\
& \cellcolor{lightpurple}SLOA ($r{=}64$) & \cellcolor{lightpurple}256 & \cellcolor{lightpurple}\underline{84.89} & \cellcolor{lightpurple}\underline{81.94} & \cellcolor{lightpurple}\underline{80.06} \\
& \cellcolor{lightblue}SLOA ($r{=}128$) & \cellcolor{lightblue}256 & \cellcolor{lightblue}\textbf{85.50} & \cellcolor{lightblue}\textbf{82.61} & \cellcolor{lightblue}\textbf{81.03} \\
\bottomrule
 \end{tabular}
}
 \label{tab:vision_main_results_b32}
\end{wraptable}
Our component ablations suggest that sequential alignment accounts for most of the improvement by compensating for upstream merging errors that subsequent operators would otherwise inherit (See~\S\ref{sec:error_accumulation} and \S\ref{app:component_ablation}). Merging the composed QK and VO operators yields a further, consistent gain over sequential alignment alone. The operator formulation also enables rank expansion. Increasing the retained rank from $r{=}64$, the original head dimension, to $r{=}128$ improves accuracy across all three settings, with larger gains when merging more tasks. This suggests that model capacity increasingly constrains merging performance as the number of tasks grows.
We use a fixed calibration budget of $N{=}256$ and report retained ranks of $r{=}64$ and $r{=}128$ to provide a controlled comparison and assess the benefit of rank expansion. These choices do not represent limits of SLOA: larger calibration sets and retained ranks yield further improvements, at the cost of additional merging time and, for rank expansion, additional inference time (\S\ref{sec:scaling_data_rank} and~\S\ref{sec:compute}). We observe similar trends for ViT-B/16 (\S\ref{appendix:sloa_vitb16}) and LoRA fine-tuned models (\S\ref{sec:vit_lora}).

\subsection{Merging encoder-based language models}
\label{sec:Merging RoBERTa}
\begin{wraptable}{r}{50mm}
\vspace{-\baselineskip}%
\caption{Avg.~GLUE metric for merged RoBERTa base models. Full per-task results are in Table~\ref{tab:roberta}.}
\label{tab:roberta_avg}
\vspace{-2.5 mm}
\resizebox{0.30\textwidth}{!}{
\begin{tabular}{lc}
\toprule
Method & Avg.\ Metric $\uparrow$ \\
\midrule
Individual (oracle) & 0.8483 \\
\midrule
\multicolumn{2}{l}{\textit{Data-Free}} \\
Weight Averaging & 0.6165 \\
Task Arithmetic & 0.6724 \\
TIES-Merging & 0.6448 \\
TSV-M & 0.7120 \\
Iso-C & 0.7115 \\
Iso-CTS & 0.6174 \\
\midrule
\multicolumn{2}{l}{\textit{Training-Free}} \\
Fisher Merging & 0.5680 \\
RegMean & 0.7066 \\
RegMean++ & 0.7139 \\
\rowcolor{lightpurple} SLOA ($r{=}64$) & \underline{0.7906} \\
\rowcolor{lightblue} SLOA ($r{=}128$) & \textbf{0.7979} \\
\bottomrule
\end{tabular}
}
\vspace{-3mm}
\end{wraptable}

To assess whether operator alignment extends beyond vision, we merge seven RoBERTa-base models fine-tuned on distinct GLUE tasks~\citep{wang2018glue}, using the checkpoints provided by~\citet{yu2024language}. These tasks cover linguistic acceptability (CoLA), sentiment classification (SST-2), paraphrase and question-pair matching (MRPC, QQP), and textual entailment (MNLI, QNLI, RTE), requiring the merged model to preserve task-specific computations across diverse input structures and decision boundaries. We compare SLOA against the same baselines used in the ViT experiment.

As shown in Table~\ref{tab:roberta_avg}, SLOA at the original rank ($r{=}64$) improves the average score over the strongest baseline by $7.7$ points. Increasing the QK/VO rank to $r{=}128$ improves scores on every task and raises this margin to $8.4$ points. The gains extend across the benchmark: SLOA achieves the highest scores on six of the seven tasks (CoLA, SST-2, MRPC, MNLI, QNLI, and RTE), trailing only TIES-Merging and TSV-M on QQP (per-task results in Table~\ref{tab:roberta}). These findings are consistent with the motivation for aligning composed attention operators. The coupled QK and VO operators govern token interactions and value transformations within each attention head; aligning them can help preserve the distinct attention patterns learned by experts fine-tuned on different language tasks.

\subsection{Merging LLaMA-2-7B LoRA experts}
\label{sec:Merging LLM LoRA}
\begin{table}[t]
\centering
\caption{LLaMA-2-7B LoRA-expert merge results (math, code, instruction-following). Avg. is the mean score over GSM8K (math), HumanEval (code, pass@1), and IFEval (instruction-following).  Bold/underline marks the best/second-best merging method per column.}
\vspace{-2mm}
\label{tab:llm_main}
\resizebox{0.76\textwidth}{!}{
\begin{tabular}{clcccc}
\toprule
\multirow{2}{*}{}
& \multirow{2}{*}{Method}
& \multicolumn{3}{c}{Benchmark}
& \multicolumn{1}{c}{\cellcolor{avggray}\textbf{Average}} \\
\cmidrule(lr){3-5}\cmidrule(lr){6-6}
& & GSM8K & HumanEval & IFEval
& \cellcolor{avggray}\textbf{Avg. $\uparrow$} \\
\midrule
\multirow{4}{*}{\rotatebox[origin=c]{90}{Experts}}
& Llama2-7B Pretrained & 4.32 & 0.00 & 20.89 & 8.40 \\
& Llama2-7B IF-Tulu & 12.59 & 1.22 & \cellcolor{avggray}66.17 & 26.66 \\
& Llama2-7B Code (Magic-Coder) & 9.70 & \cellcolor{avggray}32.93 & 26.99 & 23.21 \\
& Llama2-7B Math (Meta-Math) & \cellcolor{avggray}65.50 & 0.00 & 15.53 & 27.01 \\
\midrule
\multirow{7}{*}{\rotatebox[origin=c]{90}{Data-Free}}
& Weight Averaging & 40.11 & 18.29 & 40.67 & 33.02 \\
& Task Arithmetic ($\alpha=0.5$) & 58.98 & 15.85 & 42.70 & 39.18 \\
& TSV-M ($\alpha=0.5$) & 58.00 & 18.29 & 42.14 & 39.48 \\
& TIES ($\alpha=0.3$) & 48.14 & 20.73 & 34.57 & 34.48 \\
& KnOTS-TIES ($\lambda=0.85$) & 55.27 & 15.24 & 40.85 & 37.12 \\
& Iso-C  & 38.89 & 21.34 & 57.67 & 39.30 \\
& Iso-CTS  & 36.09 & 21.34 & 55.45 & 37.63 \\
\midrule
\multirow{5}{*}{\rotatebox[origin=c]{90}{Training-Free}}
& RegMean & 56.03 & \textbf{29.27} & 48.24 & 44.51 \\
& RegMean++ & 55.95 & \underline{29.27} & 54.16 & 46.46 \\
& \cellcolor{lightblue}SLOA-VO $r_{ov}{=}128$ (Ours) & \cellcolor{lightblue}58.61 & \cellcolor{lightblue}28.05 & \cellcolor{lightblue}57.86 & \cellcolor{lightblue}48.17 \\
& \cellcolor{lightpurple}SLOA-VO $r_{ov}{=}256$ (Ours) & \cellcolor{lightpurple}\underline{59.51} & \cellcolor{lightpurple}27.44 & \cellcolor{lightpurple}\underline{58.96} & \cellcolor{lightpurple}\underline{48.64} \\
& \cellcolor{lightyellow}SLOA-VO $r_{ov}{=}384$ (Ours) & \cellcolor{lightyellow}\textbf{59.51} & \cellcolor{lightyellow}27.44 & \cellcolor{lightyellow}\textbf{60.63} & \cellcolor{lightyellow}\textbf{49.19} \\
\bottomrule
\end{tabular}
}
\end{table}

SLOA applies directly to LoRA-adapted experts without modifying its local objective. For each task~$t$, we first form the effective fine-tuned weight $\bm{W}_t = \bm{W}_0 + \bm{B}_t \bm{A}_t$ and then construct the same functional operators used for fully fine-tuned experts. This approach also avoids the factorization ambiguity of LoRA: since $\bm{B}_t \bm{A}_t = (\bm{B}_t \bm{R})(\bm{R}^{-1} \bm{A}_t)$ for any invertible $\bm{R}$, merging $\bm{A}_t$ and $\bm{B}_t$ separately can make the result depend on an arbitrary choice of low-rank coordinates. By operating on the effective weights and their composed operators, SLOA is invariant to this ambiguity.

We evaluate this setting by merging three LoRA-adapted LLaMA-2-7B experts specialized in math, code, and instruction-following (details in \S\ref{sec:LLM_datasets}). LLaMA-2 is a decoder-only architecture with rotary positional embeddings, whose position-dependent rotations complicate the representation of the QK computation as a single composed operator. We therefore compose only the VO circuit (SLOA-VO), while merging $\mW_Q$ and $\mW_K$ individually as linear operators. We evaluate both the original VO rank, $r_{ov}{=}128$, and expanded ranks, $r_{ov} \! \in \!\{256,384\}$. Our comparisons include the baselines from \S\ref{sec:Merging ViT models} and the LoRA-specific baseline KnOTS-TIES~\citep{stoica2025knots,yadav2023ties}. We evaluate math, code, and instruction-following performance on GSM8K~\citep{cobbe2021gsm8k}, HumanEval~\citep{chen2021evaluating}, and IFEval~\citep{zhou2023instructionfollowingevaluationlargelanguage}, respectively. \looseness=-1

At the architecture-preserving rank $r_{ov}{=}128$, SLOA-VO achieves the highest average score among all merging methods, exceeding RegMean++ by $1.7$ points. Increasing $r_{ov}$ to $384$ further improves GSM8K and IFEval, with a small reduction in HumanEval performance ($27.44$ versus $28.05$ at $r_{ov}{=}128$), indicating that the benefits of additional VO capacity vary across tasks. These results extend sequential operator alignment to LoRA experts on a 7B-parameter decoder-only model. Complementary experiments on LoRA-adapted ViT-B/32 experts, using both composed QK and VO operators, are presented in \S\ref{sec:vit_lora}.

\section{Related work}
\label{sec:related_work}

\noindent\textbf{Data-free parameter-space merging.}
A broad class of data-free model merging methods considers task-specific models directly in parameter space. Uniform model soups average weights across models with the same architecture~\citep{wortsman2022model}, while Task Arithmetic merges task vectors, defined as differences between fine-tuned and pretrained weights~\citep{ilharco2023editing}. Later methods reduce task interference: TIES-Merging removes small-magnitude updates and resolves sign conflicts~\citep{yadav2023ties}, DARE randomly drops and rescales task-vector entries~\citep{yu2024language}, and spectral methods such as TSV-M and Iso-C/Iso-CTS manipulate the singular structure of task vectors to denoise or rescale merged updates~\citep{gargiulo2025task,marczak2025no}.
While attractive since they require no data or optimization, these methods merge parameter tensors instead of the computation units they implement. This is limiting in transformers, where attention depends on composed QK and OV operators and not isolated query, key, value, and output layers. SLOA instead merges these functional operators, removing dependence on arbitrary parameterizations.

\noindent\textbf{Calibration-based training-free merging.}
Another line of work uses a small calibration set while still avoiding gradient-based training. Fisher Merging~\citep{matena2022merging} weights parameters according to their Fisher information, preserving parameters estimated to be important for the task losses. RegMean~\citep{jin2023dataless} solves a closed-form regression problem that aligns hidden representations of the merged model with those of the task-specific models. RegMean++~\citep{nguyen2026regmean} improves this idea by computing statistics across layers, reducing the covariate shift caused by earlier merging errors.
SLOA belongs to this training-free regime, but differs in the object being aligned. Prior methods mainly treat layers as independent units. In contrast, SLOA defines a local operator for each \emph{functional} component: LayerNorm affine maps, linear maps, QK operators, and OV operators. This allows the same closed-form alignment principle to apply to both standard linear layers and coupled attention computations. Moreover, SLOA performs alignment \emph{sequentially} under the activations produced by the partially merged model, so later operators are merged on the activations they will actually receive at inference time.

\noindent\textbf{Test-time adaptation and training for merging.}
A separate class of methods improves merging by introducing learned coefficients, routers, task-specific components, or distillation objectives.
AdaMerging~\citep{yang2024adamerging} learns layer-wise merging coefficients using unlabeled test samples. Twin-Merging~\citep{lu2024twinmerging} decomposes task vectors into shared and private components and uses a router to select fusion weights. EMR-Merging~\citep{huang2024emrmerging} maintains both shared and task-specific parameters, requiring task identity at inference time. Representation Surgery~\citep{yang2024representation} trains an auxiliary module to align representations, while ProDistill~\citep{xu2025scalable} formulates merging as a distillation problem with learned merging coefficients.
These methods can achieve strong performance, but they require additional training, test-time data, task identity, routing, or teacher-student optimization. SLOA avoids these requirements: after collecting calibration activations, each local merge is solved in closed form, produces a single deployed model, and requires neither task identity nor test-time adaptation at inference time.

\section{Conclusion}
\label{sec:conclusion_and_limitations}
We presented \emph{Sequential Local Operator Alignment} (SLOA), a training-free framework for merging transformers by aligning \emph{local functional operators}. By treating the QK and VO products as the effective operators, matching each operator's behavior under the partially merged model's activations with a data-aware operator geometry, and merging sequentially in topological order, SLOA preserves task-specific behavior while reducing the growth of activation error across depth. The operator-level view also decouples the merged attention operator from its final factorization, making head rank an explicit capacity-control parameter.
We evaluate SLOA on ViT-B/32 and ViT-B/16 with up to 20 tasks,
RoBERTa on GLUE, LoRA-adapted ViT experts, and a VO-only variant for LoRA-adapted LLaMA-2-7B.
Across these settings, SLOA outperforms strong merging baselines at the original head rank, showing that its gains do not require widening the deployed model.
Its relative advantage becomes more pronounced as the number of merged tasks grows.
Increasing the head rank can provide further gains beyond those obtained at the original rank.

One limitation of SLOA is that, like other calibration-based training-free methods, it requires a small unlabeled calibration set that represents each task's input distribution. When such data are unavailable, data-free methods may be more suitable.
In addition, SLOA follows a greedy sequential procedure. At each step, it optimizes a well-defined local objective,
but it does not optimize a single end-to-end objective.
Finally, SLOA requires more computation during merging than data-free methods and simpler calibration-based approaches because it propagates activations, solves local regression problems, and factorizes the merged operators. Increasing the rank beyond $d_k$ also widens the attention heads, which increases inference time and memory use.
Promising directions include reducing reliance on in-distribution calibration data through synthetic, generated, or proxy inputs, and combining SLOA with data free merging when calibration data provide insufficient coverage of some tasks. Extending the composed QK formulation to attention with relative or rotary position encodings would also broaden its architectural scope.
Another direction is automatic per-head rank selection, allowing the merged model's capacity to be allocated according to deployment constraints.
More broadly, the proposed method supports a functional view of transformer merging: align local computations along the merged execution path, and choose the attention factorization to control how much of the aligned operators the deployed model retains for multi-task merging.
In light of this view, the unit of merging does not need to be the unit of parameterization.

\section*{Acknowledgments}
AM, SUS, and RM are grateful for funding from the European Research Council (ERC) under the Horizon Europe Framework Program (HORIZON) for proposal number 101170430 CollectiveMinds. Views and opinions expressed are, however, those of the authors only and do not necessarily reflect those of the European Union or the European Research Council. Neither the European Union nor the granting authority can be held responsible for them.
YEL acknowledges support from the Schmidt Futures Israeli Women’s Postdoctoral Award and the Viterbi Fellowship from the Faculty of Electrical and Computer Engineering at the Technion. YEL and SJ are supported by an Alexander von Humboldt Professorship.

\bibliographystyle{plainnat}
\bibliography{references}

\newpage
\appendix
\begin{center}
    \LARGE\textbf{Appendix}
\end{center}

\startcontents[appendices]

\printcontents[appendices]{l}{1}{
    \setcounter{tocdepth}{2}
}
\newpage

\section{Additional background}
\subsection{Model Merging}
\label{sec:model_merging}

Let there be a set of $T$ distinct tasks with corresponding training datasets $\mathcal{D} = \{D_1, \ldots, D_T\}$. We assume a homogeneous neural network architecture parameterized by $\mathbf{\Phi} \in \mathbb{R}^d$, where $d$ denotes the total number of parameters. Let $\mathbf{\Phi}_{\text{pre}} \in \mathbb{R}^d$ denote the parameters of a large-scale pretrained foundation model (e.g., CLIP~\citep{radford2021learningtransferablevisualmodels}) that serves as a shared initialization. We are given $T$ task-specific models $\{\mathbf{\Phi}_1, \ldots, \mathbf{\Phi}_T\}$, where each model $\mathbf{\Phi}_t$ is obtained by independently fine-tuning $\mathbf{\Phi}_{\text{pre}}$ exclusively on its respective dataset $D_t$.

\textbf{Model Merging.} Model merging is the process of aggregating these distinct parameter sets into a single, unified multi-task model without requiring \emph{computationally expensive joint retraining on the original training datasets}. Mathematically, we seek an aggregation function $g$ such that:
\begin{equation}
    \mathbf{\Phi}_{\text{merge}} = g(\mathbf{\Phi}_1, \ldots, \mathbf{\Phi}_T).
    \label{eq:def_model_merging}
\end{equation}

\textbf{Performance Objectives.} The fundamental objective of model merging is to construct a single parameter set $\mathbf{\Phi}_{\text{merge}}$ that encapsulates the combined capabilities of all individual experts. Let $\mathcal{M}_t: \mathbb{R}^d \to \mathbb{R}$ denote a task-specific evaluation metric (e.g., accuracy) for task $t$, where higher values indicate better performance. A successful merging function ensures that the unified model recovers the performance of the specialized experts:
\begin{equation}
    \mathcal{M}_t(\mathbf{\Phi}_{\text{merge}}) \approx \mathcal{M}_t(\mathbf{\Phi}_t), \quad \forall t \in \{1, \ldots, T\}.
\end{equation}

\subsection{Transformer architecture and the attention operation}
\label{sec:preliminaries_transformer}

The Transformer architecture \citep{vaswani2023attentionneed} has become the foundational backbone for state-of-the-art models across both vision and language domains. Structurally, a Transformer is composed of multiple identical, sequentially stacked layers  For a given layer, the input representations are processed through two primary sub-modules: a (i) \textbf{Multi-Head Attention (MHA}) block and a (ii) \textbf{Multi-Layer Perceptron (MLP)}, accompanied by Layer Normalization and residual connections.

\textbf{Multi-Head Attention (MHA).}
The core of the Transformer is the attention mechanism, which routes information across sequence tokens. Let $\mX \in \mathbb{R}^{L \times d}$ denote the input sequence of $L$ sequence length with model dimension $d$. The matrices for the layer are defined as ${\mW_Q}, {\mW_K}, {\mW_V}, {\mW_O} \in \mathbb{R}^{d \times d}$.

To allow the model to jointly attend to information from different representations
subspaces, the MHA operation divides the global matrices into $H$ independent heads.
The global parameter matrices $\mW_Q, \mW_K, \mW_V, \mW_O \in \mathbb{R}^{d \times d}$
are partitioned across $H$ heads, where each head $j$ operates on a reduced subspace
of dimension $d_k = d / H$, with head-specific sub-matrices
$\mW_Q^{(j)}, \mW_K^{(j)}, \mW_V^{(j)} \in \mathbb{R}^{d \times d_k}$ and
$\mW_O^{(j)} \in \mathbb{R}^{d_k \times d}$ for the $j$-th head. Crucially, instead of viewing MHA merely as a concatenation operation, it can be mathematically formulated as the sum of independent head contributions added directly to the residual stream. The output $\mZ_{\mathrm{MHA}} \in \mathbb{R}^{n \times d}$ of the MHA block is given by:
\begin{equation}
    \mZ_{\mathrm{MHA}} = \sum_{j=1}^{H} \mathrm{softmax}\!\left( \frac{(\mX \mW_Q^{(j)})(\mX \mW_K^{(j)})^\top}{\sqrt{d_k}} \right) \mX \mW_V^{(j)} \mW_O^{(j)}
    \label{eq:mha_sum}
\end{equation}
This formulation highlights two distinct functional circuits within each head $j$: the bilinear Query-Key (QK) circuit defined by $\mW_Q^{(j)} {\mW_K^{(j)}}^\top$, which determines attention scores, and the linear Value-Output (VO) circuit defined by $\mW_V^{(j)} \mW_O^{(j)}$, which dictates the information written to the output~\citep{elhage2021mathematical}.

\textbf{Multi-Layer Perceptron (MLP).}
Following the attention mechanism, the representations pass through a two-layer MLP block. Let $\mW_1 \in \mathbb{R}^{d \times d_{out}}$ and $\mW_2 \in \mathbb{R}^{d_{out} \times d}$ be the first and the second MLP layer matrices, respectively. For an input representation to the MLP block, which we denote as $\mZ \in \mathbb{R}^{L \times d}$ (functionally the layer-normalized sum of $\mX$ and $\mZ_{\mathrm{MHA}}$), the block applies a non-linear activation function $\sigma(\cdot)$ (e.g., ReLU or GELU). The transformation sequence maps the inputs to the final block output $\mZ_{\mathrm{MLP}} \in \mathbb{R}^{L \times d}$:
Algebraically, the operation of the entire MLP block is summarized as:
\begin{equation}
    \mZ_{\mathrm{MLP}} = \sigma(\mZ \mW_1) \mW_2
\end{equation}

\section{Extended related work}
\label{extended_related_works}
\subsection{Alignment under model symmetries and activation matching}
Neural networks contain many parameter symmetries: different parameterizations can represent the same function~\citep{dinh2017sharp, neyshabur2016path, zhao2025symmetry, bona2023parameter}. Prior work on model alignment and mode connectivity studies how to account for such symmetries before merging, for example by matching neurons, features, or permutations across independently trained networks~\citep{ainsworth2023git, singh2020model, xu2024training, zhang2025beyond}. These methods highlight an important principle: successful merging should compare models in a functionally meaningful coordinate system, rather than relying only on raw parameter coordinates.

Our work is motivated by the same non-identifiability issue, but addresses a different symmetry that appears inside transformer attention.
For a single attention head, the transformation
$
(\mW_Q,\mW_K)\mapsto(\mW_Q\mR,\mW_K\mR^{-\top})
$
leaves the QK operator $\mW_Q\mW_K^\top$ unchanged, and analogous changes of the value-head coordinates preserve the VO operator.
Thus, aligning or averaging $\mW_Q$, $\mW_K$, $\mW_V$, and $\mW_O$ separately can break a computation that is shared at the operator level.
Instead of first searching for a coordinate transformation between factorizations, SLOA removes this ambiguity by merging the invariant local operators themselves and factorizing the merged operators only after alignment.
\subsection{Transformer circuits and operator structure}
Several analyses of transformers emphasize that attention heads are naturally described through composed circuits rather than isolated projection matrices~\citep{elhage2021mathematical,dar2023analyzing,vaswani2023attentionneed, theus2026generalized}.
In this view, the QK circuit controls the pre-softmax attention scores, while the VO, or OV, circuit controls how attended token representations are written back to the residual stream.
This perspective has been influential for understanding transformer computation, but it has been less directly exploited in training-free model merging.
SLOA turns this circuit-level view into a concrete merging algorithm.
For each attention head, it constructs and aligns the QK and VO operators as residual-space maps.
This also separates the operator being merged from the rank used to implement it.
Each task-specific head realizes a low-rank operator because the head dimension is limited, but the merged multi-task operator can have higher effective rank when different tasks use different subspaces.
SLOA therefore treats rank as a capacity-control mechanism: after solving for the merged ambient operator, it can either factorize it at the original head rank or retain additional singular directions through rank expansion.
This differs from spectral task-vector methods, which apply SVD to task updates in weight space; in SLOA, the SVD is applied to the merged functional attention operator and determines how much of that operator is represented in the final head.

\section{Technical comparison to RegMean, RegMean++, rotation-symmetry matching, and  ProDistill}
\label{app:tech_compare_all}

\subsection{Comparison with RegMean}
\label{app:comparison_regmean}

We compare SLOA with Regression Mean (RegMean)~\citep{jin2023dataless}.
RegMean is one of the closest predecessors to our method because it also derives a closed-form merging rule from a regression objective.
However, RegMean and SLOA differ in the object being merged, the local objective being solved, the treatment of attention, and the role of calibration states.

\paragraph{RegMean as layer-wise linear regression.}
Consider a linear layer whose local behavior in task model $t$ is
\begin{equation}
    \mX_t \mW_t ,
\end{equation}
where $\mX_t$ denotes the input activations to that linear layer and $\mW_t$ is the corresponding task-specific weight matrix.
RegMean seeks a merged weight $\mW$ whose outputs match the outputs of the individual task layers on their own layer inputs:
\begin{equation}
    \min_{\mW}
    \sum_{t=1}^{T}
    \left\|
        \mX_t \mW
        -
        \mX_t \mW_t
    \right\|_F^2 .
    \label{eq:regmean_objective}
\end{equation}
Let
\begin{equation}
    \mG_t = \mX_t^\top \mX_t
\end{equation}
be the input Gram matrix for task model $t$.
The closed-form RegMean solution is
\begin{equation}
    \mW_{\mathrm{RegMean}}
    =
    \left(
        \sum_{t=1}^{T}
        \mG_t
    \right)^{-1}
    \left(
        \sum_{t=1}^{T}
        \mG_t \mW_t
    \right),
    \label{eq:regmean_solution}
\end{equation}
up to the normalization convention used for the empirical average.
In transformer models, RegMean applies this rule independently to each linear layer.
Other parameters, such as embeddings, biases, and normalization parameters, are typically merged by simple averaging.

\paragraph{Connection to SLOA for ordinary linear maps.}
SLOA reduces to a RegMean-like objective in a restricted special case.
For an ordinary one-sided linear map, SLOA uses the local behavior map
\begin{equation}
    \mathcal{A}(\mX;\mW) = \mX\mW .
\end{equation}
If we ignore the sequential activation mismatch by setting
\begin{equation}
    \widehat{\mX}_{t,i} = \mX_{t,i}
\end{equation}
for all calibration examples, and if we remove the additional operator-geometry regularizer, then the SLOA activation-matching term becomes
\begin{equation}
    \min_{\mW}
    \sum_{t=1}^{T}
    \sum_{i=1}^{n_t}
    \left\|
        \mX_{t,i}\mW
        -
        \mX_{t,i}\mW_t
    \right\|_F^2 ,
\end{equation}
which is exactly the RegMean regression problem for linear layers.
Thus, RegMean can be viewed as a special case of local behavior matching for isolated one-sided linear operators under task-model activations.

The purpose of SLOA is to go beyond this special case.
SLOA changes both the local inputs and the operator space:
\begin{equation}
    \mathcal{A}
    \left(
        \widehat{\mZ}_{t,i};
        \bm{\theta}
    \right)
    \approx
    \mathcal{A}
    \left(
        \mZ_{t,i};
        \bm{\theta}_t
    \right),
    \label{eq:sloa_vs_regmean_behavior}
\end{equation}
where $\widehat{\mZ}_{t,i}$ is produced by the partially merged model and $\mZ_{t,i}$ is produced by the original task model.
For linear maps, this gives an objective of the form
\begin{equation}
    \min_{\mW}
    \sum_{t=1}^{T}
    \left\|
        \widehat{\mX}_t \mW
        -
        \mX_t \mW_t
    \right\|_F^2
    +
    \sum_{t=1}^{T}
    \left\|
        \mW-\mW_t
    \right\|_{\widehat{\chi}_t}^{2}.
    \label{eq:sloa_linear_regmean_comparison}
\end{equation}
The corresponding normal equations involve both the partially merged Gram matrix
\begin{equation}
    \widehat{\mX}_t^\top \widehat{\mX}_t
\end{equation}
and the cross-state term
\begin{equation}
    \widehat{\mX}_t^\top \mX_t \mW_t .
\end{equation}
This is different from RegMean, whose objective uses the same task-model input $\mX_t$ on both sides of the regression.

\paragraph{Independent linear-layer merging versus local functional operators.}
The main limitation of RegMean from our perspective is that it treats each linear layer as the primitive unit of merging.
This is appropriate for genuinely one-sided linear maps, but it is not aligned with the computation inside transformer attention.
In a transformer head, the query and key weights do not determine attention scores independently.
Rather, the pre-softmax attention scores depend on the composed operator
\begin{equation}
    \bm{\theta}_{QK,t}^{(j)}
    =
    \frac{
        \mW_{Q,t}^{(j)}
        \big(\mW_{K,t}^{(j)}\big)^\top
    }{\sqrt{d_k}} .
\end{equation}
Similarly, the value and output weights determine the residual write map through
\begin{equation}
    \bm{\theta}_{VO,t}^{(j)}
    =
    \mW_{V,t}^{(j)}
    \mW_{O,t}^{(j)} .
\end{equation}
RegMean applies regression to the individual projection matrices, treating $\mW_Q$, $\mW_K$, $\mW_V$, and $\mW_O$ as separate linear maps.
SLOA instead treats $\bm{\theta}_{QK}$ and $\bm{\theta}_{VO}$ as the local functional operators to be merged.

This distinction matters because the individual attention factors are not identifiable from the attention computation.
For any invertible matrix $\mR$,
\begin{equation}
    \mW_Q \mW_K^\top
    =
    (\mW_Q \mR)
    (\mW_K \mR^{-\top})^\top .
\end{equation}
Thus, two models can have the same QK score operator but different query/key coordinates.
A method that merges $\mW_Q$ and $\mW_K$ independently is sensitive to this arbitrary factorization.
SLOA removes this ambiguity by merging the composed QK operator directly.
The same argument applies to the VO circuit.

\paragraph{One-sided geometry versus two-sided attention geometry.}
RegMean uses input Gram matrices for one-sided linear maps.
For a linear map $\mX\mW$, this is natural because the local discrepancy is measured after left multiplication by $\mX$.
However, the QK behavior map is not one-sided:
\begin{equation}
    \mathcal{A}(\mX;\bm{\theta}_{QK})
    =
    \mX \bm{\theta}_{QK} \mX^\top .
\end{equation}
SLOA therefore uses a two-sided metric for QK operators:
\begin{equation}
    \left\|
        \bm{\theta}
        -
        \bm{\theta}_t
    \right\|_{\widehat{\chi}_t}^{2}
    =
    \left\|
        \mLambda_t^{1/2}
        (\bm{\theta}-\bm{\theta}_t)
        \mLambda_t^{1/2}
    \right\|_F^2 ,
\end{equation}
which corresponds after vectorization to a Mahalanobis metric with covariance structure
\begin{equation}
    \mLambda_t \otimes \mLambda_t .
\end{equation}
This two-sided geometry has no analogue in RegMean because RegMean does not formulate QK as a bilinear residual-space operator.

\paragraph{Dataless statistics versus sequential calibration states.}
RegMean was designed for a dataless knowledge-fusion setting: the model provider can release the fine-tuned weights together with low-dimensional inner-product statistics, without sharing the original training data.
This is an important privacy and deployment advantage.
SLOA instead targets the training-free calibration-data setting.
It uses a small calibration set to run both the task models and the partially merged model, collecting the states required for local behavior matching and data-dependent operator regularization.
Since SLOA evaluates each local operator on the states produced by the partially merged model, it can correct for upstream merging errors.
RegMean computes each layer's merge from the task-model layer inputs and therefore implicitly assumes that previous merged layers will reproduce the task-model activations accurately.
SLOA does not make this assumption: the merged operator is evaluated on the activation distribution it will actually receive during the merged model's forward pass.

\paragraph{Rank and capacity.}
RegMean preserves the original architecture because it directly solves for each original weight matrix.
For attention, this means the head dimension remains fixed and the rank of each QK or VO product remains constrained by the original factorization.
SLOA decouples the merged operator from its final factorization.
It first solves for the merged QK or VO operator in the ambient residual-space operator space
\begin{equation}
    \Theta_{\mathrm{amb}} = \mathbb{R}^{d\times d},
\end{equation}
and only afterward factorizes the operator at a chosen rank $r$.
This allows either rank-preserving merging with $r=d_k$ or rank-expanded merging with $r>d_k$.
RegMean does not expose this rank-control mechanism because it never explicitly constructs the composed attention operator.

\subsection{Comparison with RegMean++}
\label{app:comparison_regmeanpp}

We next compare SLOA with RegMean++~\citep{nguyen2026regmean}.
RegMean++ identifies an important limitation of RegMean: independently merging each linear layer using task-model activations ignores the way representation errors propagate through the merged network.
RegMean++ addresses this by recomputing statistics progressively using activations produced by previously merged layers.
SLOA shares this sequential motivation, but differs in the local objective, the treatment of attention, the operator geometry, and the final factorization step.

\paragraph{RegMean++ as progressive RegMean.}
RegMean++ keeps the core RegMean closed-form linear-layer solve.
The key modification is how the layer statistics are collected.
In RegMean, the input statistics for a layer are obtained from the corresponding task model.
In RegMean++, the current layer is processed after earlier layers have already been merged, so the statistics for later layers are computed using activations induced by the partially merged model.
This makes the regression statistics more consistent with the actual feature distribution that the merged model will experience.

Schematically, let $\mH_{t}^{(\ell-1)}$ denote the activation entering transformer layer $\ell$ for task $t$.
RegMean uses the task-model state
\begin{equation}
    \mH_{t}^{(\ell-1)}
    =
    f_{t}^{(<\ell)}(\mX_t),
\end{equation}
whereas RegMean++ uses a merged-model state of the form
\begin{equation}
    \widehat{\mH}_{t}^{(\ell-1)}
    =
    f_{\mathrm{merge}}^{(<\ell)}(\mX_t).
\end{equation}
The resulting statistics are then used in the same RegMean-style closed-form solve for the linear layers inside the current transformer layer.
Thus, RegMean++ is progressive in the sense that earlier merged layers influence the feature statistics used to merge later layers.

\paragraph{Shared motivation with SLOA.}
SLOA agrees with the core diagnosis of RegMean++: a layer should not be merged under the assumption that all preceding layers have been merged perfectly.
If earlier merges shift the hidden states, then later layers should be aligned under the shifted hidden states that the final merged model will actually produce.
This is exactly why SLOA introduces the partially merged local input $
    \widehat{\mZ}_{t,i}^{(m)}$
for operator $o_m$.

However, SLOA uses this sequential state in a different regression problem.
At operator $o_m$, SLOA solves
\begin{equation}
    \mathcal{A}
    \left(
        \widehat{\mZ}_{t,i}^{(m)};
        \bm{\theta}
    \right)
    \approx
    \mathcal{A}
    \left(
        \mZ_{t,i}^{(m)};
        \bm{\theta}_t^{(m)}
    \right),
    \label{eq:sloa_regmeanpp_core_difference}
\end{equation}
where $\widehat{\mZ}_{t,i}^{(m)}$ is the state seen by the partially merged model and $\mZ_{t,i}^{(m)}$ is the state seen by the original task model.
Thus, the merged operator is trained to act on the actual merged-model input while matching the task-model local behavior (i.e., output).
This creates a corrective objective that explicitly bridges the shifted merged-model state and the task-model target behavior.

\paragraph{Progressive statistics versus corrective local behavior matching.}
RegMean++ modifies RegMean by making the statistics progressive.
SLOA modifies the objective itself.
For an ordinary linear map, SLOA uses
\begin{equation}
    \min_{\mW}
    \sum_{t=1}^{T}
    \left\|
        \widehat{\mX}_{t}\mW
        -
        \mX_t \mW_t
    \right\|_F^2
    +
    \sum_{t=1}^{T}
    \left\|
        \mW-\mW_t
    \right\|_{\widehat{\chi}_t}^{2}.
    \label{eq:sloa_linear_regmeanpp}
\end{equation}
The left side of the behavior-matching term uses the partially merged input $\widehat{\mX}_t$, while the target output uses the original task-model behavior $\mX_t\mW_t$.
Consequently, the closed-form solution depends not only on a merged-model Gram matrix
\begin{equation}
    \widehat{\mX}_t^\top \widehat{\mX}_t,
\end{equation}
but also on a cross-state term
\begin{equation}
    \widehat{\mX}_t^\top \mX_t \mW_t .
\end{equation}
This cross-state term directly compensates for the mismatch between the state that the merged model receives and the state under which the task-specific operator was originally used.

RegMean++ instead keeps the RegMean form of the linear-layer solve, but obtains the features progressively from the merged computation.
This improves over RegMean by incorporating cross-layer dependencies, but it still treats the current component as an isolated linear layer whose input statistics are updated.
SLOA uses the progressive state as part of a local functional matching problem between two execution paths.

\paragraph{Layer-wise progression versus operator-wise topological progression.}
RegMean++ progresses at the transformer-layer level.
Within a transformer layer, it computes statistics for linear layers and merges those linear weights using the RegMean rule.
Other parameters, including non-linear or normalization parameters, are merged by averaging.

SLOA progresses at the level of local operators in a topological order.
The local operators include LayerNorm affine maps, ordinary linear maps, QK attention operators, and VO attention operators.
This finer granularity matters inside a transformer block because the internal attention computations have distinct functional roles: QK controls token routing through attention scores,
whereas VO controls the residual write map after attention aggregation.
A layer-wise regression method can update statistics progressively, but it does not explicitly represent these two coupled attention mechanisms as separate functional operators.
SLOA does.

\paragraph{Treatment of attention.}
The most important difference is the treatment of self-attention.
RegMean++ inherits RegMean's parameterization and applies regression to linear projection matrices.
Thus, the attention projections are still treated as individual linear maps.
This means that the query, key, value, and output matrices are merged as parameter tensors rather than as the composed operators that determine attention behavior.

SLOA instead defines the QK and VO local operators as
\begin{equation}
    \bm{\theta}_{QK,t}^{(j)}
    =
    \frac{
        \mW_{Q,t}^{(j)}
        \big(\mW_{K,t}^{(j)}\big)^\top
    }{\sqrt{d_k}},
    \qquad
    \bm{\theta}_{VO,t}^{(j)}
    =
    \mW_{V,t}^{(j)}
    \mW_{O,t}^{(j)} .
\end{equation}
These are the objects that determine the pre-softmax attention scores and the residual write map, respectively.
Therefore, SLOA aligns attention at the level of local function rather than at the level of arbitrary attention-coordinate factors.

For any invertible matrix $\mR$,
\begin{equation}
    \mW_Q \mW_K^\top
    =
    (\mW_Q\mR)
    (\mW_K\mR^{-\top})^\top .
\end{equation}
Thus, two QK factorizations can be functionally identical while their individual query and key matrices differ.
RegMean++ can reduce covariate shift through progressive statistics, but it does not remove this factorization ambiguity because it still merges the individual factors.
SLOA removes the ambiguity by merging the invariant composed operator.

\paragraph{Operator geometry.}
RegMean++ uses RegMean-style Gram statistics for linear layers.
This remains a one-sided geometry appropriate for maps of the form
\begin{equation}
    \mX\mW .
\end{equation}
SLOA uses different geometries for different local operators.
For one-sided operators, such as linear maps and VO, SLOA uses a Mahalanobis geometry induced by the calibration activations:
\begin{equation}
    \left\|
        \bm{\theta}-\bm{\theta}_t
    \right\|_{\widehat{\chi}_t}^{2}
    =
    \operatorname{Tr}
    \left[
        (\bm{\theta}-\bm{\theta}_t)^\top
        \mLambda_t
        (\bm{\theta}-\bm{\theta}_t)
    \right].
\end{equation}
For QK, whose behavior is bilinear in the residual state,
\begin{equation}
    \mathcal{A}(\mX;\bm{\theta}_{QK})
    =
    \mX\bm{\theta}_{QK}\mX^\top ,
\end{equation}
SLOA uses the two-sided metric
\begin{equation}
    \left\|
        \bm{\theta}-\bm{\theta}_t
    \right\|_{\widehat{\chi}_t}^{2}
    =
    \left\|
        \mLambda_t^{1/2}
        (\bm{\theta}-\bm{\theta}_t)
        \mLambda_t^{1/2}
    \right\|_F^2 .
\end{equation}
This operator-specific geometry is absent from RegMean++, because RegMean++ does not formulate QK as a two-sided residual-space operator.

\paragraph{LayerNorm and non-linear components.}
RegMean++ improves the statistics used for linear-layer merging, but it still handles non-linear and normalization-related parameters by simple averaging.
SLOA includes LayerNorm affine parameters in the same local-operator alignment framework.
After the normalization statistics are fixed for a given input, the LayerNorm affine map is linear in $(\vgamma,\vbeta)$:
\begin{equation}
    \mathcal{A}(\mX;\vgamma,\vbeta)
    =
    \nu(\mX)\operatorname{diag}(\vgamma)
    +
    \mathbf{1}\vbeta^\top .
\end{equation}
Therefore, SLOA can align LayerNorm behavior with a closed-form local objective instead of simply averaging the affine parameters.
This is useful because LayerNorm parameters affect the scale and shift of the states that downstream operators receive.

\paragraph{Rank and capacity control.}
RegMean++ preserves the original architecture.
Since it merges existing attention projection matrices, the attention head dimension remains fixed at $d_k$.
The method therefore does not expose the spectrum of the merged QK or VO operator and cannot choose how much rank to retain after merging.

SLOA explicitly separates the merged operator from its factorization.
After solving for an ambient QK or VO operator
\begin{equation}
    \bar{\bm{\theta}}^{(j)}
    \in
    \mathbb{R}^{d\times d},
\end{equation}
SLOA factorizes it at a chosen rank $r_j$.
When $r_j=d_k$, the original head dimension is preserved.
When $r_j>d_k$, the merged model can retain additional singular directions that arise from combining task-specific low-rank operators.
This provides a capacity-control mechanism that is not available to RegMean++.

\paragraph{Different notions of ``sequential''.}
Both RegMean++ and SLOA are sequential, but the sequentiality serves different purposes.

RegMean++ is sequential because the statistics used for later layers are recomputed from activations influenced by earlier merged layers.
Its main goal is to make RegMean's input statistics more faithful to the merged model's internal feature distribution.

SLOA is sequential because each local operator is aligned under the actual execution path of the partially merged model.
The target remains the task-specific local behavior, while the input to the merged operator is the partially merged input.
Thus, each local solve is not only statistics-aware but also corrective:
\begin{equation}
    \text{merged operator on merged state}
    \quad \approx \quad
    \text{task operator on task state}.
\end{equation}
SLOA's cross-state matching explicitly accounts for that shift.

\subsection{Comparison with rotation-symmetry matching}
\label{app:comparison_rotation_symmetry}

We provide a detailed comparison with weight space rotation-symmetry matching \citep{zhang2025beyond}.
This work is closely related to ours because it also recognizes that transformer attention contains nontrivial symmetries beyond permutation symmetry.
However, the two methods use this observation in fundamentally different ways.
Rotation-symmetry matching searches for functionally equivalent parameterizations of the input models that are closer in weight space before applying a conventional model-fusion rule.
SLOA instead changes the object being merged: it constructs, aligns, and merges the local functional operators themselves.

\paragraph{Rotation-symmetry matching.}
\citep{zhang2025beyond} observe that the self-attention factors are not individually identifiable.
For a single attention head, if $\mR$ is an orthogonal matrix, then the transformation
\begin{equation}
    \mW_Q \mapsto \mW_Q \mR,
    \qquad
    \mW_K \mapsto \mW_K \mR
\end{equation}
preserves the QK score operator because
\begin{equation}
    (\mW_Q \mR)(\mW_K \mR)^\top
    =
    \mW_Q \mR \mR^\top \mW_K^\top
    =
    \mW_Q \mW_K^\top .
\end{equation}
Similarly, for the value-output path, the transformation
\begin{equation}
    \mW_V \mapsto \mW_V \mS,
    \qquad
    \mW_O \mapsto \mS^\top \mW_O ,
\end{equation}
with orthogonal $\mS$, preserves the VO operator:
\begin{equation}
    (\mW_V \mS)(\mS^\top \mW_O)
    =
    \mW_V \mW_O .
\end{equation}
Based on these invariances, rotation-symmetry matching chooses rotations that make one model's attention parameters close to another model's attention parameters in Euclidean distance.
For example, in the QK case, the matching problem can be viewed as an orthogonal Procrustes-type objective of the form
\begin{equation}
    \min_{\mR^\top \mR=\mI}
    \left\|
        \mW_{Q,1}\mR-\mW_{Q,2}
    \right\|_F^2
    +
    \left\|
        \mW_{K,1}\mR-\mW_{K,2}
    \right\|_F^2 ,
\end{equation}
with an analogous objective for the VO factors.
After this matching step, the transformed models are merged using an existing fusion method, such as weight averaging, Fisher merging, or RegMean.
Thus, rotation-symmetry matching is a \emph{preprocessing} or \emph{re-basing} step for parameter-space fusion.

\paragraph{SLOA merges operators, not matched parameter factors.}
SLOA starts from the same structural fact that QK and VO factors are not individually meaningful.
However, rather than searching for rotations that make the factors close in parameter space, SLOA eliminates the factorization ambiguity by directly forming the composed operators
\begin{equation}
    \bm{\theta}_{QK,t}^{(j)}
    =
    \frac{
        \mW_{Q,t}^{(j)}
        \big(\mW_{K,t}^{(j)}\big)^\top
    }{\sqrt{d_k}},
    \qquad
    \bm{\theta}_{VO,t}^{(j)}
    =
    \mW_{V,t}^{(j)}
    \mW_{O,t}^{(j)} .
\end{equation}
The optimization variable in SLOA is the operator $\bm{\theta}$ itself, not a rotation matrix and not the individual factors.
For each local operator, SLOA solves
\begin{equation}
    \bm{\theta}_{\mathrm{merge}}
    =
    \underset{\bm{\theta}\in\Theta_{\mathrm{amb}}}{\arg\min}
    \;
    \mathcal{L}_{\mathrm{act}}(\bm{\theta})
    +
    \frac{1}{T}
    \sum_{t=1}^{T}
    \left\|
        \bm{\theta}
        -
        \bm{\theta}_t
    \right\|_{\widehat{\chi}_t}^{2}.
\end{equation}
The merged operator is then factorized back into attention-compatible weights.
Thus, the order of operations is different and this difference is central.
Rotation matching preserves the function of each individual model before fusion, but the final fused model is still produced by merging the individual parameter factors.
SLOA instead performs the merge in the space of the functional objects that determine the local computation.

\paragraph{Why factor matching does not remove factor-level interference.}
Even after rotation matching, factor-wise merging can introduce cross terms that are not determined by the desired QK or VO operators.
Consider two QK heads, and suppose model $1$ has been rotation-matched to model $2$ using an orthogonal matrix $\mR$.
If we then average the matched factors,
\begin{equation}
    \bar{\mW}_Q
    =
    \frac{1}{2}
    \left(
        \mW_{Q,1}\mR+\mW_{Q,2}
    \right),
    \qquad
    \bar{\mW}_K
    =
    \frac{1}{2}
    \left(
        \mW_{K,1}\mR+\mW_{K,2}
    \right),
\end{equation}
then the resulting QK operator is
\begin{align}
    \bar{\mW}_Q \bar{\mW}_K^\top
    =
    \frac{1}{4}
    \Big(
        &
        \mW_{Q,1}\mW_{K,1}^\top
        +
        \mW_{Q,2}\mW_{K,2}^\top
        \nonumber
        +
        \mW_{Q,1}\mR \mW_{K,2}^\top
        +
        \mW_{Q,2}\mR^\top \mW_{K,1}^\top
    \Big).
\end{align}
The first two terms are the task-specific QK operators.
The last two terms are cross terms that depend on the particular factorization and the chosen rotation.
They are not functions only of the task operators
$\mW_{Q,1}\mW_{K,1}^\top$ and $\mW_{Q,2}\mW_{K,2}^\top$.
Thus, even if the rotation makes the factors closer, the final factor-wise average is not the same as an operator-level merge.

The same issue appears for VO.
After matching model $1$ with an orthogonal matrix $\mS$, factor averaging gives
\begin{equation}
    \bar{\mW}_V
    =
    \frac{1}{2}
    \left(
        \mW_{V,1}\mS+\mW_{V,2}
    \right),
    \qquad
    \bar{\mW}_O
    =
    \frac{1}{2}
    \left(
        \mS^\top\mW_{O,1}+\mW_{O,2}
    \right),
\end{equation}
and therefore
\begin{align}
    \bar{\mW}_V\bar{\mW}_O
    =
    \frac{1}{4}
    \Big(
        &
        \mW_{V,1}\mW_{O,1}
        +
        \mW_{V,2}\mW_{O,2}
        \nonumber
        +
        \mW_{V,1}\mS\mW_{O,2}
        +
        \mW_{V,2}\mS^\top\mW_{O,1}
    \Big).
\end{align}
Again, the cross terms are artifacts of merging the factors rather than the composed VO operators.
SLOA avoids this issue by solving directly for the merged QK and VO operators.

\paragraph{Different invariance classes.}
Rotation-symmetry matching works within an orthogonal symmetry class, optionally combined with scalar rescaling.
This is natural for parameter matching because orthogonal transformations preserve Frobenius norms and lead to efficient Procrustes-style solutions.
However, the functional non-identifiability of QK and VO operators is larger.
For any invertible matrix $\mR$, the QK product is preserved by
\begin{equation}
    \mW_Q \mapsto \mW_Q\mR,
    \qquad
    \mW_K \mapsto \mW_K\mR^{-\top},
\end{equation}
since
\begin{equation}
    (\mW_Q\mR)
    (\mW_K\mR^{-\top})^\top
    =
    \mW_Q\mR\mR^{-1}\mW_K^\top
    =
    \mW_Q\mW_K^\top .
\end{equation}
Similarly, the VO product is invariant under
\begin{equation}
    \mW_V \mapsto \mW_V\mR,
    \qquad
    \mW_O \mapsto \mR^{-1}\mW_O .
\end{equation}
SLOA is invariant to this full product-level ambiguity because it only depends on the composed operator.
If two factorizations realize the same QK or VO operator, they become exactly the same object for SLOA.
By contrast, rotation matching selects a representative within a restricted symmetry family and then still applies a parameter-space fusion rule.

\paragraph{Different objectives.}
The optimality guarantees of the two methods are for different problems.
Rotation-symmetry matching is optimal for choosing an equivalent parameterization that minimizes Euclidean distance to an anchor model under orthogonal rotation constraints.
Its objective is parameter proximity, where it
chooses an equivalent representative that is close in weight space.
This is useful because many model-fusion methods perform better when the input models are close in parameter space.

SLOA's closed-form solution is optimal for a local data-aware operator-alignment objective.
Its objective is local functional agreement:
\begin{equation}
    \mathcal{A}
    \left(
        \widehat{\mZ}_{t,i};
        \bm{\theta}
    \right)
    \approx
    \mathcal{A}
    \left(
        \mZ_{t,i};
        \bm{\theta}_t
    \right),
\end{equation}
together with a data-dependent operator-space regularizer.
Thus, SLOA does not ask for the closest equivalent factorization of a task model.
It asks for the merged local operator that best reproduces the task-specific local behaviors under the activations that the partially merged model will actually produce.

\paragraph{Sequential correction versus independent parameter matching.}
Rotation-symmetry matching is applied as a matching step between model parameters, typically module by module or layer by layer.
The matching transformation preserves the function of the source model before fusion, but it does not explicitly account for the activation distribution induced by earlier merged components.

SLOA is sequential in a different sense.
At operator $o_m$, it evaluates the merged operator on the local input
$\widehat{\mZ}_{t,i}^{(m)}$ produced by the partially merged model, while the target is the output of the task-specific operator on the task-model local input $\mZ_{t,i}^{(m)}$:
\begin{equation}
    \mathcal{A}
    \left(
        \widehat{\mZ}_{t,i}^{(m)};
        \bm{\theta}
    \right)
    \approx
    \mathcal{A}
    \left(
        \mZ_{t,i}^{(m)};
        \bm{\theta}_t^{(m)}
    \right).
\end{equation}
This makes each local merge corrective with respect to upstream merging errors.
The method therefore aligns operators under the actual execution path of the merged network, rather than only reducing parameter distance between the original task models.

\paragraph{Anchor-based matching versus symmetric multi-model alignment.}
The core rotation-matching formulation aligns one source model to an anchor model.
For multi-model fusion, this requires choosing an anchor or performing matching relative to some reference representation.
This anchor choice is natural for parameter matching, but it introduces an asymmetry among the input models.

SLOA treats all task models symmetrically in the local objective:
\begin{equation}
    \frac{1}{T}
    \sum_{t=1}^{T}
    \frac{1}{n_t}
    \sum_{i=1}^{n_t}
    \left\|
        \mathcal{A}
        \left(
            \widehat{\mZ}_{t,i};
            \bm{\theta}
        \right)
        -
        \mathcal{A}
        \left(
            \mZ_{t,i};
            \bm{\theta}_t
        \right)
    \right\|_F^2 .
\end{equation}
No single task model serves as the canonical parameter reference.
The merged operator is estimated from a balanced objective over all task-specific operators and calibration states.

\paragraph{Rank and capacity control.}
Rotation-symmetry matching preserves the original architecture during matching and subsequent fusion.
In particular, the attention head dimension remains fixed, and the matched QK and VO products remain represented through the original head bottleneck.

SLOA decouples the merged operator from its factorization rank.
The aligned QK and VO operators are first solved in the ambient residual-space operator space
$\mathbb{R}^{d\times d}$.
Only after the operator is obtained do we choose a factorization rank $r$.
When $r=d_k$, the original head dimension is preserved.
When $r>d_k$, SLOA can retain additional singular directions of the merged operator that arise from combining multiple task-specific subspaces.
This rank-expansion mechanism has no analogue in rotation-symmetry matching, because rotation matching modifies existing factors but does not expose or refactorize the merged operator spectrum.

\paragraph{Data usage and scope.}
Rotation-symmetry matching is primarily data-free: the matching transformation is computed from model parameters and can be plugged into several downstream fusion rules.
This is an advantage when calibration data are unavailable.
SLOA uses calibration data and therefore targets the training-free, data-aware setting.
The calibration data serve two roles: they define the local behavior-matching targets and they induce the data-dependent operator metric used for regularization.
Thus, SLOA trades the data-free nature of rotation matching for a more functional, data-aware merge.

\paragraph{Complementarity.}
The two methods should be viewed as complementary rather than redundant.
Rotation-symmetry matching shows that transformer attention has useful continuous parameter-space symmetries and that exploiting these symmetries can improve conventional model fusion.
SLOA takes the next step from symmetry-aware factor matching to symmetry-invariant operator merging.
Instead of choosing better representatives within the equivalence class and then merging factors, SLOA quotients out the factorization ambiguity by working directly with the composed QK and VO operators.

\subsection{Comparison to ProDistill}
\label{app:comparison_prodistill}

We provide a detailed comparison between SLOA and ProDistill~\citep{xu2025scalable}, as both methods use calibration data and proceed progressively through the network.
Despite this superficial similarity, the two methods address different merging problems and optimize fundamentally different objects.

\paragraph{Overview of ProDistill.}
ProDistill views model merging as a multi-teacher distillation problem.
Let $\bm{\theta}_0$ be the pretrained initialization and let $\bm{\theta}_t$ be the fine-tuned model for task $t$.
Writing the task vector as
$
    \bm{\tau}_t = \bm{\theta}_t - \bm{\theta}_0,
$
ProDistill parameterizes the merged model as
\begin{equation}
    \widehat{\bm{\theta}}
    =
    \bm{\theta}_0
    +
    \sum_{t=1}^{T}
    \bm{\lambda}_t \circ \bm{\tau}_t ,
\end{equation}
where $\bm{\lambda}_t$ are learned merging coefficients and $\circ$ denotes element-wise multiplication.
The method optimizes these coefficients so that the intermediate representations of the merged model match those of the fine-tuned teacher models on a small unlabeled validation set.

To reduce memory cost, ProDistill decomposes the objective layer by layer.
At layer $\ell$, it solves an optimization problem of the form
\begin{equation}
\label{eq:prodistill_layer_objective}
    \min_{\{\bm{\lambda}^{(\ell)}_t\}_{t=1}^{T}}
    \sum_{t=1}^{T}
    \frac{1}{|D_t|}
    \sum_{(z_1,z_2)\in D_t^{(\ell-1)}}
    \left\|
    \phi^{(\ell)}
    \left(
        \bm{\theta}_0^{(\ell)}
        +
        \sum_{s=1}^{T}
        \bm{\lambda}^{(\ell)}_s
        \circ
        \bm{\tau}^{(\ell)}_s,
        z_1
    \right)
    -
    \phi^{(\ell)}
    \left(
        \bm{\theta}^{(\ell)}_t,
        z_2
    \right)
    \right\|_2^2 ,
\end{equation}
where $z_1$ is the input activation produced by the partially merged model and $z_2$ is the corresponding input activation produced by the fine-tuned model.
After optimizing the coefficients for layer $\ell$, ProDistill updates the paired activations and proceeds to the next layer.
Thus, ProDistill is a progressive activation-distillation method over task-vector coefficients.

\paragraph{SLOA is training-free, whereas ProDistill is training-based.}
The most direct difference is the optimization procedure.
ProDistill trains the merging coefficients $\bm{\lambda}^{(\ell)}_t$ using gradient descent.
Although the optimization is restricted to one layer at a time, it still requires iterative updates, gradients, learning-rate choices, optimizer states, and stopping criteria.

SLOA, in contrast, is training-free.
For each local operator, SLOA solves a regularized least-squares problem in closed form:
\begin{equation}
\bm{\theta}_{\mathrm{merge}}
=
\underset{\bm{\theta}\in\Theta_{\mathrm{amb}}}{\arg\min}
\;
\mathcal{L}_{\mathrm{act}}(\bm{\theta})
+
\frac{1}{T}
\sum_{t=1}^{T}
\left\|
\bm{\theta}
-
\bm{\theta}_t
\right\|_{\widehat{\chi}_t}^{2}.
\end{equation}
The closed-form solution is possible because SLOA chooses \emph{local functional operators} whose behavior maps are linear with respect to $\bm{\theta}$.
Thus, SLOA does not learn merging coefficients by gradient descent; it directly solves for the merged local operator.

\paragraph{SLOA merges local functional operators, whereas ProDistill merges task-vector parameter coordinates.}
ProDistill retains the standard task-vector parameterization.
For each layer, the merged layer remains in the affine span
\begin{equation}
    \bm{\theta}_0^{(\ell)}
    +
    \operatorname{span}_{\circ}
    \left\{
        \bm{\tau}^{(\ell)}_1,\ldots,\bm{\tau}^{(\ell)}_T
    \right\},
\end{equation}
where the coefficients act element-wise on the task-vector coordinates.
This gives ProDistill a flexible parameter-space interpolation mechanism, but the primitive objects being combined are still the original parameter tensors.

SLOA instead changes the unit of merging.
In transformer attention, the individual matrices $\mW_Q,\mW_K,\mW_V,\mW_O$ are not the local functional units.
The attention scores depend on the composed QK operator
\begin{equation}
    \bm{\theta}_{\mathrm{QK}}
    =
    \frac{1}{\sqrt{d_k}}
    \mW_Q \mW_K^\top ,
\end{equation}
and the logits depends on the composed VO operator
\begin{equation}
    \bm{\theta}_{\mathrm{VO}}
    =
    \mW_V \mW_O .
\end{equation}
SLOA merges these composed operators directly.
This distinction is central: ProDistill optimizes a block-level representation-matching objective over task-vector coefficients, while SLOA constructs a \emph{local} operator-level objective over the functional objects that determine the computation.

\paragraph{SLOA removes attention-factorization ambiguity by construction.}
Attention heads have internal reparameterization symmetries.
For any invertible matrix $\mR$,
\begin{equation}
    \mW_Q \mW_K^\top
    =
    (\mW_Q \mR)
    (\mW_K \mR^{-\top})^\top .
\end{equation}
Thus, two attention heads can produce identical QK score operators while using different query/key coordinates.
An analogous ambiguity exists for the value/output factors.
Any method that operates directly on the individual matrices $\mW_Q,\mW_K,\mW_V,\mW_O$ must handle this coordinate ambiguity implicitly.

ProDistill uses an activation loss, so its objective can partially compensate for poor parameter-space interpolation.
However, its parameterization still acts on the individual task-vector coordinates of the original projection matrices.
It does not explicitly identify the invariant QK and VO operators, nor does it merge in the quotient space induced by attention-head reparameterizations.
Consequently, the factorization ambiguity remains inside the optimization problem.

SLOA removes this ambiguity before solving the merge.
Since the optimization variable is the composed operator itself, two different factorizations that realize the same QK or VO operator correspond to the same object in SLOA.
The merged operator is therefore invariant to internal changes of head coordinates, and factorization is postponed until after the operator has been aligned.

\paragraph{SLOA is more fine-grained than layer-wise distillation inside transformer blocks.}
ProDistill progresses layer by layer.
The layer is treated as the unit whose output representation should match the teacher representation.
This is a general design and can be applied to nonlinear modules because the optimization is performed by gradient descent.

SLOA instead uses a topological order over local operators inside the network.
For a transformer block, SLOA can separately align LayerNorm affine maps, standard linear maps, QK attention operators, and VO attention operators.
This allows the method to respect the internal computational structure of attention rather than treating the whole attention layer or block as a single black-box feature map.
In particular, SLOA distinguishes the two roles of self-attention:
\begin{itemize}
    \item the QK operator determines token-token routing through the pre-softmax attention scores;
    \item the VO operator determines what information is written back to the logit after attention weights are applied.
\end{itemize}
This decomposition is absent from ProDistill.
ProDistill aligns block outputs, whereas SLOA aligns the local functional mechanisms that produce those outputs.

\paragraph{Both methods are progressive, but the progressive state has a different meaning.}
Both ProDistill and SLOA use a progressive procedure in which later modules are merged under activations affected by earlier merges.
However, the role of these activations differs.

In ProDistill, the progressive state consists of paired features $(z_1,z_2)$, where $z_1$ is produced by the partially merged student and $z_2$ is produced by the corresponding teacher.
The layer objective trains task-vector coefficients so that the student layer output matches the teacher layer output on these paired inputs.

In SLOA, the progressive state defines the local inputs
$
    \widehat{\mZ}_{t,i}^{(m)}
$
received by the partially merged model and
$
    \mZ_{t,i}^{(m)}
$
received by the task model.
The merged operator is evaluated on the former, while the target output is generated by the task-specific local operator on the latter:
\begin{equation}
    \mathcal{A}
    \left(
        \widehat{\mZ}_{t,i}^{(m)};
        \bm{\theta}
    \right)
    \approx
    \mathcal{A}
    \left(
        \mZ_{t,i}^{(m)};
        \bm{\theta}^{(m)}_t
    \right).
\end{equation}
Thus, SLOA is progressive at the level of local operator alignment, not at the level of learned task-vector coefficients.
Each merge is corrective under the actual execution path of the partially merged network, while still solving a closed-form local regression problem.

\paragraph{SLOA decouples operator alignment from factorization rank.}
A further distinction is architectural capacity.
ProDistill keeps the original model architecture and learns coefficients for the existing task-vector parameters.
In particular, an attention head retains the original head dimension.
Thus, ProDistill does not create a mechanism for increasing the rank of the QK or VO operators after merging.

SLOA explicitly separates the ambient operator from its final factorization.
For a head with original dimension $d_k$, each task-specific QK or VO operator has rank at most $d_k$.
However, the merged multi-task operator may have larger effective rank because different tasks can use different singular directions.
SLOA first solves for the merged operator in the ambient space
$
    \Theta_{\mathrm{amb}}=\mathbb{R}^{d\times d}
$
and only afterward factorizes it at a chosen rank $r$.
When $r=d_k$, SLOA preserves the original head width.
When $r>d_k$, SLOA retains additional singular directions supported by the aligned operator.
This rank-expansion mechanism is not available in ProDistill, because ProDistill does not decouple the merged attention operator from its original factorization.

\paragraph{The two methods make different trade-offs.}
ProDistill is a general distillation-based merging framework.
Its strength is that it can optimize feature matching for arbitrary modules without requiring the module to be linear in a chosen operator.
This generality comes at the cost of gradient-based optimization and a parameter-coordinate task-vector representation.

SLOA makes the opposite trade-off.
It focuses on modules that admit a local behavior map linear in an appropriate operator.
This restriction enables closed-form merging, data-dependent operator geometry, attention-operator invariance, and post-merge rank control.
Therefore, SLOA is not simply a training-free variant of ProDistill.
It changes the mathematical object being merged: from task-vector coordinates trained to match layer features, to local functional operators solved by sequential regularized regression.

\section{Detailed derivation of local operator alignment}
\label{app: local operator alignment}
As described in \S\ref{sec:method}, the Sequential Local Operator Alignment (SLOA) method merges a single operator at a time. By explicitly writing the term $\mathcal{L}_{\mathrm{act}}(\bm{\theta})$, the unified loss objective for merging is defined as:
\begin{equation}
    L(\bm{\theta}) =
    \frac{1}{T} \sum_{t=1}^{T} \left(
    \frac{1}{n_t} \sum_{i=1}^{n_t} \left\| \mathcal{A}(\widehat{\mZ}_{t,i};\bm{\theta}) - \mathcal{A}(\mZ_{t,i};\bm{\theta}_t) \right\|_F^2
    + \left\| \bm{\theta} - \bm{\theta}_t \right\|_{\widehat{\chi}_t}^{2}
    \right)
\label{eq:unified_operator_alignment_loss}
\end{equation}

\begin{equation}
    \bm{\theta}_{\mathrm{merge}}
    =
    \underset{\bm{\theta}\in\Theta_{\mathrm{amb}}}{\arg\min}
    \;
    L(\bm{\theta})
\end{equation}
Here, $\widehat{\chi}_t = \{ \widehat{\mZ}_{t,i} \}_{i=1}^{n_t}$ is the collection of local inputs produced by the partially merged model. For standard linear operators and QK operator that taking only the activation as input, $\widehat{\mZ}_{t,i} = \widehat{\mX}_{t,i} \in \mathbb{R}^{L \times d}$, while for components like the VO circuit, it also includes the composed attention matrix.

Now, for each operator, we will show how this unified loss objective can be reduced to a structured regression problem with a closed-form solution. We will also detail the explicit form of the data-dependent regularization term and the ambient operator space $\Theta_{\mathrm{amb}}$ for each component, applying constraints (such as rank) during the subsequent factorization step for each operators mentioned in~\S\ref{sec:local_operators_transformer}.

\subsection{Optimal QK Operator}
\label{proof:qk_merging}

For the Query-Key circuit, for each attention head $j$, let $\mW_Q^{(j)}, \mW_K^{(j)} \in \mathbb{R}^{d \times d_k}$ denote the query and key matrices, respectively. The consolidated score operator is defined as $\bm{\theta}^{(j)} = \frac{1}{\sqrt{d_k}} \mW_Q^{(j)} \mW_K^{(j)\top} \in \mathbb{R}^{d \times d}$. Since the operation acts bilinearly on the input representations, when evaluated on a single calibration example $i$ from task $t$, the behavior map and the customized task-level regularization term are defined as:
\begin{align}
\mathcal{A}(\widehat{\mZ}_{t,i};\bm{\theta^{(j)}}) = \widehat{\mZ}_{t,i} \bm{\theta^{(j)}} \widehat{\mZ}_{t,i}^\top, \quad \left\|
\bm{\theta^{(j)}} - \bm{\theta^{(j)}}_t \right\|_{\widehat{\chi}_t}^{2} =
\left\|
\mLambda_t^{1/2}
(\bm{\theta^{(j)}}-\bm{\theta^{(j)}}_t)
\mLambda_t^{1/2}
\right\|_F^2
\end{align}
where $\mLambda_t \in \mathbb{R}^{d \times d}$ is a diagonal precision matrix derived from the partially merged state collection $\widehat{\chi}_t$, and $\mLambda_t^{1/2}$ is its principal square root. Hereafter, we drop the head index $j$ for notational brevity, as the alignment is solved independently for each head.

To find the optimal operator, we first relax the rank-$d_k$ structural constraint $\Theta_{\mathrm{amb}}$ to solve for an unconstrained consolidated operator (the rank constraint is subsequently enforced via Singular Value Decomposition). Substituting the behavior map into the unified loss objective (\ref{eq:unified_operator_alignment_loss}) and writing $\mathcal{A}(\mZ_{t,i};\bm{\theta}_t) = \mY_{t,i}$ as target, we have:
\begin{equation}
L(\bm{\theta}) = \frac{1}{T} \sum_{t=1}^T \left( \frac{1}{n_t} \sum_{i=1}^{n_t} \left\| \widehat{\mZ}_{t,i} \bm{\theta} \widehat{\mZ}_{t,i}^\top - \mY_{t,i} \right\|_F^2 + \left\| \mLambda_t^{1/2} (\bm{\theta} - \bm{\theta}_t) \mLambda_t^{1/2} \right\|_F^2 \right)
\end{equation}
where $n_t$ is the number of calibration samples for task $t$.

Because the relaxed objective function $L(\bm{\theta})$ is strictly convex, the global minimum is found by setting the gradient with respect to $\bm{\theta}$ to zero. Utilizing the matrix derivative identity $\nabla_{\bm{\theta}} \|\mA\bm{\theta} \mA^\top - \mY\|_F^2 = 2\mA^\top(\mA\bm{\theta} \mA^\top - \mY)\mA$ and the linearity of the gradient over the finite sum, we obtain:
\begin{equation}
\nabla_{\bm{\theta}} L(\bm{\theta}) = \frac{2}{T} \sum_{t=1}^T \left( \frac{1}{n_t} \sum_{i=1}^{n_t} \widehat{\mZ}_{t,i}^\top \left( \widehat{\mZ}_{t,i} \bm{\theta} \widehat{\mZ}_{t,i}^\top - \mY_{t,i} \right) \widehat{\mZ}_{t,i} + \mLambda_t (\bm{\theta} - \bm{\theta}_t) \mLambda_t \right) = 0
\end{equation}

Dividing by the scalar $\frac{2}{T}$ and distributing the terms inside the summations gives:
\begin{equation}
\sum_{t=1}^T \left( \frac{1}{n_t} \sum_{i=1}^{n_t} \left( \widehat{\mZ}_{t,i}^\top \widehat{\mZ}_{t,i} \bm{\theta} \widehat{\mZ}_{t,i}^\top \widehat{\mZ}_{t,i} - \widehat{\mZ}_{t,i}^\top \mY_{t,i} \widehat{\mZ}_{t,i} \right) + \mLambda_t \bm{\theta} \mLambda_t - \mLambda_t \bm{\theta}_t \mLambda_t \right) = 0
\end{equation}

We group the terms dependent on the merged parameter $\bm{\theta}$ on the left-hand side, and move the constant target terms to the right-hand side:
\begin{equation}
\sum_{t=1}^T \left( \frac{1}{n_t} \sum_{i=1}^{n_t} \left( \widehat{\mZ}_{t,i}^\top \widehat{\mZ}_{t,i} \bm{\theta} \widehat{\mZ}_{t,i}^\top \widehat{\mZ}_{t,i} \right) + \mLambda_t \bm{\theta} \mLambda_t \right) = \sum_{t=1}^T \left( \frac{1}{n_t} \sum_{i=1}^{n_t} \left( \widehat{\mZ}_{t,i}^\top \mY_{t,i} \widehat{\mZ}_{t,i} \right) + \mLambda_t \bm{\theta}_t \mLambda_t \right)
\label{eq:qk_equilibrium}
\end{equation}

To solve Equation \ref{eq:qk_equilibrium} analytically, we define $\mR_{t,i} = \widehat{\mZ}_{t,i}^\top \widehat{\mZ}_{t,i} \in \mathbb{R}^{d \times d}$, and let $\mC \in \mathbb{R}^{d \times d}$ denote the right-hand side \emph{constant} matrix. The equation simplifies to:
\begin{equation}
    \sum_{t=1}^{T} \left( \left[ \frac{1}{n_t}\sum_{i=1}^{n_t} \mR_{t,i} \bm{\theta} \mR_{t,i} \right] + \mLambda_t \bm{\theta} \mLambda_t \right) = \mC
\end{equation}

To isolate $\bm{\theta}$, we apply the vectorization operator $\operatorname{vec}(\cdot)$. Using the Kronecker product identity $\operatorname{vec}(\mA \bm{\theta} \mB) = (\mB^\top \otimes \mA) \operatorname{vec}(\bm{\theta})$ and noting that both $\mR_{t,i}$ and $\mLambda_t$ are symmetric, we obtain:
\begin{equation}
    \sum_{t=1}^{T} \left( \left[ \frac{1}{n_t}\sum_{i=1}^{n_t} (\mR_{t,i} \otimes \mR_{t,i}) \right] + (\mLambda_t \otimes \mLambda_t) \right) \operatorname{vec}(\bm{\theta}) = \operatorname{vec}(\mC)
\end{equation}

By defining the aggregated precision matrix $\mH \in \mathbb{R}^{d^2 \times d^2}$ as the operator on the left-hand side, the system reduces to the linear form:
\begin{equation}
    \mH \operatorname{vec}(\bm{\theta}) = \operatorname{vec}(\mC)
\end{equation}

Assuming $\mH$ is invertible (or utilizing the Moore-Penrose pseudo-inverse $\mH^\dagger$ in rank-deficient scenarios), we isolate the vectorized parameters and apply the inverse vectorization operator to recover the optimal unconstrained operator $\bm{\theta}_{\mathrm{merge}} \in \mathbb{R}^{d \times d}$:
\begin{equation}
    \bm{\theta}_{\mathrm{merge}} = \operatorname{vec}^{-1}\left( \mH^{-1} \operatorname{vec}(\mC) \right) \quad \square
    \label{eq:qk_optimal_closed_form_solution}
\end{equation}
The above derivation yields an exact closed-form analytical solution for the optimal QK operator, which is subsequently factorized into $\widehat{\mW}_Q$ and $\widehat{\mW}_K$ via SVD to reintroduce the rank constraint.

\textbf{Computational Remark.} While Equation \ref{eq:qk_optimal_closed_form_solution} provides an exact analytical equality, directly materializing and inverting the Kronecker matrix $\mH$ incurs an intractable time complexity of $\mathcal{O}(d^6)$ and space complexity of $\mathcal{O}(d^4)$. For standard Transformer dimensions (e.g., $d \ge 768$), this is computationally prohibitive. Therefore, in practice, we compute the optimal $\bm{\theta}_{\mathrm{merge}}$ iteratively without ever materializing $\mH$ by employing the Conjugate Gradient (CG) algorithm \footnote{\url{https://en.wikipedia.org/wiki/Conjugate_gradient_method}}.

\subsubsection{Rank-Constrained Factorization of the QK Operator}
\label{app:qk_factorization}
Following the unconstrained optimization derived in \S\ref{proof:qk_merging}, the optimal consolidated operator $\bm{\theta}_{\mathrm{merge}} \in \mathbb{R}^{d \times d}$ must be projected back onto the structural constraints of the attention mechanism. By the Eckart-Young-Mirsky theorem, the optimal rank-$r$ approximation in the Frobenius norm is obtained via the truncated Singular Value Decomposition (SVD):
\begin{equation}
    \bm{\theta}_{\mathrm{merge}} = \mU \mSigma \mV^\top \quad \implies \quad \bm{\theta}_{\mathrm{merge}}^{(r)} = \mU_r \mSigma_r \mV_r^\top
\end{equation}
where $\mU_r, \mV_r \in \mathbb{R}^{d \times r}$ contain the top $r$ left and right singular vectors, respectively, and $\mSigma_r \in \mathbb{R}^{r \times r}$ is the diagonal matrix of the top $r$ singular values.
The standard attention mechanism computes the QK logit matrix as $\frac{1}{c} \widehat{\mW}_Q \widehat{\mW}_K^\top$, where $c$ is the attention scaling denominator (e.g., $c=\sqrt{r}$ or $c=\sqrt{d_k}$). To satisfy the equality $\frac{1}{c} \widehat{\mW}_Q \widehat{\mW}_K^\top = \bm{\theta}_{\mathrm{merge}}^{(r)}$, we distribute the singular values and the scalar $c$ symmetrically across the QK weight matrices:
\begin{align}
    \widehat{\mW}_Q &= \mU_r (c \mSigma_r)^{1/2} \in \mathbb{R}^{d \times r} \label{eq:wq_svd} \\
    \widehat{\mW}_K &= \mV_r (c \mSigma_r)^{1/2} \in \mathbb{R}^{d \times r} \label{eq:wk_svd}
\end{align}

\subsection{Optimal VO Operator}
\label{proof:vo_merging}

For value-output (VO) circuits, for each head $j$, we have $\mW_{V}^{(j)} \in \mathbb{R}^{d \times d_k}$ as the value matrix, and $\mW_O^{(j)} \in \mathbb{R}^{d_k \times d}$ as the output matrix. The consolidated operator is defined as $\bm{\theta}^{(j)} = \mW_V^{(j)} \mW_O^{(j)} \in \mathbb{R}^{d \times d}$. In accordance with the sequential topological ordering described in \S\ref{sec:method}, the VO operator takes as input the activation $\widehat{\mX}_{t,i}$ as well as the attention matrix $\widehat{\mA}_{t,i} \in \mathbb{R}^{L \times L}$, which is computed using the previously aligned QK operator. We explicitly define the partially merged local input state as $\widehat{\mZ}_{t,i} = \widehat{\mA}_{t,i}\widehat{\mX}_{t,i}$.
The operation of the VO circuit is linear with respect to the input representations, and thus when evaluated on a single calibration example $i$ from task $t$, the behavior map and the customized task-level regularization term are defined as:
\begin{align}
\mathcal{A}(\widehat{\mZ}_{t,i};\bm{\theta}) = \widehat{\mZ}_{t,i} \bm{\theta}, \quad \left\|
\bm{\theta} - \bm{\theta}_t \right\|_{\widehat{\chi}_t}^{2} =
\left\|\mLambda_t^{1/2}
(\bm{\theta}-\bm{\theta}_t)
\right\|_F^2
\end{align}
where $\widehat{\chi}_t = \{ \widehat{\mZ}_{t,i}\}_{i=1}^{n_t}$ representing attention-weighted input representations. Similar to the QK case, we droped $j$ for ease of notaion, since the same operation is opertated for each attention head. $\mLambda_t$ is a diagonal matrix derived from the attention score weighted partially merged state collection $\widehat{\chi}_t$, and $\mLambda_t^{1/2}$ is its square root.

For the VO operator, we solve for $L(\bm{\theta})$ with the structural constraint $\Theta_{\mathrm{amb}} = \{rank(\bm{\theta}) \leq r \}$. Substituting these into the unified objective function (\ref{eq:unified_operator_alignment}) and writing $\mathcal{A}(\mZ_{t,i};\bm{\theta}_t) = \mY_{t,i}$ as target yields:
\begin{equation}
L(\bm{\theta}) = \frac{1}{T} \sum_{t=1}^T \left( \frac{1}{n_t} \sum_{i=1}^{n_t} \left\| \widehat{\mZ}_{t,i} \bm{\theta} - \mY_{t,i} \right\|_F^2 + \left\| \mLambda_t^{1/2} (\bm{\theta} - \bm{\theta}_t) \right\|_F^2 \right)
\end{equation}
\begin{equation}
    \bm{\theta}_{\mathrm{merge}}
    =
    \underset{\bm{\theta}\in\{rank(\bm{\theta}) \leq r \}}{\arg\min}
    \;
    L(\bm{\theta})
\end{equation}

To solve this rank-constrained minimization problem, we first eliminate the summations by stacking the task-specific matrices vertically. Because scaling the objective by a constant $T$ does not alter the minimization, minimizing $L(\bm{\theta})$ is strictly equivalent to minimizing a single, consolidated Frobenius norm:
\begin{equation}
    \min_{\operatorname{rank}(\bm{\theta}) \le r} \left\| \widehat{\mZ}\bm{\theta} - \widehat{\mY} \right\|_F^2
\end{equation}
where augmented matrices $\widehat{\mZ}$ and $\widehat{\mY}$ are defined as:
\begin{equation*}
    \widehat{\mZ} = \begin{bmatrix}
        \widehat{\mZ}_{1,1} \\ \vdots \\ \widehat{\mZ}_{1,n_1} \\ \vdots \\ \widehat{\mZ}_{T,1} \\ \vdots \\ \widehat{\mZ}_{T,n_T} \\ \mLambda_1^{1/2} \\ \vdots \\ \mLambda_T^{1/2}
    \end{bmatrix} \quad
    \widehat{\mY} = \begin{bmatrix}
        \mY_{1,1} \\ \vdots \\ \mY_{1,n_1} \\ \vdots \\ \mY_{T,1} \\ \vdots \\ \mY_{T,n_T} \\ \mLambda_1^{1/2}\bm{\theta}_1 \\ \vdots \\ \mLambda_T^{1/2}\bm{\theta}_T
    \end{bmatrix}
\end{equation*}

Let $\operatorname{trunc}_r(\cdot)$ denote the truncated Singular Value Decomposition (SVD) operator that retains only the top $r$ singular components of a matrix. The exact closed-form solution for this rank-constrained objective is given by~\cite{friedland2006generalizedrankconstrainedmatrixapproximations}:
\begin{equation}
    \bm{\theta}_{\mathrm{merge}} = (\widehat{\mZ}^\top \widehat{\mZ})^{-1/2} \operatorname{trunc}_r \!\left[ (\widehat{\mZ}^\top \widehat{\mZ})^{-1/2} \widehat{\mZ}^\top \widehat{\mY} \right]
\end{equation}
Crucially, we do not need to explicitly materialize the massive stacked matrices $\widehat{\mZ}$ and $\widehat{\mY}$. By computing the inner products of the block matrices, they elegantly collapse back into the $\widehat{\bm{G}}_t$ and $\widehat{\bm{C}}_t$ matrices:
\begin{align}
    \widehat{\mG}_t &= \sum_{i=1}^{n_t} \frac{1}{n_t}\widehat{\mZ}_{t,i}^\top \widehat{\mZ}_{t,i} + \mLambda_t \\
    \widehat{\mC}_t &= \sum_{i=1}^{n_t} \frac{1}{n_t}\widehat{\mZ}_{t,i}^\top \mY_{t,i} + \mLambda_t \bm{\theta}_t
\end{align}
Substituting these sums back into the solution, we obtain the final closed-form equation for the rank-constrained VO operator:
\begin{equation}
    \bm{\theta}_{\mathrm{merge}} = \left( \sum_{t=1}^T \widehat{\mG}_t \right)^{-1/2}\operatorname{trunc}_r \!\left[ \left( \sum_{t=1}^T \widehat{\mG}_t \right)^{-1/2} \left( \sum_{t=1}^T \widehat{\mC}_t \right) \right] \square
\end{equation}

\subsubsection{Factorization of the VO operator}
After solving our rank constraint solution for VO operator, we need to factorize the optimal merged parameter back to $\widehat{\mW}_{V}$ and $\widehat{\mW}_{O}$.
\begin{equation}
    \bm{\theta}_{\mathrm{merge}} = \mU \mSigma \mV^\top \quad \implies \quad \bm{\theta}_{\mathrm{merge}}^{(r)} = \mU_r \mSigma_r \mV_r^\top
\end{equation}
where $\mU_r, \mV_r \in \mathbb{R}^{d \times r}$ contain the top $r$ left and right singular vectors, respectively, and $\mSigma_r \in \mathbb{R}^{r \times r}$ is the diagonal matrix of the top $r$ singular values.
Unlike QK circuit, there is no scaling factor in OV circuit of standard attention
Therefore the factorization follows:
\begin{align}
    \widehat{\mW}_V &= \mU_r \mSigma_r^{1/2} \in \mathbb{R}^{d \times r} \label{eq:wc_svd} \\
    \widehat{\mW}_O &= \mV_r  \mSigma_r^{1/2} \in \mathbb{R}^{d \times r} \label{eq:wo_svd}
\end{align}

\subsection{Optimal MLP operator}
\label{proof:mlp_merging}
For MLP layers, the behavior map is defined as and the regularization term is defined as:
\begin{equation}
   \mathcal{A}(\widehat{\mZ}_{t,i};\bm{\theta}) = \widehat{\mZ}_{t,i} \bm{\theta}, \quad
    \left\| \bm{\theta} - \bm{\theta}_t \right\|_{\widehat{\chi}_t}^{2} = \left\| \mLambda_t^{1/2} (\bm{\theta} - \bm{\theta}_t) \right\|_F^2
\end{equation}

Unlike the VO circuit, the MLP layer does not have a low-rank bottleneck constraint. Therefore, we can directly solve for the optimal operator $\bm{\theta}$ without imposing any structural constraints. Substituting these into the unified objective function Eq. (\ref{eq:unified_operator_alignment_loss}) and and writing $\mathcal{A}(\mZ_{t,i};\bm{\theta}_t) = \mY_{t,i}$ as target yields:
\begin{equation}
L(\bm{\theta}) = \frac{1}{T} \sum_{t=1}^T \left( \frac{1}{n_t} \sum_{i=1}^{n_t} \left\| \widehat{\mZ}_{t,i} \bm{\theta} - \mY_{t,i} \right\|_F^2 + \left\| \mLambda_t^{1/2} (\bm{\theta} - \bm{\theta}_t) \right\|_F^2 \right)
\end{equation}
The objective reeduced to  standard regularized multivariate least squares problem (i.e., ridge regression under a Mahalanobis prior), and the closed-form solution can be derived by setting the gradient to zero. The optimal operator $\bm{\theta}$ is given by:
\begin{equation}
    \nabla_{\bm{\theta}} L({\bm{\theta}}) = \frac{2}{T} \sum_{t=1}^T \left( \frac{1}{n_t} \sum_{i=1}^{n_t} \widehat{\mZ}_{t,i}^\top (\widehat{\mZ}_{t,i} \bm{\theta} - \mY_{t,i}) + \mLambda_t (\bm{\theta} - \bm{\theta}_t) \right) = 0
\end{equation}
Rearranging the terms, we obtain:
\begin{equation}
    \left( \sum_{t=1}^T  \frac{1}{n_t} \sum_{i=1}^{n_t} \widehat{\mZ}_{t,i}^\top \widehat{\mZ}_{t,i} + \mLambda_t \right)  \bm{\theta} = \sum_{t=1}^T \left( \frac{1}{n_t} \sum_{i=1}^{n_t} \widehat{\mZ}_{t,i}^\top \mY_{t,i} + \mLambda_t \bm{\theta}_t \right)
\end{equation}
Let us define Gram matrix $\widehat{\mG}_t$ and cross-covariance matrix $\widehat{\mC}_t$ as:
\begin{equation}
    \widehat{\mG}_t = \frac{1}{n_t} \sum_{i=1}^{n_t} \widehat{\mZ}_{t,i}^\top \widehat{\mZ}_{t,i} + \mLambda_t, \quad
    \widehat{\mC}_t = \frac{1}{n_t} \sum_{i=1}^{n_t} \widehat{\mZ}_{t,i}^\top \mY_{t,i} + \mLambda_t \bm{\theta}_t
\end{equation}
The closed-form solution for the optimal MLP operator is then given by:
\begin{equation}
    \bm{\theta} = \left( \sum_{t=1}^T \widehat{\mG}_t \right)^{-1} \left( \sum_{t=1}^T \widehat{\mC}_t \right) \square
\end{equation}

\subsection{Optimal LayerNorm Operator}

For the Layer Normalization module, the operator consists of a scale vector $\vgamma \in \mathbb{R}^d$ and a bias vector $\vbeta \in \mathbb{R}^d$. Evaluated on a single calibration example $i$ from task $t$, the behavior map and the data-dependent regularization term are defined as:
\begin{align}
\mathcal{A}(\widehat{\mX}_{t,i};\vgamma,\vbeta) &= \nu(\widehat{\mX}_{t,i})\operatorname{diag}(\vgamma) + \mathbf{1}\vbeta^\top \\
\left\| \bm{\theta} - \bm{\theta}_t \right\|_{\widehat{\chi}_t}^{2} &= \left\| \mLambda_{\gamma,t}^{1/2} (\vgamma - \vgamma_t) \right\|_2^2 + \left\| \mLambda_{\beta,t}^{1/2} (\vbeta - \vbeta_t) \right\|_2^2
\end{align}
where $\nu(\cdot)$ denotes the normalization that zeroes the mean and scales to unit variance across the feature dimension, and $\mathbf{1} \in \mathbb{R}^L$ is a vector of ones. The diagonal precision matrices $\mLambda_{\gamma,t}$ and $\mLambda_{\beta,t}$ govern the task-dependent geometry for the scale and bias, respectively.

Because the affine transformation acts independently on each feature dimension $j \in \{1, \dots, d\}$, the Frobenius norm cleanly decouples. We can therefore break the global objective into $d$ independent optimization problems.

Let $\hat{\vx}_{t,i} \in \mathbb{R}^L$ denote the $j$-th column of the normalized input $\nu(\widehat{\mX}_{t,i})$, and let $\vy_{t,i} \in \mathbb{R}^L$ denote the $j$-th column of the target output $\mY_{t,i}$. Correspondingly, let the scalars $\gamma, \beta \in \mathbb{R}$ denote the $j$-th elements of the vectors $\vgamma$ and $\vbeta$, and let $\gamma_t, \beta_t \in \mathbb{R}$ denote the $j$-th elements of the task-specific parameters $\vgamma_t$ and $\vbeta_t$. Furthermore, let the scalars $\lambda_{\gamma, t}, \lambda_{\beta, t} \in \mathbb{R}$ represent the $j$-th diagonal entries of the precision matrices $\mLambda_{\gamma,t}$ and $\mLambda_{\beta,t}$. Dropping the feature dimension index $j$ for notational brevity, the objective for a single feature dimension over the empirical calibration mixture is:
\begin{equation}
    L(\gamma, \beta) = \frac{1}{T} \sum_{t=1}^T \left( \frac{1}{n_t} \sum_{i=1}^{n_t} \left\| \hat{\vx}_{t,i} \gamma + \mathbf{1}\beta - \vy_{t,i} \right\|_2^2 + \lambda_{\gamma, t} (\gamma - \gamma_t)^2 + \lambda_{\beta, t} (\beta - \beta_t)^2 \right)
\end{equation}

Because this dimension-wise objective is strictly convex, we find the global minimum by setting the partial derivatives with respect to $\gamma$ and $\beta$ to zero.

For the scale parameter $\gamma$, we obtain:
\begin{align}
    \frac{\partial L}{\partial \gamma} = 0 \implies \frac{2}{T} \sum_{t=1}^T \left( \frac{1}{n_t} \sum_{i=1}^{n_t} \hat{\vx}_{t,i}^\top (\hat{\vx}_{t,i} \gamma + \mathbf{1}\beta - \vy_{t,i}) + \lambda_{\gamma, t}(\gamma - \gamma_t) \right) = 0
\end{align}
Rearranging terms, we get:
\begin{equation}
     \sum_{t=1}^T \left( \frac{1}{n_t} \sum_{i=1}^{n_t} \hat{\vx}_{t,i}^\top \hat{\vx}_{t,i} + \lambda_{\gamma, t} \right) \gamma + \sum_{t=1}^T \left( \frac{1}{n_t}
    \sum_{i=1}^{n_t} \hat{\vx}_{t,i}^\top \mathbf{1} \right) \beta = \sum_{t=1}^T \left( \frac{1}{n_t} \sum_{i=1}^{n_t} \hat{\vx}_{t,i}^\top \vy_{t,i} + \lambda_{\gamma, t} \gamma_t \right)
    \label{eq:ln_gamma}
\end{equation}

For the bias parameter $\beta$, noting that $\mathbf{1}^\top \mathbf{1} = L$ (the sequence length), we obtain:
\begin{align}
    \frac{\partial L}{\partial \beta} = 0 &\implies \frac{2}{T} \sum_{t=1}^T \left( \frac{1}{n_t} \sum_{i=1}^{n_t} \mathbf{1}^\top (\hat{\vx}_{t,i} \gamma + \mathbf{1}\beta - \vy_{t,i}) + \lambda_{\beta, t}(\beta - \beta_t) \right) = 0 \nonumber \\
    &\implies \sum_{t=1}^T \left( \frac{1}{n_t} \sum_{i=1}^{n_t} \mathbf{1}^\top \hat{\vx}_{t,i} \right) \gamma + \sum_{t=1}^T \left( L + \lambda_{\beta, t} \right) \beta = \sum_{t=1}^T \left( \frac{1}{n_t} \sum_{i=1}^{n_t} \mathbf{1}^\top \vy_{t,i} + \lambda_{\beta, t} \beta_t \right)
    \label{eq:ln_beta}
\end{align}

Eq. (\ref{eq:ln_gamma}) and (\ref{eq:ln_beta}) form a coupled system of two linear equations. To express the closed-form solution elegantly, we define the following scalar accumulators:
\begin{align*}
    S_{xx} &= \sum_{t=1}^T \left( \frac{1}{n_t} \sum_{i=1}^{n_t} \hat{\vx}_{t,i}^\top \hat{\vx}_{t,i} + \lambda_{\gamma, t} \right), \quad
    &S_{xy} &= \sum_{t=1}^T \left( \frac{1}{n_t} \sum_{i=1}^{n_t} \hat{\vx}_{t,i}^\top \vy_{t,i} + \lambda_{\gamma, t} \gamma_t \right) \\
    S_1 &= \sum_{t=1}^T \left( L + \lambda_{\beta, t} \right), \quad
    &S_y &= \sum_{t=1}^T \left( \frac{1}{n_t} \sum_{i=1}^{n_t} \mathbf{1}^\top \vy_{t,i} + \lambda_{\beta, t} \beta_t \right) \\
    S_x &= \sum_{t=1}^T \left( \frac{1}{n_t} \sum_{i=1}^{n_t} \mathbf{1}^\top \hat{\vx}_{t,i} \right)
\end{align*}

This coupled system cleanly reduces to a symmetric $2 \times 2$ matrix equation:
\begin{equation}
    \begin{bmatrix} S_{xx} & S_x \\ S_x & S_1 \end{bmatrix} \begin{bmatrix} \gamma \\ \beta \end{bmatrix} = \begin{bmatrix} S_{xy} \\ S_y \end{bmatrix}
\end{equation}

By inverting this $2 \times 2$ matrix (e.g., via Cramer's rule), we obtain the exact closed-form solution for the optimal merged scale and bias parameters for this feature dimension:
\begin{equation}
    \gamma = \frac{S_1 S_{xy} - S_x S_y}{S_{xx} S_1 - S_x^2} \quad \beta = \frac{S_{xx} S_y - S_x S_{xy}}{S_{xx} S_1 - S_x^2}
\end{equation}
We can perform the same operation for each dimension of vectors, $\vgamma$ and $\vbeta$, in parallel to obtain the final solution.

\subsection{Rank-controlled factorization of merged attention operators}
\label{app:rank_details}

This section details the factorization step summarized in \S\ref{sec:rank_increase}.

\paragraph{Low-rank structure of attention heads.}
Let $d$ be the residual-stream dimension and $d_k$ the head dimension. For a single head, $\mW_Q,\mW_K,\mW_V\in\mathbb{R}^{d\times d_k}$ and $\mW_O\in\mathbb{R}^{d_k\times d}$, so the local operators
\begin{equation}
\bm{\theta}_{QK}=\mW_Q\mW_K^\top/\sqrt{d_k},
\qquad
\bm{\theta}_{VO}=\mW_V\mW_O
\end{equation}
are $d\times d$ matrices of rank at most $d_k$: each head implements a $d\times d$ operator through a width-$d_k$ bottleneck. SLOA merges these operators in the ambient space $\Theta_{\mathrm{amb}}=\mathbb{R}^{d\times d}$, where no rank constraint is imposed, and restores a factorized form only afterwards. All steps below are applied independently to every head, and separately to its QK and VO operators.

\paragraph{Truncated SVD.}
Given a merged operator $\bar{\bm{\theta}}\in\Theta_{\mathrm{amb}}$ with SVD $\bar{\bm{\theta}}=\mU\mSigma\mV^\top$, let $\mU_r$, $\mSigma_r$, and $\mV_r$ retain the top $r$ singular directions. By the Eckart--Young theorem, $\bar{\bm{\theta}}_r=\mU_r\mSigma_r\mV_r^\top$ is the best rank-$r$ approximation of $\bar{\bm{\theta}}$ in Frobenius norm.

\paragraph{QK factorization.}
Let $\tau$ denote the attention scaling used after factorization: $\tau=\sqrt{d_k}$ if the default implementation is kept, or $\tau=\sqrt{r}$ if the expanded head uses the standard scaling for width $r$. Setting
\begin{equation}
\mW_Q=\mU_r(\tau\mSigma_r)^{1/2},
\qquad
\mW_K=\mV_r(\tau\mSigma_r)^{1/2}
\end{equation}
gives $\mW_Q\mW_K^\top/\tau=\bar{\bm{\theta}}_{QK,r}$ for either choice of $\tau$.

\paragraph{VO factorization.}
The VO behavior map $\mA\mX\bm{\theta}_{VO}$ has no scaling, so we set
\begin{equation}
\mW_V=\mU_r\mSigma_r^{1/2},
\qquad
\mW_O=\mSigma_r^{1/2}\mV_r^\top,
\end{equation}
which gives $\mW_V\mW_O=\bar{\bm{\theta}}_{VO,r}$. In both cases the head width becomes $r$ while $d$ is unchanged. The QK and VO ranks may differ ($r_{\rm qk}$, $r_{\rm ov}$; Fig.~\ref{fig:component-ablation}); unless stated otherwise, $r=r_{\rm qk}=r_{\rm ov}$.

\paragraph{Why the merged operator may need rank above $d_k$.}
The rank $r$ controls how much of the merged operator's spectrum is retained; $r=d_k$ preserves the original head width. Different task models, however, may use different low-dimensional subspaces, and a combination of low-rank operators need not share their rank: the sum of two rank-one operators with different singular directions has rank two, and in general $\operatorname{rank}\bigl(\sum_t\bm{\theta}_t\bigr)$ can reach $\min(d,\,Td_k)$. Truncating the merged operator back to $d_k$ can therefore discard directions that are useful for multi-task behavior; $r>d_k$ retains them.

\paragraph{Does rank expansion overfit?}
Increasing $r$ beyond $d_k$ adds capacity without gradient-based training, which might be expected to overfit the calibration set. It does not, however, introduce free parameters: the factorization only retains singular directions already present in the aligned operator $\bar{\bm{\theta}}$, which is anchored to the task-specific operators by the regularizer in Eq.~(\ref{eq:unified_operator_alignment}). Consistent with this, rank expansion did not reduce average accuracy in any of our settings and typically improved it (\S\ref{sec:exp}, \S\ref{sec:scaling_data_rank}). The cost is a wider head at inference, quantified in \S\ref{sec:inference_cost}.

\section{Experimental details}
\label{app:Experimental details}
\subsection{Handiling bias terms}
\label{app:bias_terms}
Many prior works treat weight matrices and bias terms differently during the merging
operation, opting to simply average the bias terms across models. In contrast, we
handle bias terms correctly with respect to our objective via an augmentation trick
that applies uniformly to all linear layers, including MLPs and attention projections.
For instance with weight matrix $\mW \in \mathbb{R}^{d \times
d_{\text{out}}}$ and bias $\vb \in \mathbb{R}^{d_{\text{out}}}$, the standard
forward pass for a batch of $L$ sequence length is:
\begin{equation}
    \mY = \mX\mW + \textbf{1}_L \vb^\top,
    \qquad \mX \in \mathbb{R}^{l \times d},\quad
    \mY \in \mathbb{R}^{L \times d_{\text{out}}}.
\end{equation}
We rewrite this compactly by augmenting the input and weight matrix:
\begin{equation}
    \mY = \tilde{\mX}\,\tilde{\mW},
    \qquad
    \tilde{\mX} = \bigl[\mX \mid \textbf{1}_n\bigr]
    \in \mathbb{R}^{l \times (d+1)},
    \qquad
    \tilde{\mW} = \begin{bmatrix} \mW \\ \vb^\top \end{bmatrix}
    \in \mathbb{R}^{(d+1) \times d_{\text{out}}}.
\end{equation}
Our merging procedure
operates directly on the augmented matrices $\{\tilde{\mW}_t\}_{t=1}^T$,
yielding $\tilde{\mW}_{\text{merge}} \in \mathbb{R}^{(d+1) \times
d_{\text{out}}}$. The merged weight matrix and bias are then recovered by partitioning. The same augmentation applies to the QK and VO operators, where the bias terms are absorbed into the augmented weight matrices.

\subsection{Dataset description}
\label{sec:dataset_dec}
\subsubsection{Dataset for vision task.}
\label{sec:vision_datasets}
In our vision model merging experiments, we utilize the datasets and task-specific fine-tuned checkpoints provided by the FusionBench framework \citep{tang2024fusionbench}. We employ the Vision Transformer (ViT) variant of CLIP \citep{radford2021learningtransferablevisualmodels} as our base architecture. Rather than appending a randomly initialized linear classification head during fine-tuning, these models retain CLIP's native dual-encoder structure. Specifically, classification is formulated as a multi-modal contrastive task: the model computes the cosine similarity between the visual embedding of the input image and the textual embeddings generated by passing class-specific text prompts (e.g., ``a photo of a \{\textit{class}\}'') through the frozen text encoder. To maintain strict consistency with this paradigm, our evaluation reports the standard zero-shot classification accuracy utilizing these text-driven prototypes. The datasets used for model merging are listed in Table~\ref{tab:image_classification_datasets}. We also report the calibration set size and the percentage of the training set used for calibration. The calibration set size is fixed at 256 samples across all datasets, which represents a small fraction of the training data.
\begin{table}[t]
\centering
\caption{Data statistics of the 20 image classification datasets used for model merging, grouped by task set: the first 8 form 8-tasks the next 6 extend it to 14-tasks, and the final 6 extend it to 20-tasks.}
\label{tab:image_classification_datasets}
\resizebox{0.8\textwidth}{!}{
\begin{tabular}{l p{3cm} r r r r r}
\toprule
Task & Data Source & \# Classes & Train & Test & Calib. & Calib.~(\%) \\
\midrule
\multicolumn{7}{c}{\textit{8-Tasks}} \\
\midrule
SUN397      & \href{https://huggingface.co/datasets/tanganke/sun397}{tanganke/sun397}                               & 397 & 76{,}128  & 10{,}875 & 256 & 0.34  \\
GTSRB       & \href{https://huggingface.co/datasets/tanganke/gtsrb}{tanganke/gtsrb}                                 &  43 & 39{,}209  & 12{,}630 & 256 & 0.65  \\
Cars        & \href{https://huggingface.co/datasets/tanganke/stanford_cars}{tanganke/stanford\_cars}                & 196 &  8{,}144  &  8{,}041 & 256 & 3.14  \\
EuroSAT     & \href{https://huggingface.co/datasets/tanganke/eurosat}{tanganke/eurosat}                             &  10 & 21{,}600  &  5{,}400 & 256 & 1.19  \\
RESISC45    & \href{https://huggingface.co/datasets/tanganke/resisc45}{tanganke/resisc45}                           &  45 & 25{,}200  &  6{,}300 & 256 & 1.02  \\
DTD         & \href{https://huggingface.co/datasets/tanganke/dtd}{tanganke/dtd}                                     &  47 &  3{,}760  &  1{,}880 & 256 & 6.81  \\
MNIST       & \href{https://huggingface.co/datasets/ylecun/mnist}{ylecun/mnist}                                     &  10 & 60{,}000  & 10{,}000 & 256 & 0.43  \\
SVHN        & \href{https://huggingface.co/datasets/ufldl-stanford/svhn}{ufldl-stanford/svhn}                       &  10 & 73{,}257  & 26{,}032 & 256 & 0.35  \\
\midrule
\multicolumn{7}{c}{\textit{14 Tasks (+ 6 datasets)}} \\
\midrule
Flowers     & \href{https://huggingface.co/datasets/dpdl-benchmark/oxford_flowers102}{dpdl-benchmark/oxford\_flowers102} & 102 &  1{,}020 &  6{,}149 & 256 & 25.10 \\
PCAM        & \href{https://huggingface.co/datasets/1aurent/PatchCamelyon}{1aurent/PatchCamelyon}                   &   2 & 262{,}144 & 32{,}768 & 256 & 0.10  \\
FER2013     & \href{https://huggingface.co/datasets/clip-benchmark/wds_fer2013}{clip-benchmark/wds\_fer2013}         &   7 & 28{,}709  &  3{,}589 & 256 & 0.89  \\
Pet         & \href{https://huggingface.co/datasets/timm/oxford-iiit-pet}{timm/oxford-iiit-pet}                     &  37 &  3{,}680  &  3{,}669 & 256 & 6.96  \\
STL10       & \href{https://huggingface.co/datasets/tanganke/stl10}{tanganke/stl10}                                 &  10 &  5{,}000  &  8{,}000 & 256 & 5.12  \\
CIFAR100    & \href{https://huggingface.co/datasets/tanganke/cifar100}{tanganke/cifar100}                           & 100 & 50{,}000  & 10{,}000 & 256 & 0.51  \\
\midrule
\multicolumn{7}{c}{\textit{20 Tasks (+ 6 datasets)}} \\
\midrule
CIFAR10     & \href{https://huggingface.co/datasets/tanganke/cifar10}{tanganke/cifar10}                             &  10 & 50{,}000  & 10{,}000 & 256 & 0.51  \\
Food101     & \href{https://huggingface.co/datasets/ethz/food101}{ethz/food101}                                     & 101 & 75{,}750  & 25{,}250 & 256 & 0.34  \\
FashionMNIST & \href{https://huggingface.co/datasets/zalando-datasets/fashion_mnist}{zalando-datasets/fashion\_mnist} & 10 & 60{,}000  & 10{,}000 & 256 & 0.43  \\
EMNIST      & \href{https://huggingface.co/datasets/tanganke/emnist_letters}{tanganke/emnist\_letters}               &  37 & 88{,}800  & 14{,}800 & 256 & 0.29  \\
KMNIST      & \href{https://huggingface.co/datasets/tanganke/kmnist}{tanganke/kmnist}                               &  10 & 60{,}000  & 10{,}000 & 256 & 0.43  \\
RenderedSST2 & \href{https://huggingface.co/datasets/nateraw/rendered-sst2}{nateraw/rendered-sst2}                  &   2 &  6{,}920  &  1{,}821 & 256 & 3.70  \\
\bottomrule
\end{tabular}
}
\end{table}

\subsubsection{Dataset for NLU task.}
\label{sec:NLU_datasets}
For language model experiments, we use model checkpoints from the DARE~\citep{yu2024language}
work. We evaluate the performance of these models on the 7 GLUE benchmark~\citep{wang2018glue},%
\footnote{\url{https://huggingface.co/datasets/nyu-mll/glue}}
, which covers a range of NLU tasks, including grammatical acceptability (CoLA), natural language classification (MNLI, RTE, QNLI), semantic equivalence (MRPC, QQP), and sentiment analysis (SST-2). The dataset statistics and evaluation metrics for these tasks are summarized in Table~\ref{tab:glue_stats}. We use standard training and validation splits for each task, and we set aside a small calibration set of 256 examples from the training set to perform model merging. The percentage of the calibration set relative to the training set is also reported in the table.
\begin{table}[t]
\centering
\caption{Summary of GLUE tasks, Metric used, and dataset statistics}
\label{tab:glue_stats}
\begin{tabular}{l r p{2.5cm} r r r r}
\toprule
Task & \# Labels & Primary Metric & Train & Val. & Calib. & Calib.~(\%) \\
\midrule
CoLA & 2 & Matthews correlation & 8{,}551 & 1{,}043 & 256 & 2.99 \\
MNLI & 3 & Accuracy & 392{,}702 & 9{,}815 & 256 & 0.07 \\
MRPC & 2 & Avg.\ F1 \& Acc. & 3{,}668 & 408 & 256 & 6.98 \\
QNLI & 2 & Accuracy & 104{,}743 & 5{,}463 & 256 & 0.24 \\
QQP & 2 & Avg.\ F1 \& Acc. & 363{,}849 & 40{,}430 & 256 & 0.07 \\
RTE & 2 & Accuracy & 2{,}490 & 277 & 256 & 10.28 \\
SST-2 & 2 & Accuracy & 67{,}349 & 872 & 256 & 0.38 \\
\bottomrule
\end{tabular}
\end{table}

\subsubsection{Dataset for LLM experiments}
\label{sec:LLM_datasets}
For the LLM merging experiments, we did not use an off-the-shelf pool of LoRA experts for LLaMA-2-7B \citep{touvron2023llama2openfoundation}; instead, we constructed our own task pool. We train three LoRA experts \citep{hu2022lora} independently from the LLaMA-2-7B base checkpoint, each targeting all linear projections at rank 32($\alpha{=}64$), for 3 epochs with a cosine learning-rate schedule (peak learning rate $2\times10^{-4}$) and an effective batch size of 32: a \textit{math} expert on MetaMathQA~\citep{yu2023metamath}, a \textit{code} expert on Magicoder-Evol-Instruct-110K~\citep{wei2024magicoder}, and an \textit{instruction-following} expert on
the persona-conditioned instruction-following split of Tulu-3-SFT~\citep{lambert2025tulu3pushingfrontiers}. Each expert is trained with a single, fixed prompt template applied to every example (the code expert uses Magicoder's own instruction/response wrapper; the math and instruction-following experts share an Alpaca-style instruction/response template) see~\ref{fig:llm_templates}, and the same template is reused uniformly at both calibration and evaluation time.  We evaluate each task on its corresponding
standard benchmark rather than a held-out split of its own training
corpus: GSM8K \citep{cobbe2021gsm8k} for math (exact-match accuracy under
flexible answer extraction), HumanEval \citep{chen2021evaluating} for code
(pass@1), and IFEval \citep{zhou2023instructionfollowingevaluationlargelanguage} for instruction-following (prompt-level strict accuracy).

All three are evaluated zero-shot through \texttt{lm-evaluation-harness} \footnote{\url{https://github.com/eleutherai/lm-evaluation-harness}}, with each task's training template baked into its eval prompt. Dataset statistics, calibration-set sizes, and calibration coverage (a fixed 256 examples per task, drawn from the same training split and formatted with the same template used for fine-tuning) are summarized in Table~\ref{tab:llm_data_stats}.
\begin{table}[t]
\centering
\caption{Summary of the LLM fine-tuning corpora, evaluation benchmarks, and dataset statistics. Each row is one LoRA expert (rank 32, all-linear modules) fine-tuned from LLaMA-2-7B.}
\label{tab:llm_data_stats}
\resizebox{\textwidth}{!}{
\begin{tabular}{l l l r r r r r}
\toprule
Task & Train Corpus & Benchmark & Primary Metric & Train & Eval & Calib. & Calib.~(\%) \\
\midrule
Math & MetaMathQA & GSM8K & Exact match (flexible extract) & 395,000 & 1,319 & 256 & 0.065 \\
Code & Magicoder-Evol-Instruct-110K & HumanEval & pass@1 & 111,183 & 164 & 256 & 0.230 \\
Instruction-following & Tulu-3 Persona-IF & IFEval & Prompt-level strict accuracy & 29,980 & 541 & 256 & 0.854 \\
\bottomrule
\end{tabular}
}
\end{table}

\begin{figure}[t]
    \centering
    \begin{subfigure}{0.42\linewidth}
        \centering
        \fbox{%
            \parbox{0.7\linewidth}{%
                \raggedright
                \ttfamily\small
                \#\#\# Instruction:\\
                \{instruction\}\\[0.8em]
                \#\#\# Response:\\
            }%
        }
        \caption{Alpaca-style template (math, instruction-following).}
        \label{fig:alpaca_template}
    \end{subfigure}
    \hfill
    \begin{subfigure}{0.42\linewidth}
        \centering
        \fbox{%
            \parbox{0.9\linewidth}{%
                \raggedright
                \ttfamily\small
                You are an exceptionally intelligent coding assistant that consistently delivers accurate and reliable responses to user instructions.\\[0.8em]
                @@ Instruction\\
                \{instruction\}\\[0.8em]
                @@ Response\\
            }%
        }
        \caption{Magicoder template (code).}
        \label{fig:magicoder_template}
    \end{subfigure}

    \caption{Prompt templates applied uniformly at training, calibration, and evaluation time for each LoRA expert. The math and instruction-following experts share the Alpaca-style template \subref{fig:alpaca_template}; the code expert uses Magicoder's own template \subref{fig:magicoder_template}.}
    \label{fig:llm_templates}
\end{figure}

\section{Additional experiments and ablation analysis of SLOA}
\label{sec:ablation}
We study how SLOA scales with calibration data, rank expansion, and attention-circuit capacity. We also investigate the behavior of SLOA on LoRA finetuned models.
\begin{figure}[t]
    \centering
    \begin{subfigure}{0.30\linewidth}
    \centering
    \includegraphics[width=\linewidth]{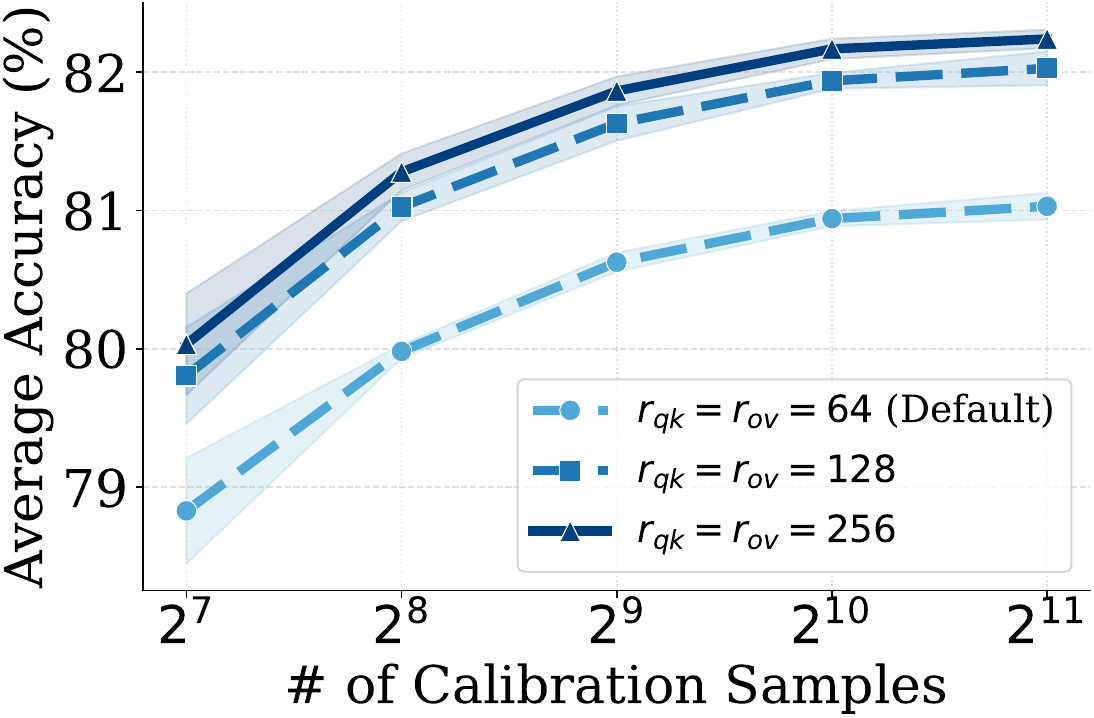}
    \caption{ViT-B/32}
    \label{fig:calib_vitb32}
    \end{subfigure}
    \begin{subfigure}{0.30\linewidth}
    \centering
    \includegraphics[width=\linewidth]{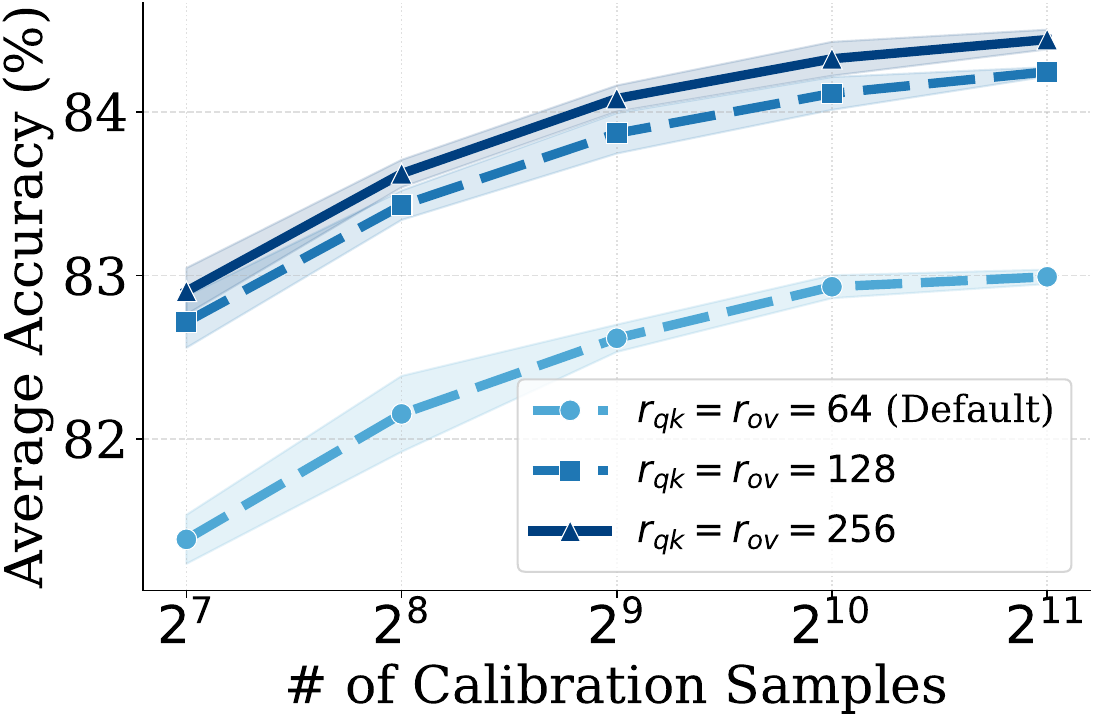}
    \caption{ViT-B/16}
    \label{fig:calib_vitb16}
    \end{subfigure}
    \begin{subfigure}{0.30\linewidth}
    \centering
    \includegraphics[width=\linewidth]{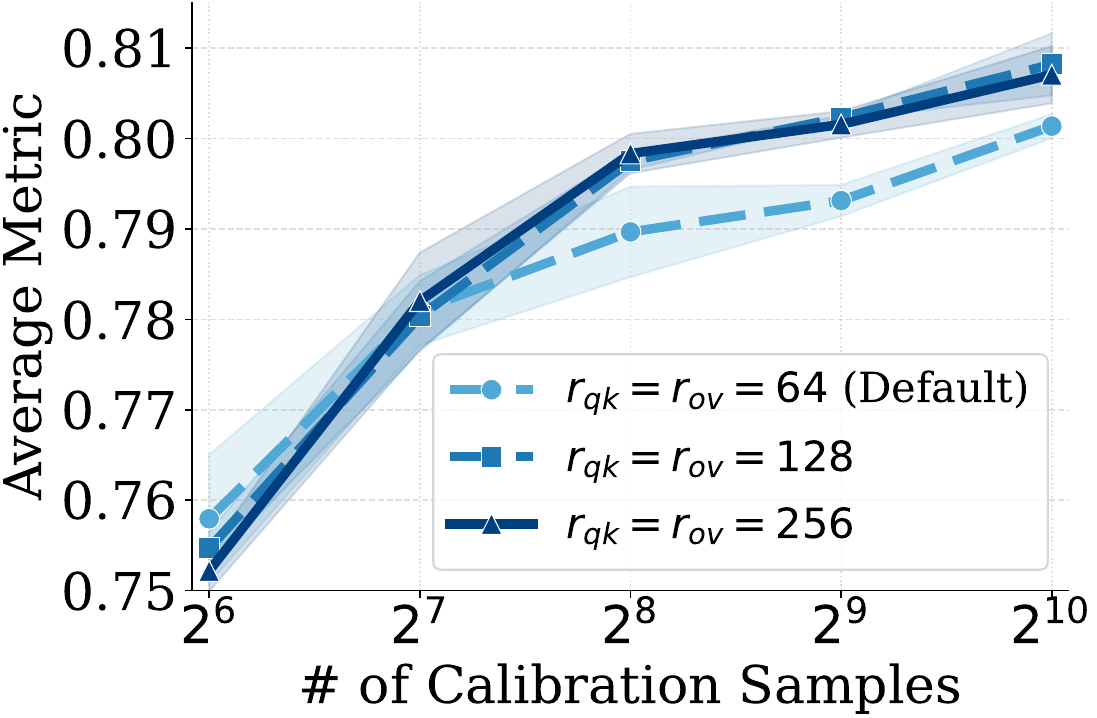}
    \caption{RoBERTa}
    \label{fig:calib_roberta}
    \end{subfigure}
    \caption{Average performance of SLOA as the number of calibration data increases for \subref{fig:calib_vitb32}~ViT-B/32 and \subref{fig:calib_vitb16}~ViT-B/16 on 20 vision tasks and \subref{fig:calib_roberta}~RoBERTa on 7 NLU tasks.}
    \label{fig:rank_samplenumber}
\end{figure}
\begin{table}[t]
\centering
 \caption{
Performance comparison of model merging methods on ViT-B/16 across 8, 14, and 20 vision classification tasks. For SLOA, $r$ denotes the retained factorization rank used for both QK and VO local operators. The setting $r=64$ matches the original ViT-B/16 attention-head dimension.
}
\begin{tabular}{clcccc}
\toprule
\multirow{2}{*}{}
& \multirow{2}{*}{Method}
& \multirow{2}{*}{\# Samples}
& \multicolumn{3}{c}{Avg. Accuracy $\uparrow$}  \\
\cmidrule(lr){4-6}
& & & 8 Tasks & 14 Tasks & 20 Tasks \\
\midrule
& Fine-tuned            & --  & 92.34 & 91.31 & 91.61 \\
\midrule
\multirow{6}{*}{\rotatebox[origin=c]{90}{Data-Free}}
& Weight Averaging       & --  & 72.28 & 69.72 & 64.81 \\
& Task Arithmetic        & --  & 77.17 & 70.92 & 64.23 \\
& TIES-Merging           & --  & 80.21 & 71.48 & 66.31 \\
& TSV-M                  & --  & 86.70 & 82.09 & 78.31 \\
& Iso-C                  & --  & 88.36 & 82.39 & 75.82 \\
& Iso-CTS                & --  & \textbf{89.01} & 84.54 & 78.50 \\
\midrule
\multirow{5}{*}{\rotatebox[origin=c]{90}{Training-Free}}
& Fisher Merging         & 256 & 73.67 & 71.00 & 66.50 \\
& RegMean                & 256 & 86.13 & 81.43 & 75.53 \\
& RegMean++              & 256 & 87.25 & 82.23 & 76.64 \\
& \cellcolor{lightpurple}SLOA ($r{=}64$) & \cellcolor{lightpurple}256 & \cellcolor{lightpurple}87.16 & \cellcolor{lightpurple}84.19 & \cellcolor{lightpurple}82.12 \\
& \cellcolor{lightblue}SLOA ($r{=}128$) & \cellcolor{lightblue}256 & \cellcolor{lightblue}88.04 & \cellcolor{lightblue}\textbf{85.16} & \cellcolor{lightblue}\textbf{83.43} \\
\bottomrule
\end{tabular}
\label{tab:vision_main_results_b16}
\end{table}

\subsection{Scaling to a larger vision backbone: ViT-B/16}
\label{appendix:sloa_vitb16}
To test whether SLOA's advantage persists as model capacity grows, we repeat the ViT-B/32 multi-task merging experiments (\S\ref{sec:Merging ViT models}) on the larger ViT-B/16 backbone across the same 8-, 14-, and 20-task benchmarks (Table~\ref{tab:vision_main_results_b16}); as a complementary scaling direction, we also evaluate SLOA on a much larger, 7B-parameter decoder-only LLM in the main text (\S\ref{sec:Merging LLM LoRA}, Table~\ref{tab:llm_main}).
The trend mirrors ViT-B/32: SLOA's margin over the best baseline widens as the number of merged tasks grows. On 8 tasks, the data-free baseline Iso-CTS is narrowly the strongest method overall (89.01 vs.\ 88.04 for SLOA ($r=128$)); by 14 tasks, SLOA (85.16) has overtaken it (84.54); and by 20 tasks, SLOA leads by nearly 5 points (83.43 vs.\ 78.50). As on ViT-B/32, increasing $r$ from 64 to 128 improves performance at every task count, with the gain growing alongside the number of merged tasks.
\citet{ortizjimenez2023task} show that larger, better-pretrained vision backbones tend to be more weight-disentangled than smaller ones, so task-specific fine-tuning directions interfere less under simple weight-space combination. This may explain why data-free baselines remain competitive with SLOA on ViT-B/16 when only a few tasks are merged. However, as the number of merged tasks grows and interference increases, SLOA, which aligns functional operators rather than raw weights, wins by an increasingly clear margin.

\subsection{Merging LoRA-adapted ViT experts}
\label{sec:vit_lora}
\begin{table}
\centering
 \caption{
Normalized per-task accuracy of model merging methods on ViT-B/32 LoRA (KnOTS, $r{=}16$~\citep{stoica2025knots}) experts across 8 vision tasks. Normalized accuracy is the mean over tasks of (merged/fine-tuned) accuracy $\times$ 100; the fine-tuned LoRA oracle's raw average accuracy is 84.28\%.
 }
 \vspace{-1 mm}
 \resizebox{0.6\textwidth}{!}{
 \begin{tabular}{clcc}
\toprule
\multirow{2}{*}{}
& \multirow{2}{*}{Method}
& \multirow{2}{*}{\# Samples}
& \multirow{2}{*}{Norm. Acc.\ (\%) $\uparrow$} \\
& & & \\
\midrule
& Fine-tuned (LoRA, oracle) & -- & 100.00 \\
\midrule
\multirow{8}{*}{\rotatebox[origin=c]{90}{Data-Free}}
& Simple Avg & -- & 63.76 \\
& TA & -- & 63.70 \\
& TIES & -- & 63.70 \\
& DARE-TIES & -- & 63.70 \\
& KnOTS-TIES & -- & 68.00 \\
& KnOTS-DARE-TIES & -- & 63.90 \\
& Iso-C & -- & 73.60 \\
& Iso-CTS & -- & 73.70 \\
\midrule
\multirow{5}{*}{\rotatebox[origin=c]{90}{Training-Free}}
& Fisher Merging & 256 & 62.81 \\
& RegMean & 256 & 72.72 \\
& RegMean++ & 256 & 81.11 \\
& \cellcolor{lightpurple}SLOA (r=64) & \cellcolor{lightpurple}256 & \cellcolor{lightpurple}82.66 \\
& \cellcolor{lightblue}SLOA (r=128) & \cellcolor{lightblue}256 & \cellcolor{lightblue}\textbf{88.07} \\
\bottomrule
 \end{tabular}
}

 \label{tab:vision_lora_b32_8tasks}
\end{table}

SLOA extends naturally to LoRA-adapted experts: there is no additional difficulty caused by the LoRA parameterization itself. For a LoRA-adapted task $t$-weight $W_t = W_0 + \Delta W_t$, with $\Delta W_t = B_t A_t$, we first form the effective fine-tuned weight $W_t$ and then construct the same functional QK, VO, linear, and normalization operators used for fully fine-tuned experts; no change to the SLOA local objective is required. This distinction is important because LoRA's factors are themselves non-identifiable: $B_t A_t = (B_t R)(R^{-1} A_t)$ for any invertible $R$, so methods that operate separately on $A_t$ and $B_t$ can depend on an arbitrary low-rank coordinate system. SLOA does not merge the individual LoRA factors; it merges the effective functional operator and is therefore invariant to this factorization ambiguity.
We evaluate this on ViT-B/32 LoRA experts (KnOTS checkpoints) \citep{stoica2025knots} across the same 8-task vision benchmark. Table~\ref{tab:vision_lora_b32_8tasks} shows that SLOA achieves the highest normalized accuracy among all methods: SLOA ($r=128$) reaches 88.07\%, 7.0 points above the strongest baseline. Notably, even SLOA ($r=64$) already exceeds all baselines at 82.66\%, despite each task-specific LoRA update being rank 16 -- reinforcing the view that the subspace spanned by the merged operator can exceed the rank of any individual task expert.
\subsection{Scaling with calibration samples and rank}
\label{sec:scaling_data_rank}
Figure~\ref{fig:rank_samplenumber} reports average accuracy as a function of calibration samples for vision (ViT-B/32 and ViT-B/16, 20 tasks) and language (RoBERTa, 7 tasks). Across both modalities and all rank configurations, SLOA improves steadily as calibration data increase. This is consistent with our unified objective in Eq.~(\ref{eq:unified_operator_alignment}): larger calibration sets provide a better-conditioned system for estimating the merged weights. Notably, even with only 128 samples, our SLOA outperforms all baselines (see Tables~\ref{tab:vision_main_results_b32} -\ref{tab:roberta}), demonstrating strong data efficiency.
Increasing the rank from the default $r = r_{\rm qk} = r_{\rm ov} = 64$ to $128$ and $256$ further improves performance. On vision tasks, higher rank consistently helps across sample sizes, reflecting the greater diversity of the 20-task setting. On NLU tasks, the improvement from $64$ to $128$ is clear, while $256$ gives diminishing returns, suggesting that the effective rank required for merging grows with task number and diversity.

\subsection{Error accumulation across network depth}
\label{sec:error_accumulation}
\begin{figure}[t]
    \centering
    \begin{subfigure}{0.38\linewidth}
    \centering
    \includegraphics[width=\linewidth]{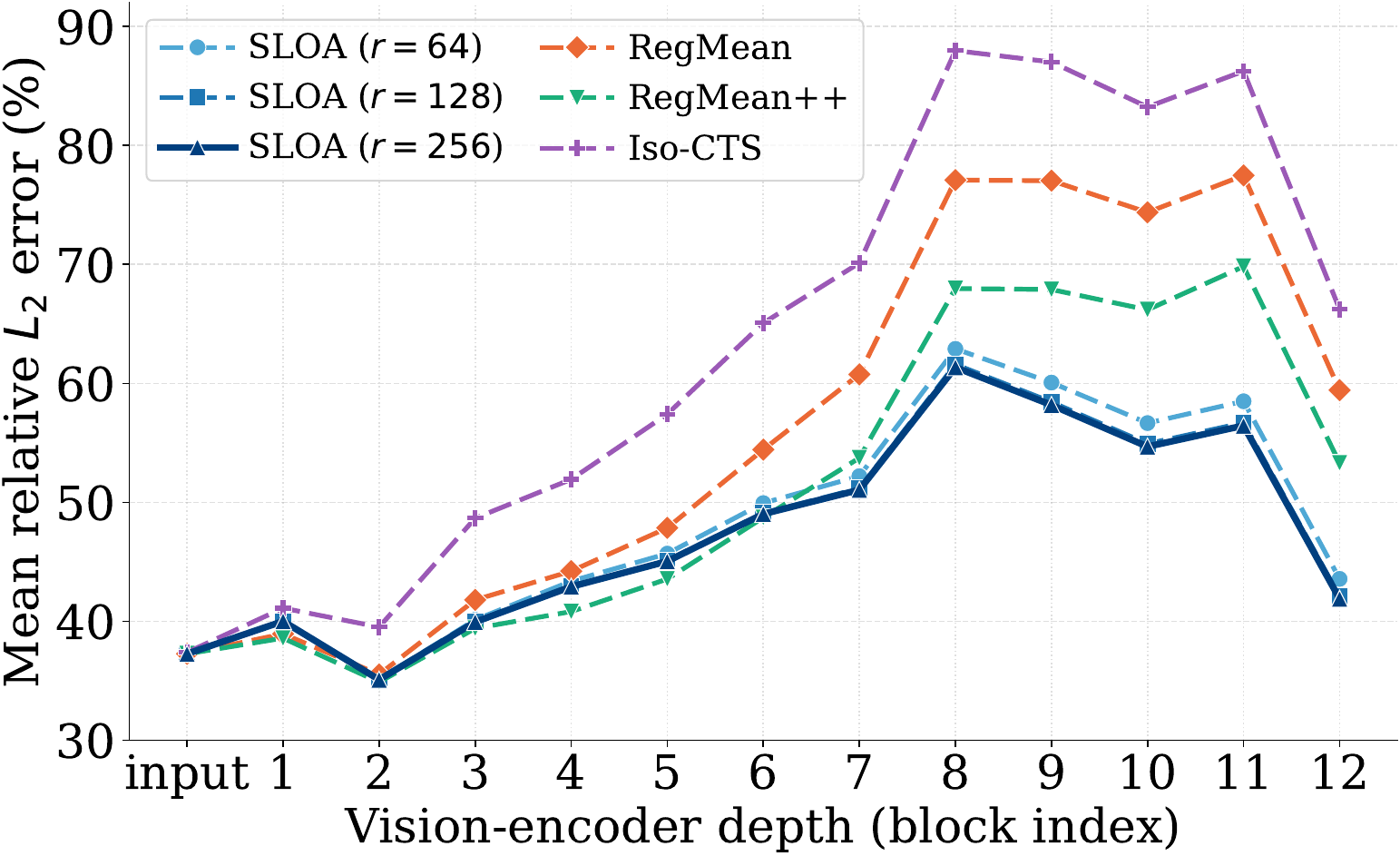}
    \caption{Layer-wise merging error on ViT-B/16.}
    \label{fig:error_propagation}
    \end{subfigure}
    \hspace{0.02\linewidth}
    \begin{subfigure}{0.38\linewidth}
    \centering
    \includegraphics[width=\linewidth]{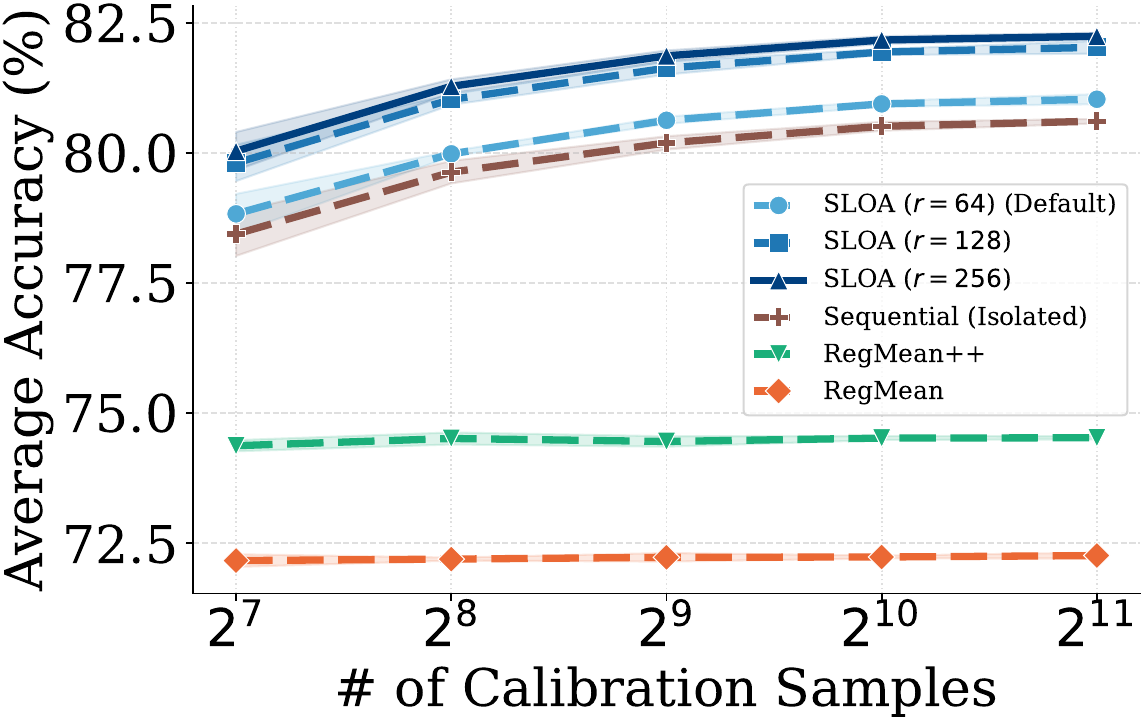}
    \caption{Baseline data scaling on ViT-B/32.}
    \label{fig:sloa_vs_baseline_calib}
    \end{subfigure}
    \caption{Diagnosing the source of SLOA's gains :~\subref {fig:error_propagation} error accumulation across network depth on ViT-B/16, and \subref{fig:sloa_vs_baseline_calib}~merging accuracy versus the number of calibration samples on ViT-B/32, comparing SLOA at increasing rank ($r=64,128,256$) against sequential correction in isolation (no operator composition), RegMean, and RegMean++.}
    \label{fig:ablation_a1_a2}
\end{figure}

To directly assess whether and how merging error compounds across the network's topological ordering of operators, we measure, at each vision-encoder block, the mean relative $L_2$ error between the merged model's activation and the corresponding task-specific expert's activation, averaged across all 20 tasks. We report this on ViT-B/16 in the 20-task setting, the most challenging regime in our vision benchmarks, comparing SLOA ($r \in \{64, 128, 256\}$) against RegMean, RegMean++, and Iso-CTS~(best performing data-free method). Figure~\ref{fig:error_propagation} shows that error compounds with depth for every method, but at sharply different rates. Through the first six blocks, SLOA and RegMean++ track closely; the two only begin to diverge from block 7 onward, once upstream discrepancies have had several blocks to accumulate.  SLOA's error also grows more slowly relative to its own input-level error, peaking at 1.65--1.69$\times$ its starting value versus 1.87$\times$ for RegMean++ and 2.08$\times$ for RegMean. This depth-dependent divergence, rather than a uniform improvement at every block, localizes SLOA's advantage to the later stages of the network, where uncorrected merging error would otherwise compound through the topological ordering of blocks. Expanding the rank from $r=64$ to $r=256$ narrows the error gap further.

\subsection{Decoupling the QK and OV circuits}
\label{sec:decouple_qk_ov_circuit}
To test whether the QK and OV circuits should always use the same rank, we vary $r_{\rm qk}$ and $r_{\rm ov}$ independently on ViT-B/32 with 20 tasks and 1024 calibration samples, while fixing the other rank at 64. Figure~\ref{fig:component-ablation} shows three main trends. First, expanding either circuit alone improves performance only modestly and saturates after roughly $r=256$. Second, QK expansion is more effective than OV expansion, indicating that query-key interactions are a stronger bottleneck for task-specific representation. Third, jointly expanding both circuits gives the best and most sustained gains. Thus, we conclude that the two circuits contribute complementary information, and therefore in practice, scaling $r_{\rm qk}$ and $r_{\rm ov}$ together is the most effective design.

\subsection{Sensitivity to regularization strength}
\label{sec:alpha_sensitivity}
The regularization term in Eq.~(\ref{eq:unified_operator_alignment}) is controlled by $\alpha$ and stabilizes the solution when calibration data is limited. Figure~\ref{fig:alpha-sensitivity} evaluates this effect on ViT-B/32 with 8 tasks. With few samples, especially 64 or 128, larger $\alpha$ improves performance by anchoring the solution toward the task-specific weights. With more samples, 256 or 512, the activation-alignment term is already sufficiently constrained, so strong regularization becomes unnecessary and can reduce performance. The optimal $\alpha$ therefore decreases as the calibration set grows, approaching $\alpha \approx 0.05$ at larger sample sizes. These results suggest that SLOA is robust to $\alpha$ when enough data is available, while the regularizer is most useful in the low-data regime.

\subsection{Component ablation.}
\label{app:component_ablation}
\begin{figure}[t]
\centering
\begin{minipage}{0.312\textwidth}
    \centering
    \includegraphics[width=\linewidth]{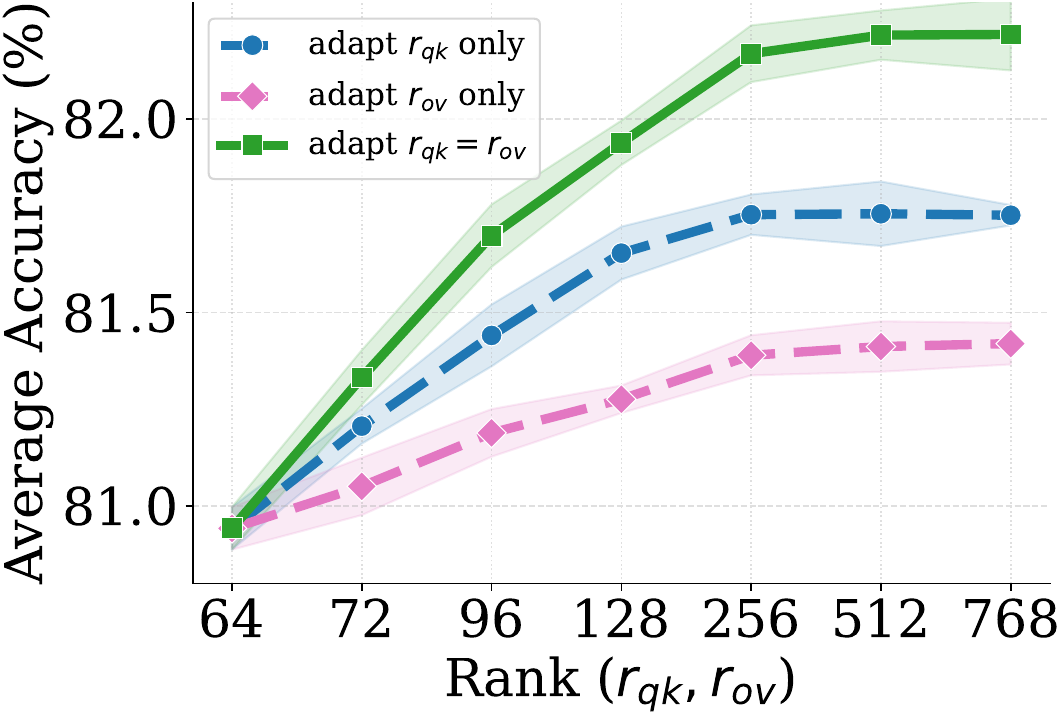}
    \captionof{figure}{Decoupling QK and OV circuit ranks.
    Avg. accuracy of SLOA on ViT-B/32 merging 20 vision tasks.}
    \label{fig:component-ablation}
\end{minipage}
\hfill
\begin{minipage}{0.312\textwidth}
    \centering
    \includegraphics[width=\linewidth]{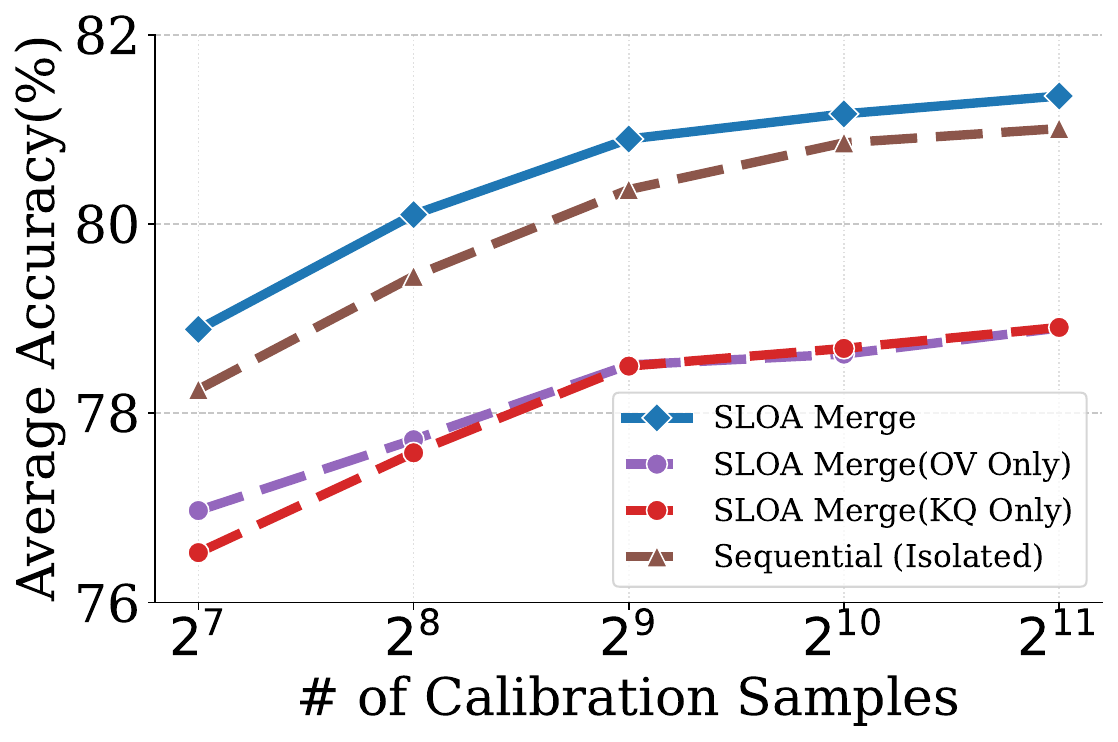}
    \captionof{figure}{Component ablation of SLOA.
    Avg. accuracy on ViT-B/32 merging 20 tasks with different calib. samples.}
    \label{fig:qk-ov-ranks}
\end{minipage}
\hfill
\begin{minipage}{0.312\textwidth}
    \centering
    \includegraphics[width=\linewidth]{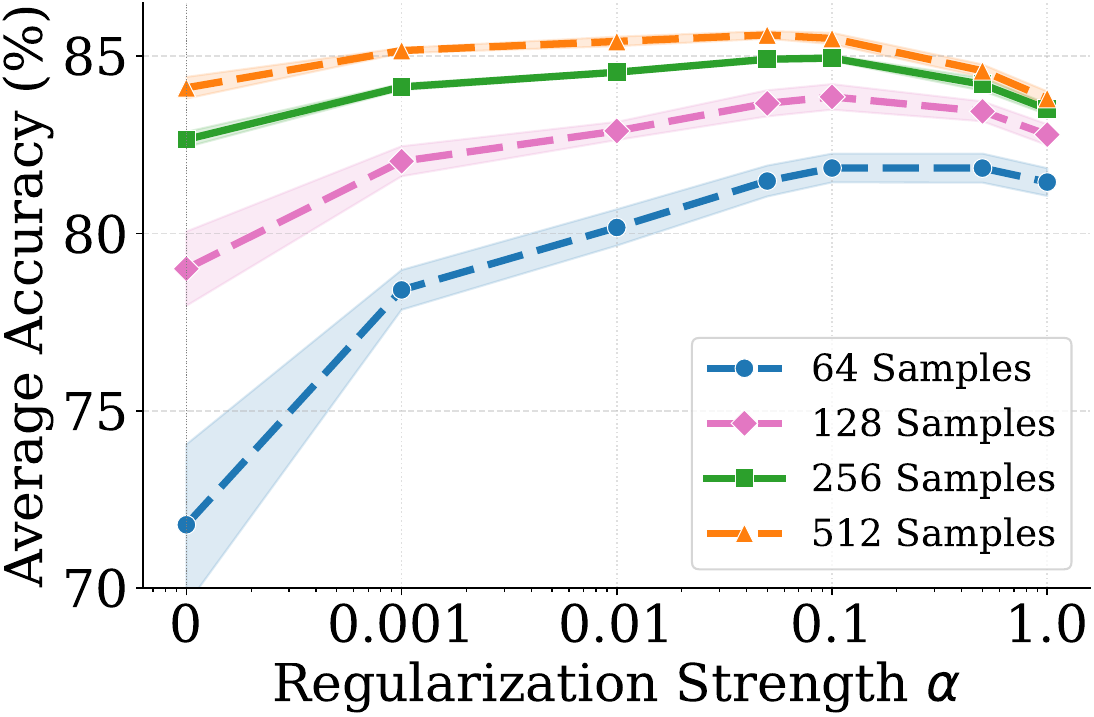}
    \captionof{figure}{Sensitivity of SLOA to $\alpha$.
    Avg. accuracy on ViT-B/32 merging 8 tasks with different regularization strength }
    \label{fig:alpha-sensitivity}
\end{minipage}
\end{figure}

SLOA extends standard linear-layer merging with three components: sequential operator alignment, the QK operator, and the OV operator (\S\ref{sec:local_operators_transformer}). Figure~\ref{fig:qk-ov-ranks} compares four configurations on ViT-B/32 with 20 tasks: sequential alignment with isolated weight merging, sequential alignment plus QK only, sequential alignment plus OV only, and the full SLOA. Sequential alignment alone is already strong, confirming that the sequential operator structure is a major source of improvement. However, adding only QK or only OV degrades performance, as the two circuits are tightly coupled inside self-attention; modifying one without the other introduces a functional error that propagates through later layers. In contrast, combining both operators recovers this loss and improves over sequential alignment alone, showing that QK and OV should be merged jointly.

\subsection{Sequential correction versus training-free baselines.}
Figure~\ref{fig:sloa_vs_baseline_calib} examines how these gains behave as the amount of calibration data changes. RegMean and RegMean++ estimate Gram statistics; once these estimates stabilize, additional calibration samples make little further difference, and both curves plateau early. SLOA's cross-state behavior-matching objective, by contrast, continues to benefit from additional samples, since more calibration data improves alignment across a broader range of inputs. The isolated sequential-correction baseline again outperforms RegMean and RegMean++ throughout, consistent with the component ablation above. Composing the QK and OV operators, however, is what allows their retained rank to be increased beyond what the original per-matrix factorization permits, and doing so yields a further, consistent gain over the isolated sequential baseline at every calibration budget. Sequential correction and operator composition are therefore complementary rather than redundant: the former limits the accumulation of merging error across depth, while the latter supplies the additional capacity needed to represent the union of task-specific directions once that error has been corrected.

\section{Computational Cost and Runtime Analysis}
\label{sec:compute}
\begin{table}[t]
\centering
\caption{Wall-clock merging cost and accuracy when merging 20 task-specific ViT-B/16 checkpoints. Calibration-based methods use $N=256$ samples per task; data-free methods use none.}
\label{tab:merge_cost_baselines}
\begin{tabular}{llrr}
\toprule
Method & Data requirement & Merge time (s) & Accuracy (\%) \\
\midrule
Task Arithmetic & Data-free & 14.63 & 64.23 \\
TIES-Merging & Data-free & 39.64 & 66.31 \\
TSV-M & Data-free & 153.34 & 78.31 \\
Iso-C & Data-free & 17.09 & 75.82 \\
Iso-CTS & Data-free & 167.24 & 78.50 \\
\midrule
Fisher Merging & Calibration-based ($N=256$) & 112.70 & 66.50 \\
RegMean & Calibration-based ($N=256$) & 84.11 & 75.53 \\
RegMean++ & Calibration-based ($N=256$) & 115.08 & 76.64 \\
SLOA (Ours), $r=64$ & Calibration-based ($N=256$) & 175.95 & 82.12 \\
SLOA (Ours), $r=128$ & Calibration-based ($N=256$) & 182.44 & 83.43 \\
SLOA (Ours), $r=256$ & Calibration-based ($N=256$) & 186.74 & 83.74 \\
\bottomrule
\end{tabular}
\end{table}

\subsection{Merging cost versus baselines}
\label{sec:merging_cost_vs_baseline}
Table~\ref{tab:merge_cost_baselines} compares wall-clock merging time and accuracy across all methods when merging 20 task-specific ViT-B/16 checkpoints, using $N=256$ calibration samples per task for the calibration-based methods. Being data-free is not synonymous with being computationally inexpensive: for example, TSV-M and Iso-CTS take comparable time due to concatenating weight matrices across different tasks and performing singular value decomposition. SLOA's dominant offline costs are activation propagation through the merged network, solving the local operator objective at each layer, and defactorizing the composed operators back into weight updates. Critically, this cost is incurred once, when constructing the merged checkpoint: we pay it a single time, and the resulting accuracy gain persists over the model's entire deployment lifetime. After merging, SLOA requires no task identity, routing, test-time calibration, or online adaptation -- the merged checkpoint is used exactly like any single fine-tuned model. Increasing the retained rank $r$ increases merging time only modestly (from 175.95s to 186.74s, from $r=64$ to $r=256$), since it enlarges the local operators being solved and defactorized rather than introducing new computation; we quantify the resulting effects on memory usage and inference cost separately below.

\subsection{Scaling cost with task count and calibration budget}
\label{sec:merging_vs_task_counts}
Tables~\ref{tab:scaling_tasks} and~\ref{tab:scaling_calib} characterize how merging cost scales along the two axes: the number of tasks being merged, and the calibration budget. As the number of tasks grows from 8 to 20, accuracy decreases on both backbones, reflecting the increasing interference among a larger and more heterogeneous task pool, while merge time grows roughly in line with the number of task-specific checkpoints that must be processed. Accuracy improves monotonically with more calibration data on both backbones. Peak GPU memory usage is reported in Tables~\ref{tab:memory_tasks} and~\ref{tab:memory_calib}. GPU usage is a function of the number of parameters, the amount of data processed for operator alignment, and the number of tasks we need to merge. ViT-B/32 stays under 6GB across every setting tested, while ViT-B/16 grows to 20.2GB at $N=2048$, still within a single modern GPU budget.

\begin{table}[t]
\centering
\caption{Accuracy and merge time versus the number of merged tasks, at $N=256$ calibration samples per task and $r=128$.}
\label{tab:scaling_tasks}
\begin{tabular}{c cc cc}
\toprule
& \multicolumn{2}{c}{ViT-B/32} & \multicolumn{2}{c}{ViT-B/16} \\
\cmidrule(lr){2-3}\cmidrule(lr){4-5}
\# Tasks & Accuracy (\%) & Merge time (s) & Accuracy (\%) & Merge time (s) \\
\midrule
8  & 85.50 $\pm$ 0.05 & 74.2 $\pm$ 0.5  & 88.04 $\pm$ 0.24 & 107.50 $\pm$ 1.76 \\
14 & 82.61 $\pm$ 0.04 & 103.3 $\pm$ 3.1 & 85.16 $\pm$ 0.09 & 136.50 $\pm$ 0.49 \\
20 & 81.03 $\pm$ 0.12 & 130.6 $\pm$ 1.3 & 83.43 $\pm$ 0.09 & 180.49 $\pm$ 1.88 \\
\bottomrule
\end{tabular}
\end{table}

\begin{table}[t]
\centering
\caption{Accuracy and merge time versus calibration samples per task, at 20 tasks and $r=128$.}
\label{tab:scaling_calib}
\begin{tabular}{c cc cc}
\toprule
& \multicolumn{2}{c}{ViT-B/32} & \multicolumn{2}{c}{ViT-B/16} \\
\cmidrule(lr){2-3}\cmidrule(lr){4-5}
$N$ & Accuracy (\%) & Merge time (s) & Accuracy (\%) & Merge time (s) \\
\midrule
128  & 79.77 $\pm$ 0.38 & 99.31 $\pm$ 2.13  & 82.67 $\pm$ 0.12 & 124.96 $\pm$ 0.92 \\
256  & 81.03 $\pm$ 0.12 & 130.64 $\pm$ 1.26 & 83.43 $\pm$ 0.09 & 180.49 $\pm$ 1.88 \\
512  & 81.63 $\pm$ 0.13 & 147.59 $\pm$ 3.28 & 83.86 $\pm$ 0.08 & 245.69 $\pm$ 0.58 \\
1024 & 81.92 $\pm$ 0.08 & 177.78 $\pm$ 1.89 & 84.12 $\pm$ 0.11 & 376.18 $\pm$ 0.57 \\
2048 & 82.01 $\pm$ 0.07 & 252.40 $\pm$ 2.37 & 84.24 $\pm$ 0.03 & 635.90 $\pm$ 2.24 \\
\bottomrule
\end{tabular}
\end{table}

\begin{table}
\centering
\caption{Peak GPU memory (GB) vs. the number of merged tasks, at $N=256$ calib. samples.}
\label{tab:memory_tasks}
\begin{tabular}{lccc}
\toprule
Backbone & 8 tasks & 14 tasks & 20 tasks \\
\midrule
ViT-B/32 & 2.1 & 2.5 & 2.9 \\
ViT-B/16 & 7.0 & 8.0 & 9.0 \\
\bottomrule
\end{tabular}
\end{table}
\begin{table}[H]
\centering
\caption{Peak GPU memory (GB) versus calib. samples per task, at 8 tasks.}
\label{tab:memory_calib}
\begin{tabular}{lccccc}
\toprule
Backbone & $N=128$ & $N=256$ & $N=512$ & $N=1024$ & $N=2048$ \\
\midrule
ViT-B/32 & 1.4 & 2.1 & 2.4 & 3.2 & 5.5 \\
ViT-B/16 & 3.8 & 7.0 & 8.1 & 10.9 & 20.2 \\
\bottomrule
\end{tabular}
\end{table}

\begin{table}[H]
\centering
\caption{Inference cost by deployed rank on ViT-B/16 (20-task merge). Inference measured with batch size 64, FP16, on a single A100 GPU, averaged over 100 iterations.}
\label{tab:inference_cost}
\resizebox{\textwidth}{!}{
\begin{tabular}{lcrrrr}
\toprule
Deployed model & QK/VO rank & Accuracy (\%) & Throughput (img/s) $\uparrow$ & Latency (ms/img) $\downarrow$ & Peak memory (GB) $\downarrow$ \\
\midrule
All baseline merges (shared original architecture) & 64 / 64 & 64.23--78.50 & $\approx$3257 & 0.307 & 0.60 \\
SLOA (original rank, $r=64$) & 64 / 64 & 82.37 & 3261.1 & 0.307 & 0.60 \\
SLOA (expanded rank, $r=128$) & 128 / 128 & 83.53 & 2711.5 & 0.369 & 0.65 \\
SLOA (expanded rank, $r=256$) & 256 / 256 & 83.74 & 1999.1 & 0.500 & 0.83 \\
\bottomrule
\end{tabular}
}
\end{table}
\subsection{Inference cost}
\label{sec:inference_cost}
Table~\ref{tab:inference_cost} isolates the cost of deploying the merged model itself, separate from the one-time merging cost reported above. Since every baseline in Table~\ref{tab:merge_cost_baselines} merges by combining weights within the original 64/64 attention-head architecture, all of them, share an identical inference profile: throughput of approximately 3257 images/s, 0.307 ms/image latency, and 0.60GB peak memory. At the default rank ($r=64$, no expansion), SLOA's merged model is indistinguishable in inference cost from every baseline, while still improving accuracy to 82.37\%. Rank expansion to $r=128$ or $r=256$ increases inference time and GPU memory, both direct consequences of the wider attention head, in exchange for further accuracy gains (83.53\% and 83.74\%, respectively). This is a deliberate, user-controlled trade-off between capacity and test-time cost.

\textbf{Computational resources.} We performed our SLOA merging method and conducted all the ablation studies, including benchmarking on a single NVIDIA DGX A100 with 48 GB memory.

\section{Task-wise full tables}
\label{sec:full_tables}
\begin{table}[t]
\centering
\caption{
Multi-task performance on seven GLUE tasks with merged RoBERTa models. Training-free methods use 256 calibration samples. SLOA denotes rank-preserving operator alignment, while SLOA ($r=128$) expands the factorization rank of the merged QK and VO operators.
}
\label{tab:roberta}
\begin{center}
\resizebox{\textwidth}{!}{%
\begin{tabular}{clccccccccc}
\toprule
\multirow{2}{*}{}
& \multirow{2}{*}{Method}
& \multirow{2}{*}{\# Samples}
& \multicolumn{7}{c}{Accuracy}
& \multicolumn{1}{c}{\cellcolor{avggray}\textbf{Average}} \\
\cmidrule(lr){4-10}\cmidrule(lr){11-11}
& & & CoLA & SST2 & MRPC & QQP & MNLI & QNLI & RTE
& \cellcolor{avggray}\textbf{Avg. $\uparrow$} \\
\midrule
& Individual & --
& 0.6018 & 0.9404 & 0.8922 & 0.9141 & 0.8720 & 0.9271 & 0.7906
& \avgcell{0.8483} \\

\midrule
\multirow{6}{*}{\rotatebox[origin=c]{90}{Data-Free}}
& Weight Averaging & --
& 0.1808 & 0.8188 & 0.8116 & 0.7381 & 0.4383 & 0.7106 & 0.6173
& \avgcell{0.6165} \\

& Task Arithmetic & --
& 0.2909 & 0.8463 & 0.8188 & 0.7869 & 0.5591 & 0.7584 & 0.6462
& \avgcell{0.6724} \\

& TIES-Merging & --
& 0.1951 & 0.8429 & 0.8333 & \textbf{0.8351} & 0.6501 & 0.7309 & 0.4260
& \avgcell{0.6448} \\
& TSV-M & --
& 0.3477 & 0.8739 & 0.8446 & \underline{0.8286} & 0.6163 & 0.8124 & 0.6606
& \avgcell{0.7120} \\

& Iso-C & --
& 0.2639 & 0.8601 & 0.8130 & 0.8211 & 0.7303 & 0.7882 & 0.7040
& \avgcell{0.7115} \\

& Iso-CTS & --
& 0.4136 & 0.7798 & 0.6385 & 0.6500 & 0.5277 & 0.6590 & 0.6534
& \avgcell{0.6174} \\

\midrule
\multirow{5}{*}{\rotatebox[origin=c]{90}{Training-Free}}
& Fisher Merging & 256
& 0.1126 & 0.8719 & 0.7419 & 0.7808 & 0.4227 & 0.5311 & 0.5150
& \avgcell{0.5680} \\

& RegMean & 256
& 0.2797 & 0.8933 & 0.8069 & 0.7915 & 0.7251 & 0.8336 & 0.6161
& \avgcell{0.7066} \\

& RegMean++ & 256
& 0.2433 & 0.9170 & 0.7740 & 0.8099 & 0.7543 & 0.8299 & 0.6691
& \avgcell{0.7139} \\

& \cellcolor{lightpurple}SLOA ($r{=}64$) & \cellcolor{lightpurple}256
& \cellcolor{lightpurple}\underline{0.5608} & \cellcolor{lightpurple}\underline{0.9232} & \cellcolor{lightpurple}\textbf{0.8814} & \cellcolor{lightpurple}0.8152 & \cellcolor{lightpurple}\underline{0.7678} & \cellcolor{lightpurple}\underline{0.8755} & \cellcolor{lightpurple}\underline{0.7100}
& \cellcolor{lightpurple}\underline{0.7906} \\

& \cellcolor{lightblue}SLOA ($r{=}128$) & \cellcolor{lightblue}256
& \cellcolor{lightblue}\textbf{0.5721} & \cellcolor{lightblue}\textbf{0.9239} & \cellcolor{lightblue}\underline{0.8810} & \cellcolor{lightblue}0.8166 & \cellcolor{lightblue}\textbf{0.7753} & \cellcolor{lightblue}\textbf{0.8824} & \cellcolor{lightblue}\textbf{0.7341}
& \cellcolor{lightblue}\textbf{0.7979} \\

\bottomrule
\end{tabular}
}
 \end{center}
\end{table}

This section reports the per-task results behind the averages in the main paper. Tables~\ref{tab:vitb32_8tasks_full}, \ref{tab:vitb32_14tasks_full}, and~\ref{tab:vitb32_20tasks_full} extend Table~\ref{tab:vision_main_results_b32} with task-wise accuracies for ViT-B/32 merged over 8, 14, and 20 vision tasks, respectively. Table~\ref{tab:roberta} likewise extends Table~\ref{tab:roberta_avg} with the per-task scores for RoBERTa on the seven GLUE tasks. All methods share the setup of the main experiments: calibration-based methods use $N{=}256$ samples per task, and SLOA is reported at the original rank ($r{=}64$) and with rank expansion ($r{=}128$). SLOA's average gains are broad rather than driven by a few tasks. Data-free methods use no calibration data, so for fixed checkpoints and hyperparameters, their merged model, and therefore their accuracy, is the same in every run; we report standard deviations only for calibration-based methods, where the variation comes from the random draw of calibration samples.
\begin{table}[t]
\centering
\caption{Per-task accuracy (\%) of merged ViT-B/32 models on 8 vision tasks. Training-free methods use $N{=}256$ calibration samples per task; their averages are mean$\pm$std over three seeds. For SLOA, $r$ is the retained rank of the QK and VO operators ($r{=}64$ preserves the original architecture). Bold/underline marks the best/second-best merging method per column. Data-free methods are deterministic given the checkpoints, so standard deviations are reported only for calibration-based methods, computed over three random draws of the calibration set.}
\label{tab:vitb32_8tasks_full}
\resizebox{\ifdim\width>\textwidth\textwidth\else\width\fi}{!}{%
\begin{tabular}{lccccccccc}
\toprule
Method & SUN397 & GTSRB & Cars & EuroSAT & RESISC45 & DTD & MNIST & SVHN & Avg. \\
\midrule
Fine-tuned (individual) & 74.90 & 98.91 & 78.47 & 99.07 & 95.14 & 79.79 & 99.58 & 97.27 & 90.39 \\
\midrule
Simple Avg. & 65.39 & 54.92 & 62.42 & 75.67 & 70.63 & 50.53 & 86.28 & 64.49 & 66.29 \\
Task Arithmetic & 64.30 & 62.71 & 61.42 & 78.30 & 70.57 & 51.76 & 93.01 & 73.88 & 69.49 \\
TIES-Merging & 61.40 & 78.68 & 60.15 & 69.04 & 69.62 & 51.06 & 97.72 & 85.92 & 71.70 \\
TSV-M & 68.59 & 91.27 & \underline{72.01} & 93.41 & 85.14 & 64.36 & 98.76 & 90.79 & 83.04 \\
Iso-C & 67.73 & 93.51 & 71.61 & 90.00 & 84.98 & 65.05 & 98.82 & 89.39 & 82.64 \\
Iso-CTS & 68.30 & 92.75 & \textbf{72.39} & 88.81 & 85.97 & 67.87 & 98.76 & 88.09 & 82.87 \\
\midrule
Fisher Merging & 65.84 & 61.71 & 65.74 & 79.74 & 76.54 & 54.47 & 85.28 & 63.67 & 69.12{\scriptsize$\pm$0.13} \\
RegMean & \underline{69.03} & 83.56 & 70.61 & 95.64 & 85.50 & 65.87 & 98.41 & 92.41 & 82.63{\scriptsize$\pm$0.11} \\
RegMean++ & \textbf{69.34} & 90.29 & 70.75 & 95.64 & 86.89 & 68.60 & 98.98 & 94.45 & 84.37{\scriptsize$\pm$0.12} \\
SLOA (r=64) & 65.15 & \underline{95.05} & 66.92 & \underline{97.62} & \underline{89.29} & \underline{69.80} & \underline{99.20} & \underline{96.07} & \underline{84.89}{\scriptsize$\pm$0.08} \\
SLOA (r=128) & 66.11 & \textbf{95.70} & 68.35 & \textbf{97.65} & \textbf{90.22} & \textbf{70.35} & \textbf{99.28} & \textbf{96.31} & \textbf{85.50}{\scriptsize$\pm$0.05} \\
\bottomrule
\end{tabular}%
}
\end{table}

\begin{table}[t]
\centering
\caption{Per-task accuracy (\%) of merged ViT-B/32 models on 14 vision tasks. Setup and notation as in Table~\ref{tab:vitb32_8tasks_full}.}
\label{tab:vitb32_14tasks_full}
\resizebox{\ifdim\width>\textwidth\textwidth\else\width\fi}{!}{%
\begin{tabular}{lccccccccccccccc}
\toprule
Method & {SUN397} & {GTSRB} & {Cars} & {EuroSAT} & {RESISC45} & {DTD} & {MNIST} & {SVHN} & {Flowers} & {PCAM} & {FER2013} & {Pets} & {STL10} & {CIFAR100} & {Avg.} \\
\midrule
Fine-tuned (individual) & 74.90 & 98.91 & 78.47 & 99.07 & 95.14 & 79.79 & 99.58 & 97.27 & 88.62 & 87.96 & 71.62 & 92.37 & 97.55 & 88.38 & 89.26 \\
\midrule
Simple Avg. & 64.73 & 45.63 & 60.40 & 66.96 & 67.06 & 47.13 & 76.57 & 50.66 & 67.25 & 65.27 & 51.62 & 84.25 & 97.20 & 70.39 & 65.37 \\
Task Arithmetic & 64.41 & 50.00 & 59.63 & 67.74 & 67.29 & 47.77 & 80.75 & 53.93 & 66.17 & 69.80 & 53.05 & 84.25 & 96.62 & 69.14 & 66.47 \\
TIES-Merging & 62.10 & 63.95 & 54.58 & 62.93 & 65.29 & 50.21 & 92.56 & 65.70 & 58.22 & 77.10 & 54.95 & 81.30 & 94.84 & 62.44 & 67.58 \\
TSV-M & 67.30 & 87.80 & 63.85 & 91.04 & 81.22 & 61.17 & 98.35 & 85.23 & 68.01 & 82.03 & \underline{65.53} & 87.82 & 97.18 & 71.06 & 79.11 \\
Iso-C & \underline{70.02} & 81.45 & 63.66 & 83.85 & 80.10 & 60.96 & 96.36 & 75.98 & 70.34 & 83.29 & 65.30 & 87.87 & 97.08 & \textbf{74.99} & 77.94 \\
Iso-CTS & \textbf{70.31} & 83.76 & \textbf{68.42} & 85.30 & 81.54 & 62.18 & 97.03 & 76.04 & 74.58 & 83.16 & \textbf{67.02} & 87.87 & \underline{97.29} & \underline{74.86} & 79.24 \\
\midrule
Fisher Merging & 64.51 & 47.80 & 62.96 & 66.09 & 69.47 & 48.62 & 73.94 & 48.50 & 69.42 & 64.12 & 56.02 & 87.68 & 96.96 & 71.58 & 66.26{\scriptsize$\pm$0.21} \\
RegMean & 66.88 & 72.91 & 66.17 & 91.63 & 79.68 & 58.94 & 96.65 & 81.92 & 72.92 & 80.23 & 63.80 & 88.67 & 97.02 & 73.65 & 77.93{\scriptsize$\pm$0.04} \\
RegMean++ & 67.54 & 80.60 & \underline{67.04} & 94.38 & 81.78 & 60.73 & 98.03 & 91.25 & 75.96 & 77.41 & 64.67 & \textbf{90.08} & \textbf{97.47} & 73.95 & 80.06{\scriptsize$\pm$0.03} \\
SLOA (r=64) & 62.68 & \underline{88.83} & 63.19 & \underline{96.96} & \underline{85.97} & \underline{66.03} & \underline{98.95} & \underline{94.99} & \underline{77.62} & \underline{87.95} & 64.13 & 89.47 & 97.13 & 73.26 & \underline{81.94}{\scriptsize$\pm$0.01} \\
SLOA (r=128) & 63.18 & \textbf{90.49} & 64.62 & \textbf{97.17} & \textbf{87.25} & \textbf{66.77} & \textbf{99.03} & \textbf{95.32} & \textbf{78.07} & \textbf{88.24} & 64.84 & \underline{89.79} & 97.21 & 74.49 & \textbf{82.61}{\scriptsize$\pm$0.04} \\
\bottomrule
\end{tabular}%
}
\end{table}

\begin{table}[t]
\centering
\caption{Per-task accuracy (\%) of merged ViT-B/32 models on 20 vision tasks. Setup and notation as in Table~\ref{tab:vitb32_8tasks_full}.}
\label{tab:vitb32_20tasks_full}
\resizebox{\ifdim\width>\textwidth\textwidth\else\width\fi}{!}{%
\begin{tabular}{lccccccccccccccccccccc}
\toprule
Method & {SUN397} & {GTSRB} & {Cars} & {EuroSAT} & {RESISC45} & {DTD} & {MNIST} & {SVHN} & {Flowers} & {PCAM} & {FER2013} & {Pets} & {STL10} & {CIFAR100} & {CIFAR10} & {Food101} & {FMNIST} & {EMNIST} & {KMNIST} & {SST2} & {Avg.} \\
\midrule
Fine-tuned (individual) & 74.90 & 98.91 & 78.47 & 99.07 & 95.14 & 79.79 & 99.58 & 97.27 & 88.62 & 87.96 & 71.62 & 92.37 & 97.55 & 88.38 & 97.60 & 88.40 & 94.75 & 95.62 & 98.23 & 71.28 & 89.78 \\
\midrule
Simple Avg. & 64.22 & 43.14 & 59.56 & 60.85 & 64.76 & 46.44 & 71.79 & 47.22 & 66.29 & 63.95 & 50.14 & 83.92 & 96.99 & 69.80 & 92.66 & 80.38 & 71.31 & 14.98 & 11.41 & 61.78 & 61.08 \\
Task Arithmetic & 62.03 & 48.95 & 53.74 & 58.15 & 60.87 & 45.96 & 79.45 & 48.43 & 60.95 & 73.45 & 51.32 & 82.23 & 94.89 & 64.64 & 91.38 & 71.88 & 73.86 & 17.81 & 12.14 & 59.86 & 60.60 \\
TIES-Merging & 65.02 & 49.22 & 59.77 & 60.78 & 66.22 & 48.51 & 79.27 & 52.37 & 66.55 & 66.75 & 51.66 & 84.06 & \underline{97.05} & 70.40 & 93.33 & 80.45 & 72.43 & 17.32 & 12.16 & 61.94 & 62.76 \\
TSV-M & 65.50 & 83.79 & 55.80 & 87.11 & 75.76 & 58.62 & 97.61 & 80.28 & 65.00 & 80.57 & \underline{64.29} & 86.29 & 96.55 & 68.92 & 93.89 & 76.37 & 84.27 & 35.91 & 46.48 & \underline{70.18} & 73.66 \\
Iso-C & \textbf{67.13} & 80.72 & 47.76 & 78.19 & 73.16 & 57.07 & 96.82 & 75.47 & 63.16 & 79.79 & 62.75 & 83.73 & 96.05 & 71.96 & 94.36 & 73.38 & 82.60 & 33.99 & 35.11 & 70.02 & 71.16 \\
Iso-CTS & \underline{66.61} & 82.52 & 57.67 & 80.74 & 75.13 & 57.87 & 97.10 & 75.94 & 70.56 & 83.69 & \textbf{64.54} & 84.96 & 96.58 & \textbf{72.95} & 94.62 & 74.06 & 83.58 & 38.29 & 40.48 & \textbf{70.35} & 73.41 \\
\midrule
Fisher Merging & 63.94 & 43.62 & 61.63 & 59.00 & 65.82 & 46.49 & 68.27 & 45.50 & 68.09 & 62.36 & 52.94 & 86.39 & 96.95 & 70.35 & 93.07 & 81.16 & 70.36 & 16.08 & 13.84 & 68.95 & 61.74{\scriptsize$\pm$0.06} \\
RegMean & 65.74 & 65.87 & \underline{63.88} & 88.40 & 77.14 & 55.85 & 91.32 & 74.54 & 70.66 & 79.60 & 61.76 & 87.48 & 96.53 & 72.14 & 94.58 & \underline{82.21} & 80.06 & 22.75 & 44.82 & 68.63 & 72.20{\scriptsize$\pm$0.03} \\
RegMean++ & 66.55 & 73.95 & \textbf{65.21} & 92.90 & 79.34 & 57.87 & 93.65 & 87.43 & 73.48 & 75.69 & 63.55 & \textbf{89.26} & \textbf{97.42} & \underline{72.83} & \underline{94.92} & \textbf{83.57} & 82.97 & 29.06 & 43.70 & 66.94 & 74.51{\scriptsize$\pm$0.12} \\
SLOA (r=64) & 60.23 & \underline{84.31} & 59.91 & \underline{96.26} & \underline{83.66} & \underline{63.67} & \underline{98.01} & \underline{93.48} & \underline{75.94} & \underline{87.21} & 63.12 & 88.03 & 96.95 & 70.68 & 94.66 & 78.30 & \underline{90.06} & \underline{72.88} & \underline{82.88} & 61.01 & \underline{80.06}{\scriptsize$\pm$0.18} \\
SLOA (r=128) & 60.85 & \textbf{86.24} & 61.50 & \textbf{96.54} & \textbf{85.05} & \textbf{64.11} & \textbf{98.24} & \textbf{94.01} & \textbf{76.45} & \textbf{87.92} & 63.82 & \underline{88.65} & 97.00 & 71.64 & \textbf{95.01} & 79.50 & \textbf{90.60} & \textbf{76.03} & \textbf{85.11} & 62.27 & \textbf{81.03}{\scriptsize$\pm$0.12} \\
\bottomrule
\end{tabular}%
}
\end{table}

\end{document}